\documentclass[journal]{IEEEtran}

\IEEEoverridecommandlockouts 

\makeatletter
\def\cl@chapter{\@elt {theorem}}
\makeatother

\usepackage{times}

\usepackage{textcomp}
\usepackage{lipsum}
\usepackage{flushend}

\usepackage[utf8]{inputenc}

\usepackage{url}
\usepackage{ifpdf}

\ifpdf
\usepackage[table]{xcolor}
\usepackage{graphics}
\usepackage[pdftex]{graphicx}
\DeclareGraphicsExtensions{.pdf,.PDF,.png,.PNG,.jpg,.JPG}
\usepackage{hyperref}
\hypersetup{
  colorlinks,%
  citecolor=black,%
  filecolor=black,%
  linkcolor=black,%
  urlcolor=black
}
\else
\usepackage{graphicx}
\DeclareGraphicsExtensions{.eps,.EPS,.ps,.PS,}
\fi

\usepackage{amsmath}
\usepackage{amssymb} 
\usepackage{stmaryrd}           
\usepackage{mathptmx}
\usepackage[mathscr]{euscript}
\usepackage{bm}
\usepackage{nicefrac}

\DeclareMathAlphabet{\mathcal}{OMS}{cmsy}{m}{n} 

\newcommand\imCMsym[4][\mathord]{%
  \DeclareFontFamily{U} {#2}{}
  \DeclareFontShape{U}{#2}{m}{n}{
    <-6> #25
    <6-7> #26
    <7-8> #27
    <8-9> #28
    <9-10> #29
    <10-12> #210
    <12-> #212}{}
  \DeclareSymbolFont{CM#2} {U} {#2}{m}{n}
  \DeclareMathSymbol{#4}{#1}{CM#2}{#3}
}

\imCMsym{cmmi}{124}{\CMjmath}

\usepackage[ruled,vlined]{algorithm2e}

\usepackage{tabularx}
\usepackage{balance}
\usepackage{booktabs}
\usepackage{multirow}
\usepackage{multicol}
\usepackage{array}
\usepackage{adjustbox}
\usepackage[british]{babel}
\usepackage{dcolumn}
\usepackage{colortbl}
\usepackage{threeparttable}
\usepackage{enumitem}

\usepackage[binary-units=true]{siunitx}

\makeatletter
\if@twocolumn
  \renewcommand\normalsize{%
   \@setfontsize\normalsize\@xpt{12.5pt}%
   \abovedisplayskip=3 mm plus6pt minus 4pt
   \belowdisplayskip=3 mm plus6pt minus 4pt
   \abovedisplayshortskip=0.0 mm plus6pt
   \belowdisplayshortskip=2 mm plus4pt minus 4pt
   \let\@listi\@listI}%

  \renewcommand\small{%
   \@setfontsize\small{8.5pt}\@xpt
   \abovedisplayskip 8.5\p@ \@plus3\p@ \@minus4\p@
   \abovedisplayshortskip \z@ \@plus2\p@
   \belowdisplayshortskip 4\p@ \@plus2\p@ \@minus2\p@
   \def\@listi{\leftmargin\leftmargini
               \parsep 0\p@ \@plus1\p@ \@minus\p@
               \topsep 4\p@ \@plus2\p@ \@minus4\p@
               \itemsep0\p@}%
   \belowdisplayskip \abovedisplayskip}

\else
  \if@smallext
   \renewcommand\normalsize{%
   \@setfontsize\normalsize\@xpt\@xiipt
   \abovedisplayskip=3 mm plus6pt minus 4pt
   \belowdisplayskip=3 mm plus6pt minus 4pt
   \abovedisplayshortskip=0.0 mm plus6pt
   \belowdisplayshortskip=2 mm plus4pt minus 4pt
   \let\@listi\@listI}%

  \renewcommand\small{%
   \@setfontsize\small\@viiipt{9.5pt}%
   \abovedisplayskip 8.5\p@ \@plus3\p@ \@minus4\p@
   \abovedisplayshortskip \z@ \@plus2\p@
   \belowdisplayshortskip 4\p@ \@plus2\p@ \@minus2\p@
   \def\@listi{\leftmargin\leftmargini
               \parsep 0\p@ \@plus1\p@ \@minus\p@
               \topsep 4\p@ \@plus2\p@ \@minus4\p@
               \itemsep0\p@}%
   \belowdisplayskip \abovedisplayskip}
 \else
  \renewcommand\normalsize{%
   \@setfontsize\normalsize{9.5pt}{11.5pt}%
   \abovedisplayskip=3 mm plus6pt minus 4pt
   \belowdisplayskip=3 mm plus6pt minus 4pt
   \abovedisplayshortskip=0.0 mm plus6pt
   \belowdisplayshortskip=2 mm plus4pt minus 4pt
   \let\@listi\@listI}%

  \renewcommand\small{%
   \@setfontsize\small\@viiipt{9.25pt}%
   \abovedisplayskip 8.5\p@ \@plus3\p@ \@minus4\p@
   \abovedisplayshortskip \z@ \@plus2\p@
   \belowdisplayshortskip 4\p@ \@plus2\p@ \@minus2\p@
   \def\@listi{\leftmargin\leftmargini
               \parsep 0\p@ \@plus1\p@ \@minus\p@
               \topsep 4\p@ \@plus2\p@ \@minus4\p@
               \itemsep0\p@}%
   \belowdisplayskip \abovedisplayskip}
  \fi
\let\footnotesize\small
\makeatother

\usepackage{microtype}

\usepackage{scrtime}
\usepackage[dont-mess-around]{fnpct}

\usepackage{etoolbox}

\newcolumntype{P}[1]{>{\centering\arraybackslash}p{#1}}

\usepackage[capitalise,noabbrev]{cleveref}

\usepackage{bbm}
\usepackage{pifont}
\usepackage{tikz}
\newcommand{\tikzcircle}[2][red,fill=red]{\tikz[baseline=-0.5ex]\draw[#1,radius=#2] (0,0) circle ;}%

\DeclareMathOperator*{\argmin}{arg\,min}
\makeatletter
\newcommand{\algorithmfootnote}[2][\footnotesize]{%
  \let\old@algocf@finish\@algocf@finish
  \def\@algocf@finish{\old@algocf@finish
    \leavevmode\rlap{\begin{minipage}{\linewidth}
    #1#2
    \end{minipage}}%
  }%
}
\makeatother

\usepackage[nolist,nohyperlinks]{acronym}
\newacro{CNN}{Convolutional Neural Network}
\newacro{DNN}{Deep Neural Network}
\newacro{GPS}{Global Positioning System}
\newacro{GNSS}{Global Navigation Satellite System}
\newacro{NLOS}{non-line-of-sight}
\newacro{ADAS}{Advanced Driver Assistance Systems}
\newacro{LIDAR}[LiDAR]{Light Detection and Ranging}
\newacro{HD map}{High Definition map}
\newacro{EV}{Embedding Vector}
\newacro{SLAM}{Simultaneous Localization and Mapping}
\newacro{MLP}{Multi Layer Perceptron}
\newacro{IMU}{Inertial Measurement Unit}
\newacro{ML}{Machine Learning}
\newacro{SfM}{Structure from Motion}
\newacro{PnP}{Perspective-n-Points}
\newacro{ASPP}{Atrous Spatial Pyramid Pooling}
\newacro{RANSAC}{RANdom SAmple Consensus}
\newacro{CV}{Computer Vision}
\newacro{VLAD}{Vector of Locally Aggregated Descriptor}
\newacro{SVD}{Singular Value Decomposition}
\newacro{ICP}{Iterative Closest Point}
\newacro{RMSE}{Root Mean Square Error}
\newacro{UAV}{Unmanned Aerial Vehicle}

\definecolor{pvgreen}{RGB}{145,188,128}
\definecolor{pvblue}{RGB}{97,145,189}
\definecolor{pvorange}{RGB}{224,151,37}
\definecolor{pvpurple}{RGB}{151,117,164}
\definecolor{pathred}{RGB}{228, 26, 28}
\definecolor{pathgreen}{RGB}{77, 175, 74}

\newcommand{\globaldesc}{f(P)}

\newcommand{\algref}[1]{Algo.~\ref{#1}}

\def\net{LCDNet}

\begin{document}

\title{\net: Deep Loop Closure Detection and\\Point Cloud Registration for LiDAR SLAM}

\author{Daniele Cattaneo$^{1}$, Matteo Vaghi$^{2}$, Abhinav Valada$^{1}$
\thanks{$^{1}$ Department of Computer Science, University of Freiburg, Germany.}
\thanks{$^{2}$ Department Informatica, Sistemistica e Comununicazioni, Università degli studi di Milano - Bicocca, Italy.}
}
\markboth{\textcopyright IEEE 2022}%
{Shell \MakeLowercase{\textit{et al.}}: Bare Demo of IEEEtran.cls for IEEE Journals}

\maketitle

\begin{abstract}
Loop closure detection is an essential component of \ac{SLAM} systems, which reduces the drift accumulated over time. Over the years, several deep learning approaches have been proposed to address this task, however their performance has been subpar compared to handcrafted techniques, especially while dealing with reverse loops. In this paper, we introduce the novel \net\ that effectively detects loop closures in LiDAR point clouds by simultaneously identifying previously visited places and estimating the 6-DoF relative transformation between the current scan and the map. \net\ is composed of a shared encoder, a place recognition head that extracts global descriptors, and a relative pose head that estimates the transformation between two point clouds. We introduce a novel relative pose head based on the unbalanced optimal transport theory that we implement in a differentiable manner to allow for end-to-end training. Extensive evaluations of \net\ on multiple real-world autonomous driving datasets show that our approach outperforms state-of-the-art loop closure detection and point cloud registration techniques by a large margin, especially while dealing with reverse loops. Moreover, we integrate our proposed loop closure detection approach into a LiDAR SLAM library to provide a complete mapping system and demonstrate the generalization ability using different sensor setup in an unseen city.\looseness=-1
  
\end{abstract}


\section{Introduction}

\begin{figure}
\footnotesize
\centering
\includegraphics[width=\linewidth]{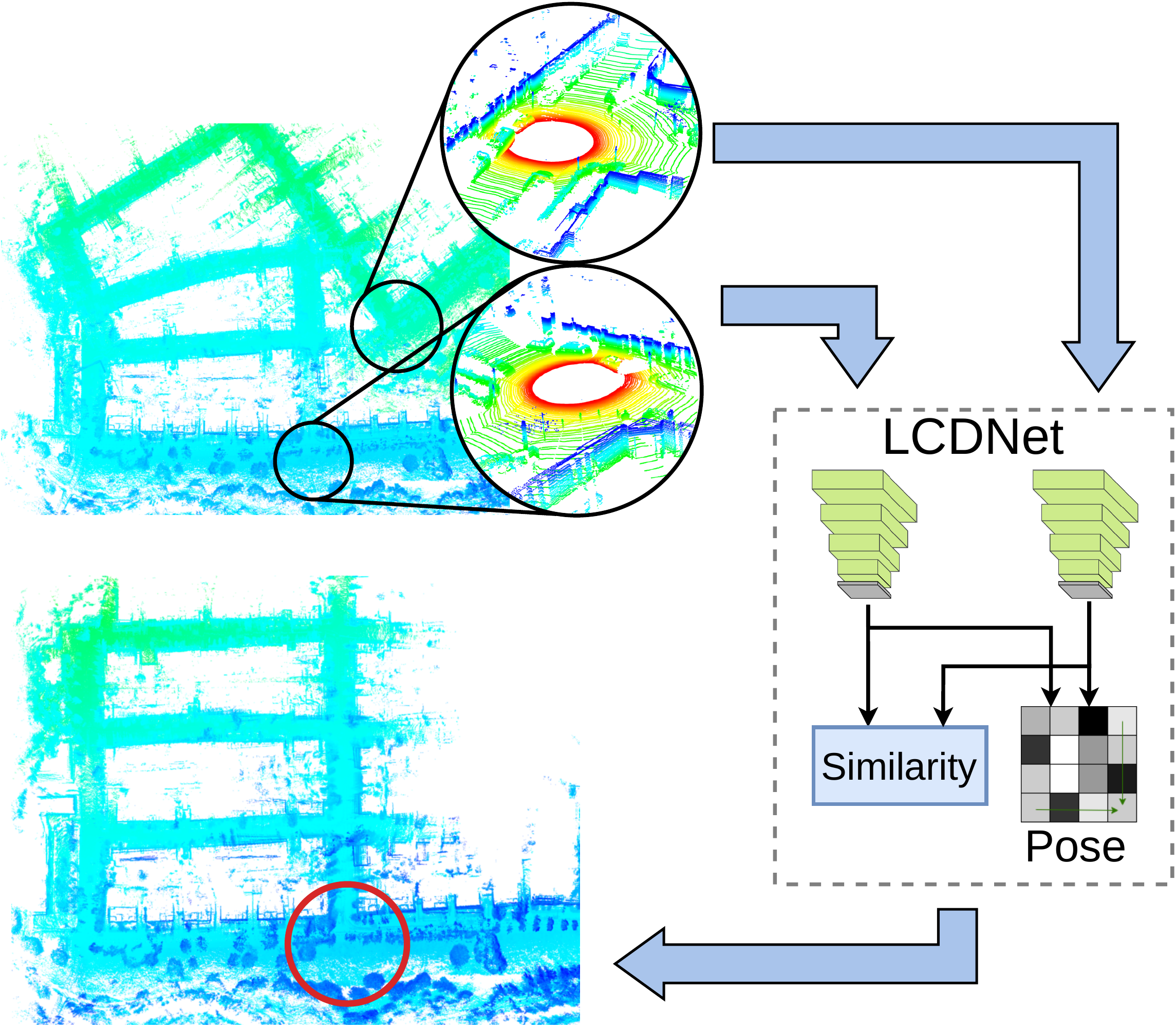}
\caption{Our proposed \net\ detects loops by computing the similarity between two point clouds and predicting the relative pose between them. This is a crucial component of any SLAM system, as it reduces the drift accumulated over time.}
\label{fig:coverfig}
\vspace{-3mm}
\end{figure}

\IEEEPARstart{S}{imultaneous} Localization and Mapping (SLAM) is an essential task for autonomous mobile robots as it is a critical precursor for other tasks in the navigation pipeline. A failure in the \ac{SLAM} system will adversely affect all the subsequent tasks and negatively impact the functioning of the robot. Therefore, improving the robustness of \ac{SLAM} systems has garnered significant interest from both industry and academia in the past decades, as demonstrated by the widespread adoption in many fields such as self-driving cars~\cite{cattaneo2020cmrnetpp}, unmanned aerial vehicles~\cite{mittal2019vision}, agricultural robots~\cite{tanke2012automation}, and autonomous marine vehicles~\cite{valada2014development}. The goal of any \ac{SLAM} system is to build a map of an unknown environment by exploiting onboard sensor data (such as \ac{GPS}, cameras, \ac{LIDAR}, and \acp{IMU}), and at the same time localize the robot within the built map.

A typical \ac{SLAM} pipeline consists of three main components: (i) consecutive scan alignment in which subsequent scans are aligned by leveraging information such as odometry, scan matching, or \acp{IMU} information; (ii) loop detection to identify places that were previously visited, and (iii) loop closure to align the current scan to the previously visited place and accordingly correct the map. Since the first step by itself usually drifts over time due to its incremental nature, the map will no longer be consistent when the robot navigates through a place that was previously visited. Therefore, steps (ii) and (iii) are employed to reduce the accumulated drift by adding a new constraint to the pose graph when a loop is detected. Subsequently, all the previous poses are corrected according to this constraint, thus generating a consistent map.

Although several vision-based \ac{SLAM} systems have been proposed~\cite{artal2015, cummis2008}, their loop closure methods often fail in case of strong variation due to illumination, appearance, or viewpoint changes. LIDARs on the other hand, are invariant to illumination changes and provide an accurate geometric reconstruction of the surrounding environment. Hence, they are often preferred over cameras for \ac{SLAM} approaches due to their inherent robustness. Standard \ac{LIDAR}-based loop detection methods extract local keypoints~\cite{steder2011, bosse2013} or global handcrafted descriptors~\cite{He2016, Kim2018}, and compare the descriptor of the current scan with that of previous scans to identify loops. However, most of these approaches require an ad hoc function to compare the descriptors of two point clouds, which drastically impacts the runtime when the number of past scans increases.

Driven by the significant strides achieved by \acp{DNN} in different fields, many recent works~\cite{Uy_2018_CVPR, schaupp2019, chen2020rss, kong2020semantic, zhu2020} employ \acp{DNN} to address the loop detection task in LiDAR-based \ac{SLAM} systems. While these approaches are generally faster than handcrafted methods, their performance is not on par with these state-of-the-art methods, especially in the case of reverse loops. Another challenge faced while detecting a loop closure is related to aligning the current point cloud with the built map. A common approach is to leverage standard techniques for scan matching such as the \ac{ICP}~\cite{zhang1994iterative} algorithm or one of its variants~\cite{ segal2009generalized, yang2015go, Agamennoni2016_IROS}. Although \ac{ICP} is generally able to successfully align two point clouds when they are relatively similar and close, it can fall in local minima when the initial pose between the point clouds is very different.
This is often the case when faced with reverse direction loops. To overcome this problem, some methods also provide an estimate of the rotation between the two point clouds. This estimate can then be used as an initial guess in the \ac{ICP} algorithm to aid the alignment to converge to the correct solution.

Several recent works~\cite{Aoki_2019_CVPR, yew2020,cao21pcam,wang2019deep} have been proposed to address the point cloud registration task by leveraging the advancement in deep learning. Although these approaches achieve impressive results in registering single objects and outperform standard techniques, the protocol used to test these approaches only consider a relatively small initial rotation misalignment (up to 45$^\circ$). However, the point clouds can be rotated by 180$^\circ$ in the loop closure task. Recent work has shown that some of these methods have a very low success rate when the initial misalignment is larger than 120$^\circ$~\cite{Aoki_2019_CVPR}.

In this paper, we propose the novel \net\ for loop closure detection which performs both loop detection and point cloud registration (see~\Cref{fig:coverfig}). Our method combines the ability of \acp{DNN} to extract distinctive features from point clouds, with algorithms from the transport theory for feature matching. \net\ is composed of a shared backbone that extracts point features, followed by the place recognition head that extracts global descriptors and the relative pose head that estimates the transformation between two point clouds. One of the core components of our \net\ is the Unbalanced Optimal Transport (UOT) algorithm that we implement in a differentiable manner. UOT allows us to effectively match the features extracted from the two point clouds, reject outliers, and handle occluded points, while still being able to train the network in an end-to-end manner. As opposed to existing loop closure detection methods that estimate the relative yaw rotation between two point clouds, our proposed \net\ estimates the full 6-DoF relative transformation {under driving conditions} between them which significantly helps the subsequent \ac{ICP} refinement to converge faster.

We train our proposed \net\ on sequences from the KITTI odometry~\cite{Geiger2012CVPR} and KITTI-360~\cite{Xie2016CVPR} datasets, and evaluate it on the unseen sequences on both datasets. Moreover, we found that there is a lack of a standard protocol for evaluating loop closure detection methods in the existing literature. Different works evaluated their approaches using different metrics such as precision-recall curve, average precision, Receiver Operating Characteristic (ROC) curve, recall@k, and maximum F1-score. Even among the methods that use the same metric for evaluation, there are still substantial differences in the other parameters chosen for computing the metrics which makes the performance of existing methods not directly comparable. For example, the definition of a true loop can span from scans within three meters~\cite{kong2020semantic} up to scans within 15~meters~\cite{zhu2020}. Therefore, in this work, we evaluate existing state-of-the-art approaches using a uniform evaluation protocol to provide a fair comparison. Exhaustive comparisons demonstrate that our proposed \net\ outperforms both handcrafted methods as well as \ac{DNN}-based methods and achieves state-of-the-art performance on both loop closure detection and point cloud registration tasks. Furthermore, we present detailed ablation studies on the architectural topology of \net\ and also present results from integrating \net\ into a recent LiDAR \ac{SLAM} library~\cite{liosam2020shan}. Additionally, we demonstrate the generalization ability of our proposed approach using experiments with a different sensor setup from an autonomous driving scenario in a completely different city.

The main contributions of this work can be summarized as follows:
{
\begin{enumerate}[topsep=0pt]
  \item We propose \net, a novel approach for loop closure detection that effectively detects reverse loops.
  \item We propose an end-to-end trainable relative pose regression network based on the unbalanced optimal transport theory that can register two point clouds that only partially overlap and with an arbitrary initial misalignment.
  \item We comprehensively evaluate existing state-of-the-art loop closure detection methods using a uniform evaluation protocol, we perform extensive evaluations of \net\ on multiple autonomous driving datasets, and we present detailed ablation studies that demonstrate the efficacy of our contributions.
  \item We study the generalization ability of our approach to unseen environments and different sensor setups by evaluating \net\ on our own recorded dataset around the city of Freiburg, Germany.
  \item We integrate our network into a \ac{SLAM} library to provide a complete system for localization and mapping and we make the code, the entire \ac{SLAM} system, and the evaluation tools publicly available at \url{http://rl.uni-freiburg.de/research/lidar-slam-lc}.
\end{enumerate}
}
The remainder of the paper is organized as follows: we review existing methods that are related to our approach in \Cref{sec:related}. In \Cref{sec:technical}, we detail our proposed \net\ and the integration into the \ac{SLAM} system. We then present experiments that demonstrate the effectiveness and robustness of \net\ in \Cref{sec:experiments}. Finally, we present our conclusions in \Cref{sec:conclusions}.\looseness=-1

\section{Related Works}
\label{sec:related}

In this section, we provide an overview of the state-of-the-art techniques for vision-based and LiDAR-based loop closure detection, followed by methods for point cloud registration.

{\parskip=5pt
\noindent\textit{Loop Closure Detection}: Techniques for loop closure detection can primarily be categorized into visual and LiDAR-based methods.
Traditionally, vision-based techniques for loop closure detection rely on handcrafted features for identifying and representing relevant parts of scenes depicted within images, and exploit a Bag-of-Words model to combine them \cite{artal2015, cummis2008}. In the last few years, deep learning approaches~\cite{Arandjelovic_2016_CVPR, zhang_2017_ICAC} have been proposed that achieve successful results. These techniques employ \ac{DNN} for computing global descriptors to provide a compact representation of images and perform direct comparisons between descriptors for searching matches between similar places. Recently, \cite{liu2019loop} proposed a novel approach that employs \acp{DNN} for extracting local features from intermediate layers and organizes them in a word-pairs model. Although vision-based methods achieve impressive performance, they are not robust against adverse environmental situations such as challenging light conditions and appearance variations that can arise during long-term navigation. As loop closure detection is a critical task within \ac{SLAM} systems, in this work, we exploit LiDARs for the sensing modality since they provide more reliable information even in challenging conditions in which visual systems fail.}

3D LiDAR-based techniques have gained significant interest in the last decade, as LiDARs provide rich 3D information of the environment with high accuracy and their performance is not affected by illumination changes. Similar to vision-based approaches, LiDAR-based techniques also exploit local features. Most methods use 3D keypoints \cite{3dsurf2010, Zhong2009} that are organized in a bag-of-words model for matching point clouds \cite{steder2011}. \cite{bosse2013} propose a keypoint based approach in which a nearest neighbor voting paradigm is employed to determine if a set of keypoints represent a previously visited location. Recently, \cite{sivavoxel2020} propose a voxel-based method that divides a 3D scan into voxels and extracts multiple features from them through different modalities, followed by learning the importance of voxels and types of features.

Another category of techniques represents point clouds through global descriptors. \cite{He2016} propose an approach that directly produces point clouds fingerprints. In particular, this method relies on density signatures extracted from multiple projections of 3D point clouds on different 2D planes. \cite{Kim2018} introduces a novel global descriptor called Scan Context that exploits bird-eye-view representation of a point cloud together with a space partitioning procedure to encode the 2.5D information within an image.
In a similar approach, \cite{IRIS_2020} propose a method to extract binary signature images from 3D point clouds by employing LoG-Gabor filtering with thresholding operations to obtain a descriptor. The main drawback of these approaches is that they require an ad-hoc function to compare the global descriptor of two point clouds which drastically impacts the runtime when the number of scans to compare increases.

Recently, \ac{DNN}-based techniques have also been proposed for computing descriptors from 3D point clouds. \cite{Uy_2018_CVPR} propose PointNetVLAD which is composed of PointNet~\cite{Qi_2017_CVPR} with a NetVLAD layer~\cite{Arandjelovic_2016_CVPR} and yields compact descriptors. \cite{schaupp2019} propose OREOS which computes 2D projection of point clouds on cylindrical planes and is subsequently fed into a \ac{DNN} that computes global descriptors and estimate their yaw discrepancy.
More recently, the OverlapNet~\cite{chen2020rss} architecture was introduced, which estimates the overlap and relative yaw angle between a pair of point clouds. The overlap estimate is then used for detecting loop closures while the yaw angle estimation is provided to the Iterative Closest Point (ICP) algorithm as the initial guess for the point clouds alignment. While \ac{DNN}-based methods are generally faster than classical techniques, and show promising results in sequences that contain loops only in the same direction, their performance drastically decreases when they are faced with reverse loops.

Recently, techniques that exploit graph structures by matching semantic graphs have been proposed~\cite{kong2020semantic, zhu2020}. These approaches first extract semantic information and perform instance retrieval, followed by defining graph vertices on the object centroids. Subsequently, features are extracted by considering handcrafted descriptors or by processing nodes through a Dynamic Graph CNN~\cite{dgccn2019}. Finally, loop closures are identified by comparing vertices between graphs. However, computing the exact correspondences between two graphs is still an open problem and existing methods are only suitable when a few vertices are considered or they can only provide an approximated solution~\cite{Bailey2000}. In this work, we exploit recent advancements in deep learning and propose a \ac{DNN}-based approach for detecting loop closure by combining high-level voxel features with fine-grained point features. Our approach effectively detects loops in challenging scenarios such as reverse loops and outperforms state-of-the-art handcrafted and learning-based techniques.

\begin{figure*}
  \centering
  \includegraphics[width=0.9\linewidth]{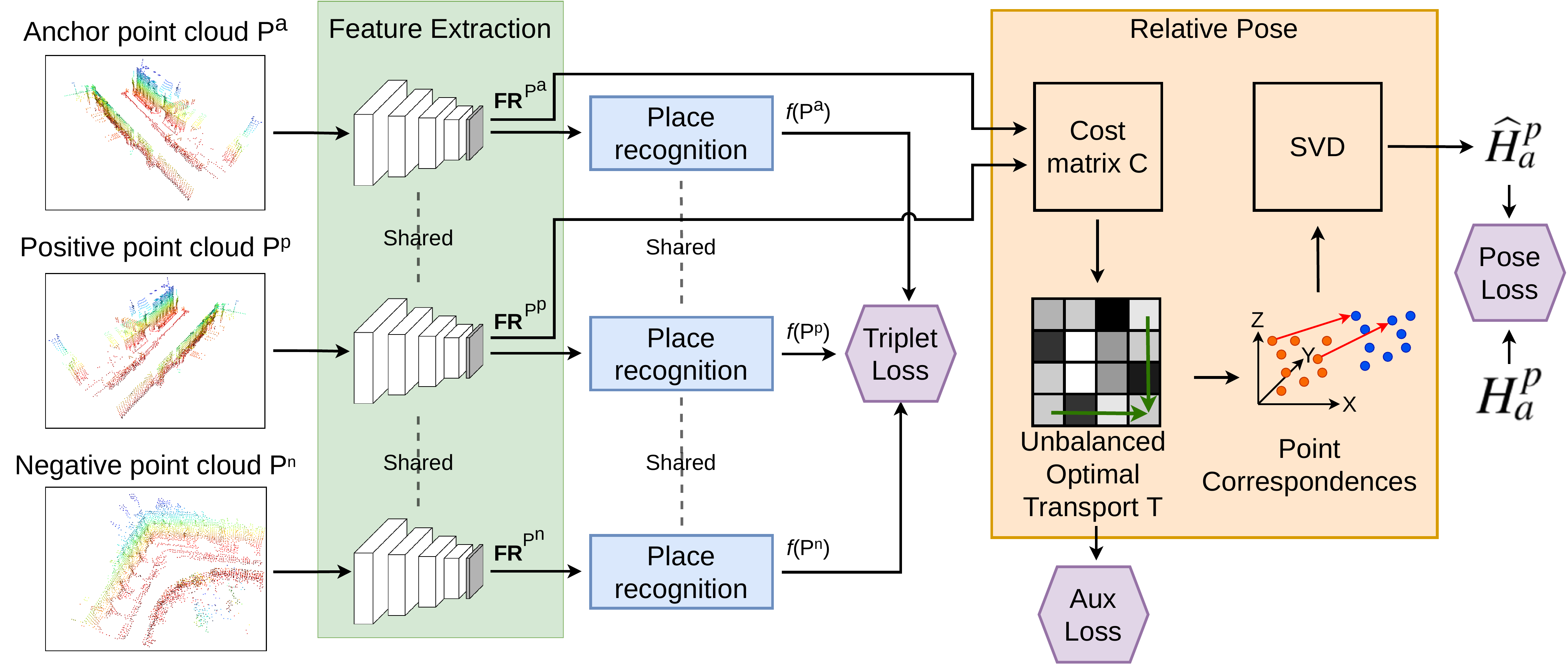}
  \caption{Overview of our proposed \net\ which is composed of a shared feature extractor ({\color{pvgreen}green}), a place recognition head ({\color{pvblue}blue}) that generates global descriptors, and a relative pose head ({\color{pvorange}orange}) that estimate the transformation between two point clouds. We use three loss functions to train \net\ (triplet loss, aux loss, and pose loss) which are depicted in {\color{pvpurple}purple}. The topology of the feature extractor is further illustrated in \Cref{fig:pvrcnn}.}
  \label{fig:overview}
  \vspace{-3mm}
\end{figure*}

{\parskip=5pt
\noindent\textit{Point Cloud Registration}: Point clouds registration represents the task of finding a rigid transformation to accurately align a pair of point clouds. The \ac{ICP} algorithm~\cite{zhang1994iterative} is one standard method that is often employed to tackle this task.
 Although \ac{ICP} is one of the most popular methods, the main drawback concern the initial rough alignment of point clouds which is required to reach an acceptable solution, and the algorithm complexity which increases drastically with the number of points. Other methods tackle the registration problem globally without requiring a rough initial alignment. Traditionally, these techniques exploit local features~\cite{Rusu_ICRA_2009} for finding matches between point clouds and employ algorithms such as \ac{RANSAC}~\cite{fischler1981random} for estimating the final transformation. However, the presence of noise in the input data and outliers generated from incorrect matches can lead to an inaccurate result. To address these problems, \cite{Zhou_ECCV_2016} proposes a global registration approach that ensures fast and accurate alignment, even in the presence of many outliers.}

{Recent years have also seen the introduction of deep learning methods that tackle the registration problem. A typical approach is to employ a \ac{DNN} for extracting features which are then used in the later stages to perform point clouds alignment.
\cite{Aoki_2019_CVPR} propose such an approach known as PointNetLK, which exploits the PointNet~\cite{Qi_2017_CVPR} architecture for feature extraction and employs a variation of the Lucas and Kanade algorithm~\cite{lucasKanade} to perform registration.
Deep Closest Point (DCP)~\cite{wang2019deep} is another approach that employs a Siamese architecture, attention modules, and differentiable \ac{SVD} to regress a rigid transform for aligning two input point clouds.
Recently, \cite{yew2020} proposes a \ac{DNN}-based method called RPM-Net which is inspired by Robust Point Matching (RPM). RPM-Net employs two different neural networks to extract features and predict annealing parameters that are required for RPM. However, these methods are only capable of aligning point clouds that are relatively close to each other (up to 45$^\circ$ rotation misalignment), and completely fail to register point clouds that are more than 120$^\circ$ apart~\cite{Aoki_2019_CVPR}.}
In contrast to the aforementioned methods, the approach that we propose in this work does not require any initial guess as input and can handle both outliers and occluded points. Moreover, unlike existing \ac{DNN}-based methods, our approach effectively aligns point clouds with arbitrary initial rotation misalignment.


\section{Technical Approach}
\label{sec:technical}
In this section, we detail our proposed \net\ for loop closure detection and point cloud registration from \ac{LIDAR} point clouds. An overview of the proposed approach is depicted in \Cref{fig:overview}. The network consists of three main components: feature extraction, global descriptor head, and 6-DoF relative pose estimation head. We first describe each of the aforementioned components and the associated loss functions for training, followed by the approach for integrating \net\ into the \ac{SLAM} system.

\subsection{Feature Extraction}
\label{sec:feature_extractor}

\begin{figure}
  \centering
  \includegraphics[width=\linewidth]{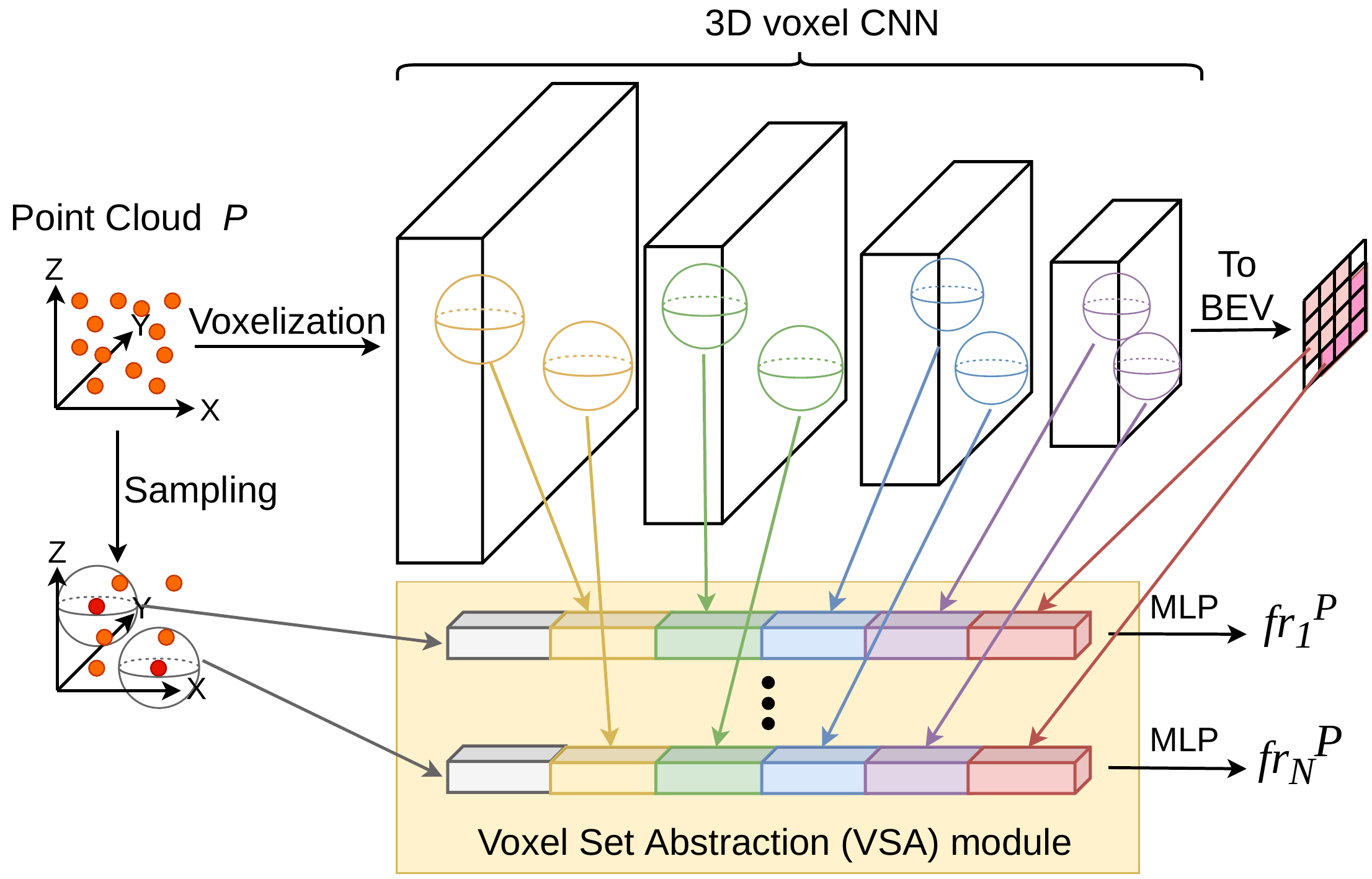}
  \caption{Network topology of the PV-RCNN architecture that we build upon for feature extractor component of our proposed LCDNet.
  }
  \label{fig:pvrcnn}
\vspace{-3mm}
\end{figure}

We build the feature extractor stream of our network based upon the PV-RCNN~\cite{shi2020pv} architecture that was proposed for 3D object detection. PV-RCNN effectively combines the ability of voxel-based methods for extracting high-level features, with fine-grained features provided by PointNet-type architectures. We make several changes to the standard architecture to adapt it to our task. We illustrate the topology of our adapted PV-RCNN in \Cref{fig:pvrcnn}.

The input to the network is a point cloud $P \in \mathbb{R}^{Jx4}$ ($J$ points with 4 values each: x, y, z, and intensity). The output of our feature extractor network is a set of $N$ keypoints' feature $\mathbf{FR}^P = \{fr_{1}^P, \ \dots\ , fr_N^P\}$, where $fr_{i}^P \in \mathbb{R}^D$ is the D-dimensional feature vector for the $i$-th keypoint. Since we are interested in the feature extraction, and not in the object detection head, we only use the 3D voxel \ac{DNN} and the Voxel Set Abstraction (VSA) module, and we discarded the region proposal network, the ROI-grid pooling, and the fully connected layers towards the end of the architecture. The 3D voxel \ac{DNN} first converts the point cloud into a voxel grid of size $L \times W \times H$, where voxel features are averaged across all the points that lay within the same voxel. Subsequently, we extract a feature pyramid using sparse 3D convolutions and downsampling. In particular, we use four pyramid blocks composed of 3D sparse convolutions, with downsampling rates of $1 \times$, $2 \times$, $4 \times$, and $8 \times$, respectively. Finally, we convert the coarsest feature map into a 2D Bird's-Eye-View (BEV) feature map by stacking the features along the $Z$ axis.

The VSA module, on the other hand, aggregates all the pyramid feature maps together with the BEV feature map and the input point cloud into a small set of $N$ keypoints features. To do so, we first downsample the point cloud using the Farthest Point Sampling (FPS) algorithm~\cite{gonzalez1985clustering} to select $N$ uniformly distributed keypoints. The VSA module is an extension of the Set Abstraction (SA) level~\cite{qi_NIPS_2017}. The standard SA aggregate neighbors point features in the raw point cloud, whereas, the VSA aggregate neighbors voxel features in the 3D sparse feature map. For every selected keypoint $kp_i$, and every layer $l$ of the pyramid feature map, the keypoint features $f_i^l$ are computed as
{
\begin{equation}
  f_i^l = MP(MLP(\mathcal{M}(S_i^l))),
\end{equation}}
where $MP$ is the max-pooling operation, MLP denotes a \ac{MLP}, and $\mathcal{M}$ randomly samples the set of neighbor voxel features $S_i^l$, which is computed as
\begin{equation}
  \label{eq:voxelfeature}
  S_i^l = \left\{ \left[ fvox_j^l ; v_j^l - kp_i\right] ; \text{s. t.} \left\lVert v_j^l - kp_i \right\rVert^2 < r \right\},
\end{equation}
where $fvox_j^l$ is the feature of the voxel $j$ at level $l$, $v_j^l$ denotes the coordinates of the voxel $j$ at level $l$, and $r$ is the neighbor radius. This operation is performed at every level of the pyramid to yield
\begin{equation}
  f_i^{pv} = \left[ f_i^1, f_i^2, f_i^3, f_i^4 \right].
\end{equation}
We perform a similar operation for the input raw point cloud, as well as the BEV feature map, yielding the aggregated keypoint features
\begin{equation}
  f_i^{3D} = \left[ f_i^{pv}, f_i^{raw}, f_i^{bev} \right].
\end{equation}
Lastly, we employ a \ac{MLP} on the aggregated keypoint features to generate the final keypoint feature vectors as
\begin{equation}
  \label{eq:fri}
  fr_i = MLP(f_i^{3D}).  
\end{equation}

As opposed to the original PV-RCNN that processes only the points that lay in the camera Field Of View (FOV), we require the full \ang{360} surrounding view. Therefore, we use a voxel grid size of $\pm\, \SI{70.4}{\meter}$, $\pm\, \SI{70.4}{\meter}$ and $[\SI{-1}{\meter}, \SI{3}{\meter}]$ in the $x, y$ and $z$ dimensions, respectively. We use a voxel size of $\SI{0.1}{\meter} \times \SI{0.1}{\meter} \times \SI{0.1}{\meter}$.
We demonstrate the ability of our feature extractor in generating discriminative keypoint features by comparing it with different state-of-the-art backbones in the ablation studies presented in \Cref{sec:ablation}. Moreover, we also investigate the best choice for the dimensionality $D$ of the keypoint features.

\subsection{Global Descriptor}

In order to generate a global descriptor for a given point cloud, we aggregate the keypoints' feature set $\mathbf{FR}^P$ obtained from the feature extractor into a compact G-dimensional vector. To do so, we first employ the NetVLAD layer~\cite{Arandjelovic_2016_CVPR} which converts the (N x D)-dimensional $\mathbf{FR}^P$ set into a (K x D)-dimensional vector $\mathbf{V}(\mathbf{FR}^P)$ by learning a set of $K$ cluster centers $\{c_1,\ \dots\ , c_K\}$, $c_k \in \mathbb{R}^D$. NetVLAD mimics the original \ac{VLAD}~\cite{jegou2010} using differentiable operations. It replaces the k-means clustering with learnable clusters and replacing the hard assignment with a soft assignment defined as
\begin{equation}
  \label{eq:soft_ass}
  a_k(fr_i^P) = \frac{e^{\mathbf{w}_k^\top fr_i^P + b_k}}{\sum_{k'=1}^K e^{\mathbf{w}_{k'}^\top fr_i^P + b_{k'}}},
\end{equation}
where $\mathbf{w}_k \in \mathbb{R}^D$ and $b_k \in \mathbb{R}$ are the learnable weights and bias. In practice, $a_k(fr_i^P)$ represents the probability of assigning the feature vector $fr_i^P$ to the cluster center $c_k$. The final NetVLAD descriptor $\mathbf{V}(\mathbf{FR}^P) = [ \mathbf{V}_1(\mathbf{FR}^P),\ \dots\, \mathbf{V}_K(\mathbf{FR}^P)]$ is computed by combining the original \ac{VLAD} formulation with the soft assignment defined in~\cref{eq:soft_ass} as
\begin{equation}
  \mathbf{V}_k(\mathbf{FR}^P) = \sum_{i=1}^{N} a_k(fr_i^P) (fr_i^P - c_k).
\end{equation}

We use the NetVLAD layer instead of max-pooling employed in PointNet~\cite{Qi_2017_CVPR}, as it has demonstrated superior performance for point cloud retrieval~\cite{Uy_2018_CVPR}. To further reduce the dimensionality of the final global descriptor, we employ a simple \ac{MLP} that compresses the (K $\times$ D)-dimensional vector $\mathbf{V}(\mathbf{FR}^P)$ into a G-dimensional compact descriptor. We then obtain the final global descriptor $\globaldesc \in \mathbb{R}^G$ by employing the Context Gating (CG) module~\cite{miech17loupe} on the output of the \ac{MLP}.
The CG module re-weights the output of the \ac{MLP} using a self-attention mechanism as
\begin{equation}
  Y(X) = \sigma(WX + b) \odot X,
\end{equation}
where $X$ is the \ac{MLP} output, $\sigma$ is the element-wise sigmoid operation, $\odot$ is the element-wise multiplication, $W$ and $b$ are the weights and bias of the \ac{MLP}. The CG module captures dependencies among features by down-weighting or up-weighting features based on the \textit{context} while considering the full set of features as a whole, thus focusing the attention on more discriminative features.

\subsection{Relative Pose Estimation}
\label{sec:posehead}

Given two point clouds $P$ and $S$, the third component of our architecture estimates the 6-DoF transformation to align the source point cloud $P$ with the target point cloud $S$ {under driving conditions}. We perform this task by matching the keypoints' features $\mathbf{FR}^P$ and $\mathbf{FR}^S$ computed using our feature extractor from \cref{sec:feature_extractor}. Due to the sparse nature of \ac{LIDAR} point clouds and the keypoint sampling step which is performed in the feature extractor, a point in $P$ might not have a single matching point in $S$, but it can lay in between two or more points in $S$. Therefore, a one-to-one mapping is not desirable in our task.

In order to address this problem, we employ the Sinkhorn algorithm~\cite{sinkhorn1964relationship}, which can be used to approximate the \textit{optimal transport} (OT) theory in a fast, highly parallelizable and differentiable manner. Recent work has shown benefits of using the Sinkhorn algorithm with \acp{DNN} for several tasks such as feature matching~\cite{sarlin2020}, scene flow~\cite{puy2020b}, shape correspondence~\cite{eisenberger2020}, and style transfer~\cite{kolkin2019}. The discrete Kantorovich formulation of the optimal transport is defined as
\begin{equation}
  \label{eq:ot}
\begin{split}
  T = \argmin_{A \in \mathbb{R}^{N \times N}} \Bigg\{ \sum_{i, j} C_{ij} A_{ij};\ \text{s. t. A is doubly stochastic} \Bigg\},
\end{split}
\end{equation}
where $C_{ij}$ is the cost of matching the $i$-th point in $P$ to the $j$-th point in $S$.
In order to employ the Sinkhorn algorithm, we add an entropic regularization term:
\begin{equation}
  \label{eq:otregularized}
  T = \argmin_{A \in \mathbb{R}^{N \times N}} \Bigg\{ \sum_{i, j} C_{ij} A_{ij} + \lambda A_{ij} \left( \log A_{ij} - 1 \right) \Bigg\},
\end{equation}
where $\lambda$ is a parameter that controls the sparseness of the mapping (as $\lambda \rightarrow 0$, $T$ converges to a one-to-one mapping). However, both \cref{eq:ot,eq:otregularized} are subject to $A$ being a doubly stochastic matrix (mass preservation constraint), \textit{i.e.,} every point in $\mathbf{FR}^P$ has to be matched to one or more points in $\mathbf{FR}^S$, and vice versa. In our point cloud matching task, some points in $\mathbf{FR}^P$ might not have a matching in $\mathbf{FR}^S$, for example when a car is present in one point cloud but is absent in the other, or in the case of occlusions. Therefore, we need to relax the mass prevention constraint. One common approach to overcome this problem is by adding a dummy point in both $P$ and $S$ (\textit{i.e.,} add a dummy row and column to $A$). Another way is to reformulate the problem as \textit{unbalanced optimal transport} (UOT) which allows mass creation and destruction, and is defined as
\begin{equation}\small
  \label{eq:otunbalanced}
  \begin{split}
  T = &\argmin_{A \in \mathbb{R}^{N \times N}} \Bigg\{ \left( \sum_{i, j} C_{ij} A_{ij} + \lambda A_{ij} \left( \log A_{ij} - 1 \right) \right) +  \\
  & \rho \left( KL \left( \sum_{i} A_{ij}|U(1,N) \right) + KL \left( \sum_{j} A_{ij}|U(1,N) \right) \right) \Bigg\},
\end{split}
\end{equation}
where KL is the Kullback--Leibler divergence, $U$ is the discrete uniform distribution, and $\rho$ is a parameter that controls how much mass is preserved.
{The UOT formulation, compared to the standard OT, reduces the negative effect caused by incorrect point matching and is more robust to the stochasticity induced by keypoint sampling~\cite{fatras2021unbalanced}.}
A recent extension to the Sinkhorn algorithm \cite{chizat2018scaling} that approximates the unbalanced optimal transport is shown in~\algref{alg:ot}. We set the cost matrix $C$ as cosine distance between the keypoints' features $C_{ij} = 1 - \nicefrac{FR_i^P \cdot FR_j^S}{\left\lVert FR_i^P \right\rVert \left\lVert FR_j^S \right\rVert}$. Instead of setting $\lambda$ and $\rho$ manually, we learn them using back propagation.

\begin{algorithm}[t]
  \SetAlgoLined
  \DontPrintSemicolon
  \KwData{Cost matrix $C$, number of iterations $L$, parameters $\lambda$ and $\rho$}
  \KwResult{Unbalanced Optimal Transport $T$}
  \Begin{
    $K \leftarrow e^{-C / \lambda}$\;
    $a \leftarrow \mathbbm{1}_N / N$\;
    $b \leftarrow \mathbbm{1}_N / N$\;
    $v \leftarrow \mathbbm{1}_N / N$\;
    \For{$i \leftarrow 1$ \KwTo $L$}{
      $u \leftarrow [a \oslash (Kv)]^{\rho/(\rho+\lambda)}$\;
      $v \leftarrow [b \oslash (K^\intercal u)]^{\rho/(\rho+\lambda)}$\;
    }
    $T \leftarrow u \odot K \odot v^\intercal$\;
  }
  \caption{Unbalanced Optimal Transport}
  \label{alg:ot}
  \algorithmfootnote{where $\oslash$ is the element wise division, and $\odot$ is the element-wise multiplication.}
\end{algorithm}

Once we estimate the unbalanced optimal transport $T$, which represents the set of soft correspondence between keypoints' features $\mathbf{FR}^P$ and $\mathbf{FR}^S$, together with their respective 3D keypoints' coordinates $P$ and $S$, we compute for every keypoint $p_j \in P$ its projected coordinates in $S$ as 
\begin{equation}
    \hat{s}_j = \frac{\sum_{k=1}^K T_{jk} s_k}{\sum_{k=1}^K T_{jk}}.
\end{equation}

Finally, to estimate the rigid body transformation between the original point cloud $P$ and its projection $\hat{S}$ in $S$ we use the weighted \ac{SVD}. Since both~\algref{alg:ot} and \ac{SVD} are differentiable, we train our relative pose head in an end-to-end manner by comparing the predicted transformation $\widehat{H}_{P}^{S}$ with the groundtruth transformation $H_{P}^{S}$.

Once the network has been trained, we replace the UOT-based relative position head with a \ac{RANSAC}-based registration method that exploits the features extracted by our network to find correspondences. In this way, we can train the network in an end-to-end manner, and at the same time estimate accurate relative poses using the robust \ac{RANSAC} estimator during inference.

\subsection{Loss Function}

We train our global descriptors using the triplet loss~\cite{facenet}. Given an anchor point cloud $P^a$, a positive sample $P^p$ (point cloud of the same place), and a negative sample $P^n$ (point cloud of a different place), the triplet loss enforce the distance between the descriptors of positive samples to be smaller than the distance between negative samples descriptors. More formally, the triplet loss is defined as
\begin{equation}
  \label{eq:triplet_loss}
  \mathcal{L}_{trp} = [d(f(P^a), f(P^p)) - d(f(P^a), f(P^n)) + m)]_+,
\end{equation}
where $d(\cdot)$ is a distance function, $m$ is the desired separation margin, and $[x]_+$ means $max(0, x)$.

Instead of selecting the triplets in advance (offline mining) for every anchor in the batch, we randomly select a positive sample, and we select the negative sample randomly from all the samples in the batch that depict a different place (online negative mining). We compute the relative pose transformation only for positive pairs, and we train the model by comparing the anchor point cloud $P^a = \{p^a_1,\ \ldots\ , p^a_J\}$ transformed using the predicted transformation $\widehat{H}_{a}^{p}$ and the groundtruth transformation $H_{a}^{p}$ as
\begin{equation}
  \label{eq:pose_loss}
  \mathcal{L}_{pose} = \frac{1}{J} \sum_{j=1}^J \left\lvert \widehat{H}_{a}^{p} p^a_j - H_{a}^{p} p^a_j \right\rvert .
\end{equation}

We add an auxiliary loss on the matches estimated by the unbalanced optimal transport $T$ as
\begin{equation}
  \label{eq:aux_loss}
  \mathcal{L}_{OT} = \frac{1}{J} \sum_{j=1}^J \left\lvert \frac{\sum_{k=1}^K T_{jk} p^p_k}{\sum_{k=1}^K T_{jk}} - H_{a}^{p} p^a_j \right\rvert .
\end{equation}

The final loss function is a linear combination of the three aforementioned components:
\begin{equation}
  \label{eq:total_loss}
  \mathcal{L}_{total} = \mathcal{L}_{trp} + \mathcal{L}_{pose} + \beta \mathcal{L}_{OT},
\end{equation}
{ where $\beta$ is loss balancing term which we empirically set to 0.05. Consequently, due to the combination of triplet loss, UOT, and data augmentation, the shared feature extractor learns to yield distinctive, rotation and translation invariant keypoints' features through backpropagation.}

\subsection{SLAM System}
\label{sec:slam}

We integrate our proposed \net\ into a recently proposed \ac{SLAM} system, namely LIO-SAM~\cite{liosam2020shan} which achieves state-of-the-art performance on large-scale outdoor environments. LIO-SAM is a tightly coupled LiDAR inertial odometry framework built atop a factor graph. The framework takes a LiDAR point cloud and \ac{IMU} measurements as input. It includes four types of constraints that are added to the factor graph: \ac{IMU} preintegration, LiDAR odometry, \ac{GPS} measurements (optional), and loop closure. In order to reduce the computational complexity, LIO-SAM selectively chooses LiDAR scans as keyframes only when the robot moves more than a predefined threshold since the last saved keyframe. The scans in between two keyframes are then discarded. We replaced the Euclidean distance-based loop closure detection provided in LIO-SAM with our \net. From a technical perspective, for every keyframe $\mathbb{F}_i$ added to the LIO-SAM factor graph, we compute and store its global descriptor $f(\mathbb{F}_i)$ in a database. When a new keyframe $\mathbb{F}_{i+1}$ is added to the graph, we retrieve the point cloud with the most similar descriptor (excluding the past M keyframes) from the database:
\begin{equation}
  W = \argmin_{j \in \{1, \ldots, i-M\}} \left\lVert f(\mathbb{F}_{i+1}) - f(\mathbb{F}_j) \right\rVert .
\end{equation}

{If the distance between the two descriptors is below a certain threshold $th$, we set $\mathbb{F}_W$ as a loop candidate, and we estimate the 6-DoF transformation between the two point clouds $\widehat{H}_{i+1}^{W}$ provided by the relative pose head as described in~\cref{sec:posehead}. Finally, we further refine the transformation using \ac{ICP} with $\widehat{H}_{i+1}^{W}$ as initial guess, and we add the loop closure factor to the pose graph only if the ICP fitness score is higher than a threshold $th_{icp}$. By using this additional geometric consistency check, we can discard the few remaining false positive detection. It is important to note that no \ac{IMU} nor \ac{GPS} measurements are used in the loop detection step}.


\section{Experimental Evaluation}
\label{sec:experiments}

In this section, we first describe the datasets that we evaluate on, followed by the implementation details and the training protocol that we employ. {We then present quantitative and qualitative results from experiments that are designed to demonstrate that our proposed \net\ can (i) effectively detect loop closures even in challenging condition such as loops in the reverse direction, (ii) align two point clouds without any prior initial guess, (iii) robustly align point clouds that only partly overlap,
(iv) provide an accurate initial guess for further ICP alignment,
(v) integrate with an existing SLAM system to provide a fully featured localization and mapping framework, (vi) generalize to unseen environments.}

\subsection{Datasets}
\label{sec:dataset}

\begin{table*}
  \centering
  \caption{Statistics of evaluation datasets.}
  \label{tab:dataset}
  \setlength\tabcolsep{5.5pt}
  \begin{tabular}{lccccccccccccc}
  \toprule
  & \multicolumn{6}{c}{KITTI} & \multicolumn{6}{c}{KITTI-360} & Freiburg \\ \cmidrule(lr){2-7} \cmidrule(lr){8-13} \cmidrule(lr){14-14}
  & 00 & 05 & 06 & 07 & 08 & 09 & 00 & 02 & 04 & 05 & 06 & 09 & - \\ \midrule
  Num. of scans  & \num{4541} & \num{2761} & \num{1101} & \num{1101} & \num{4071} & \num{1591} & \num{10514} & \num{18235} & \num{11052} & \num{6291} & \num{9186} & \num{13247} & \num{25612} \\
  Num. of loops & \num{790} & \num{492} & \num{69} & \num{97} & 334 & \num{18} & \num{2452} & \num{4690} & \num{2218} & \num{2008} & \num{2433} & \num{4670} & \num{13851} \\
  Num. of pairs & \num{10499} & \num{6534} & \num{2138} & \num{2497} & \num{2960} & \num{252} & \num{24499} & \num{43894} & \num{21165} & \num{20361} & \num{22822} & \num{53858} & $\sim$ 411M \\ 
  Route direction & \textit{Same} & \textit{Same} & \textit{Same} & \textit{Same} & \textit{Reverse} & \textit{Same} & \textit{Both} & \textit{Both} & \textit{Both} & \textit{Both} & \textit{Both} & \textit{Both} & \textit{Both} \\
  {\% Reverse Loops} & {3\%} & {5\%} & {0\%} & {0\%} & {100\%} & {0\%} & {67\%} & {87\%} & {92\%} & {88\%} & {61\%} & {46\%} & {20\%} \\
  \bottomrule
  \end{tabular}
  \vspace{-3mm}
\end{table*}

We evaluate our proposed approach on three different autonomous driving datasets. We detail the list of sequences that we use for training and testing, together with the respective number of loop closures and route direction of revisited places in~\cref{tab:dataset}. Note that we do not include the sequences without loops.

{\parskip=5pt
\noindent\textit{KITTI}: The KITTI odometry dataset~\cite{Geiger2012CVPR} contains 11 sequences with \ac{LIDAR} point clouds and groundtruth poses, six of which contain loops. However, the groundtruth for some of these sequences is not aligned to nearby loop closures. Therefore, we use the groundtruth provided with the SemanticKITTI dataset~\cite{behley2019iccv} which is consistent for all the sequences. Most of the KITTI odometry sequences contain loop closures from the same driving direction, except for sequence 08 which contains reverse loop closures. We evaluate our approach on sequences 00 and 08 as they contain the highest number of loops and reverse loops, respectively.}

{\parskip=5pt
\noindent\textit{KITTI-360}: The recently released KITTI-360 dataset~\cite{Xie2016CVPR} consists of nine sequences, six of which contain loops. KITTI-360 contains more loops and reverse loops than the standard KITTI dataset (see~\cref{tab:dataset}). We evaluate our approach on two of the sequences in KITTI-360 that contain the highest number of loop closures: sequence 02 and sequence 09.}

{{\parskip=5pt
\noindent\textit{Freiburg}: We recorded our own dataset by driving around the city of Freiburg, Germany, across different days. We used a car equipped with a Velodyne HDL-64E LiDAR sensor and an Applanix POS LV positioning system. The resulting dataset includes many loops, both from the same and reverse directions. Moreover, differently from the KITTI and KITTI-360 datasets, our Freiburg dataset includes many dynamic objects. The Freiburg dataset is thus used to evaluate the generalization ability of our approach to a different city, different sensor setup, and across different days by training the models on KITTI and KITTI-360, and evaluating them on our own dataset collected in Freiburg, without any re-training or fine-tuning.}}

\subsection{Implementation and Training Details}

Following \cite{Kim2018}, we consider two point clouds as a real loop if the distance between the groundtruth poses is less than four meters. Moreover, we do not search for loop candidates in the past 50 scans to avoid detecting loops in nearby scans.
We train \net\ on sequences 05, 06, 07, and 09 of the KITTI dataset, validate it on sequences 00 and 08, and test it on the KITTI-360 dataset.
We also train a second model, denoted as \textbf{\net$_\dagger$}, which is trained on sequences 00, 04, 05, and 06 of the KITTI-360 dataset, validated on sequences 02 and 09, and tested on the KITTI dataset.

We train all models for $150$ epochs on a server with 4 NVIDIA TITAN RTX GPUs, using a batch size of 24 positive pairs. We use the ADAM optimizer to update the weights of the network, with an initial learning rate of $0.004$ which is halved after epochs $40$ and $80$, and a weight decay of $5\cdot 10^{-6}$. In all the experiments, if not otherwise specified, we set the number of keypoints $N=4096$, the intermediate feature dimension $D=640$, the output feature dimension $G=256$, the number of NetVLAD clusters $K=64$, the triplet margin $m=0.5$, and the distance function in \cref{eq:triplet_loss} as the L2 distance. The number of iterations for the Sinkhorn algorithm is set to $L=5$.

In order to help the network to learn viewpoint-invariant features, we apply a random rigid body transformation to each point cloud, with a maximum translation of $[\pm\, \SI{1.5}{\meter}]$ on the $x$ and $y$ axes, and $[\pm\, \SI{0.25}{\meter}]$ on the $z$ axis; the maximum rotation of $[\pm\, \ang{180}]$ for the yaw (to simulate loop closures from different directions), and $[\pm\, \ang{3}]$ for roll and pitch.

\subsection{Evaluation of Loop Closure Detection}
\label{sec:experiment_lcd}

To evaluate the loop closure detection performance of \net, we use precision-recall curves and the Average Precision (AP) metric under two different evaluation protocols.

\begin{table*}
  \centering
  \caption{Comparison with the state of the art in terms of the average precision evaluated on the KITTI and KITTI-360 datasets.}
  \setlength\tabcolsep{8pt}
  \label{tab:comparison-ap}
\begin{tabular}{clccccp{0.3cm}cccc}
	\toprule
	& Method & \multicolumn{4}{c}{Protocol 1} & & \multicolumn{4}{c}{Protocol 2} \\
	\cmidrule(lr){3-6}	\cmidrule(lr){8-11}
	&& \multicolumn{2}{c}{KITTI} & \multicolumn{2}{c}{KITTI-360} & & \multicolumn{2}{c}{KITTI} & \multicolumn{2}{c}{KITTI-360} \\
	\cmidrule(lr){3-4}	\cmidrule(lr){5-6} \cmidrule(lr){8-9} \cmidrule(lr){10-11}
	&& 00 & 08 & 02 & 09 & & 00 & 08 & 02 & 09 \\
	\midrule
	\multirow{4}{*}{\rotatebox[origin=c]{90}{\parbox[c]{1cm}{\centering\scriptsize Handcrafted}}} & M2DP~\cite{He2016} & 0.93 & 0.05 & 0.15 & 0.66 & & 0.31 & 0.01 & 0.03 & 0.17 \\
	&Scan Context~\cite{Kim2018} & 0.96 & 0.65 & 0.81 & 0.90 & & 0.47 & 0.21 & 0.32 & 0.31 \\
	&ISC~\cite{9196764} & 0.83 & 0.31 & 0.41 & 0.65 & & 0.14 & 0.05 & 0.03 & 0.04 \\
	&LiDAR-Iris~\cite{IRIS_2020} & 0.96 & 0.64 & 0.83 & 0.91 & & 0.42 & 0.17 & 0.25 & 0.26 \\ \midrule
	\multirow{4}{*}{\rotatebox[origin=c]{90}{\parbox[c]{1cm}{\centering\scriptsize DNN-based}}} & OverlapNet~\cite{chen2020rss} & 0.95 & 0.32 & 0.14 & 0.70 & & 0.60 & 0.20 & 0.05 & 0.33 \\
	&SG\_PR~\cite{kong2020semantic} & 0.49 & 0.13 & - & - & & 0.23 & 0.13 & - & - \\
	&\textbf{\net} & \underline{0.97} & \underline{0.94} & \underline{0.95} & \underline{0.98} & & \underline{0.62} & \underline{0.73} & \underline{0.69} & \underline{0.79} \\
	&\textbf{\net$_\dagger$} & \textbf{0.998} & \textbf{0.96} & \textbf{0.97} & \textbf{0.99} & & \textbf{0.89} & \textbf{0.76} & \textbf{0.73} & \textbf{0.80} \\
	\bottomrule
\end{tabular}
\vspace{-3mm}
\end{table*}

{\parskip=5pt
\noindent\textit{Protocol 1}: In the first protocol, we evaluate our approach in a real loop closure setting. For each scan $i$ of the sequence, we compute the similarity between the global descriptor $f(P^i)$ and the descriptor of all the previous scans, excluding the nearby scan as detailed in~\cref{sec:dataset}. We select scan $j$ with the highest similarity as the loop candidate, and if the similarity between the two descriptors is higher than a threshold $th$, then we consider the pair $(i, j)$ as a loop. In such a case, we further check the distance of the groundtruth poses between the two scans: if the distance is less than four meters, then we consider it as a true positive, and as a false positive otherwise.
On the other hand, if the similarity is lower than the threshold, but if a scan within four meters around the current scan $i$ exists, then we consider it as a false negative.}

{\parskip=5pt
\noindent\textit{Protocol 2}: {In the second protocol, for each scan, we take into account all the previous scans, not only the one with the highest similarity. For every pair of scans, if the similarity between the two descriptors is higher than the threshold, we consider the pair as loop closure, and we compare against the groundtruth to compute precision and recall. Although in a real-world loop closure application only the most similar scan matters, if an approach is able to detect loops when the scans are very similar, but fails in more challenging scenarios (such as occlusions), this will not be reflected in the protocol 1 results. In protocol 2, on the other hand, all pairs of scans are considered, and thus approaches that better deal with challenging situations will achieve better results. Also in this protocol, we ignore nearby scans to avoid matching consecutive scans.
}}

\begin{figure*}
\centering
\footnotesize
\setlength{\tabcolsep}{1cm}
{\renewcommand{\arraystretch}{0.5}
\begin{tabular}{P{6.2cm}P{6.2cm}}
\multicolumn{1}{c}{Protocol 1} & \multicolumn{1}{c}{Protocol 2} \\
\\
\raisebox{-0.4\height}{\includegraphics[width=\linewidth]{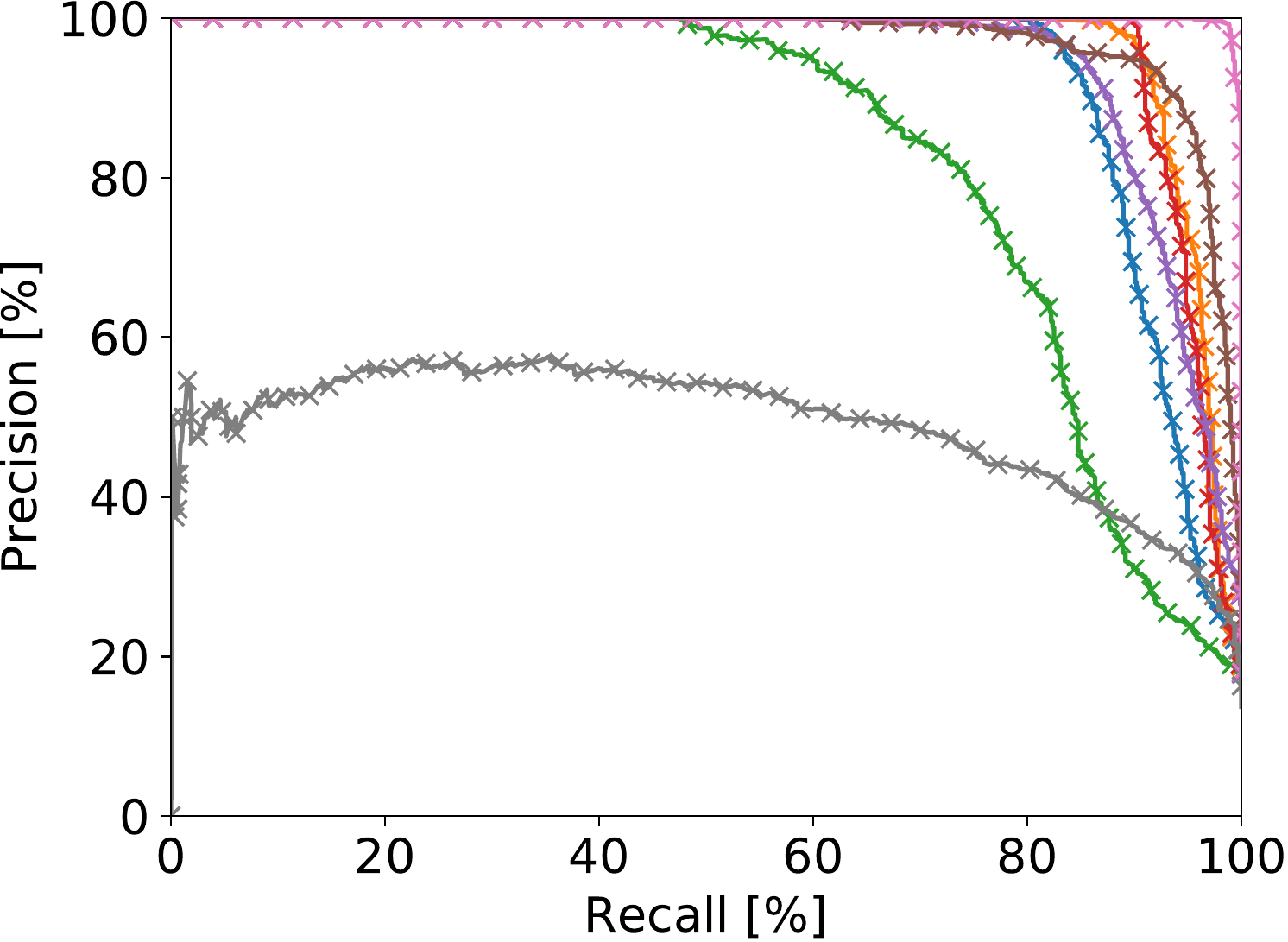}} & \raisebox{-0.4\height}{\includegraphics[width=\linewidth]{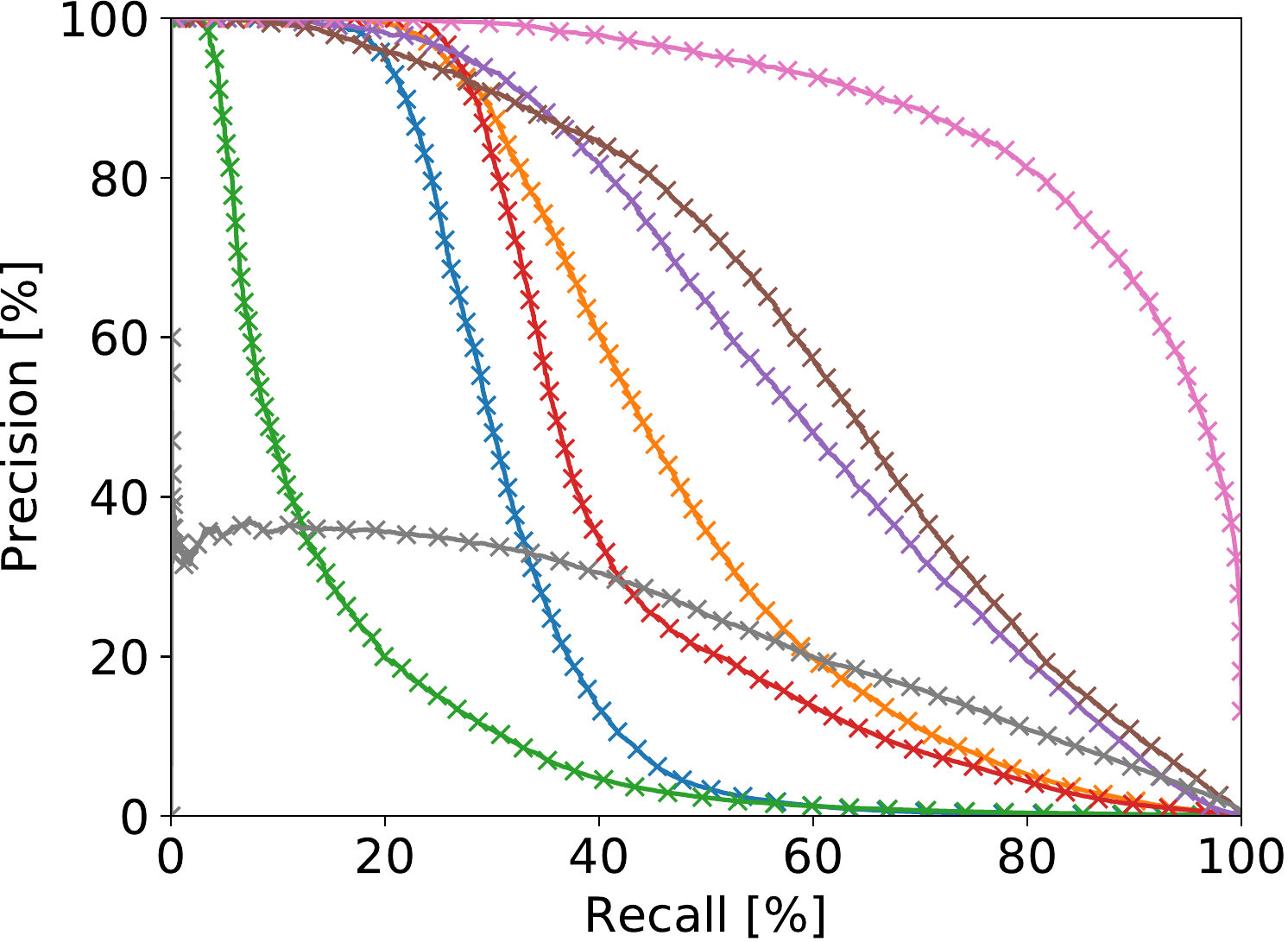}} \\
\multicolumn{2}{c}{(a)~KITTI sequence 00} \\
\\
\\
\includegraphics[width=\linewidth]{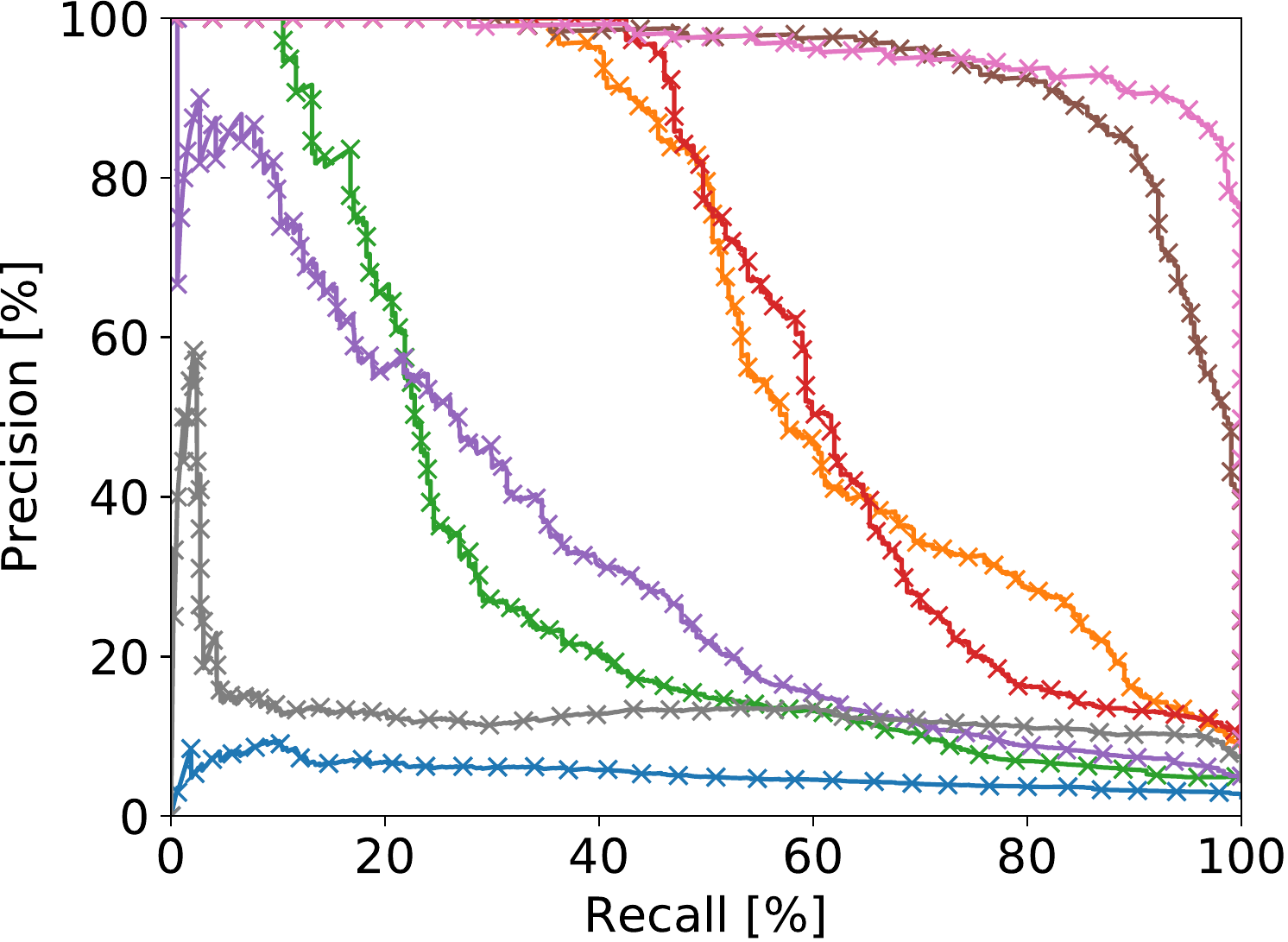} & \includegraphics[width=\linewidth]{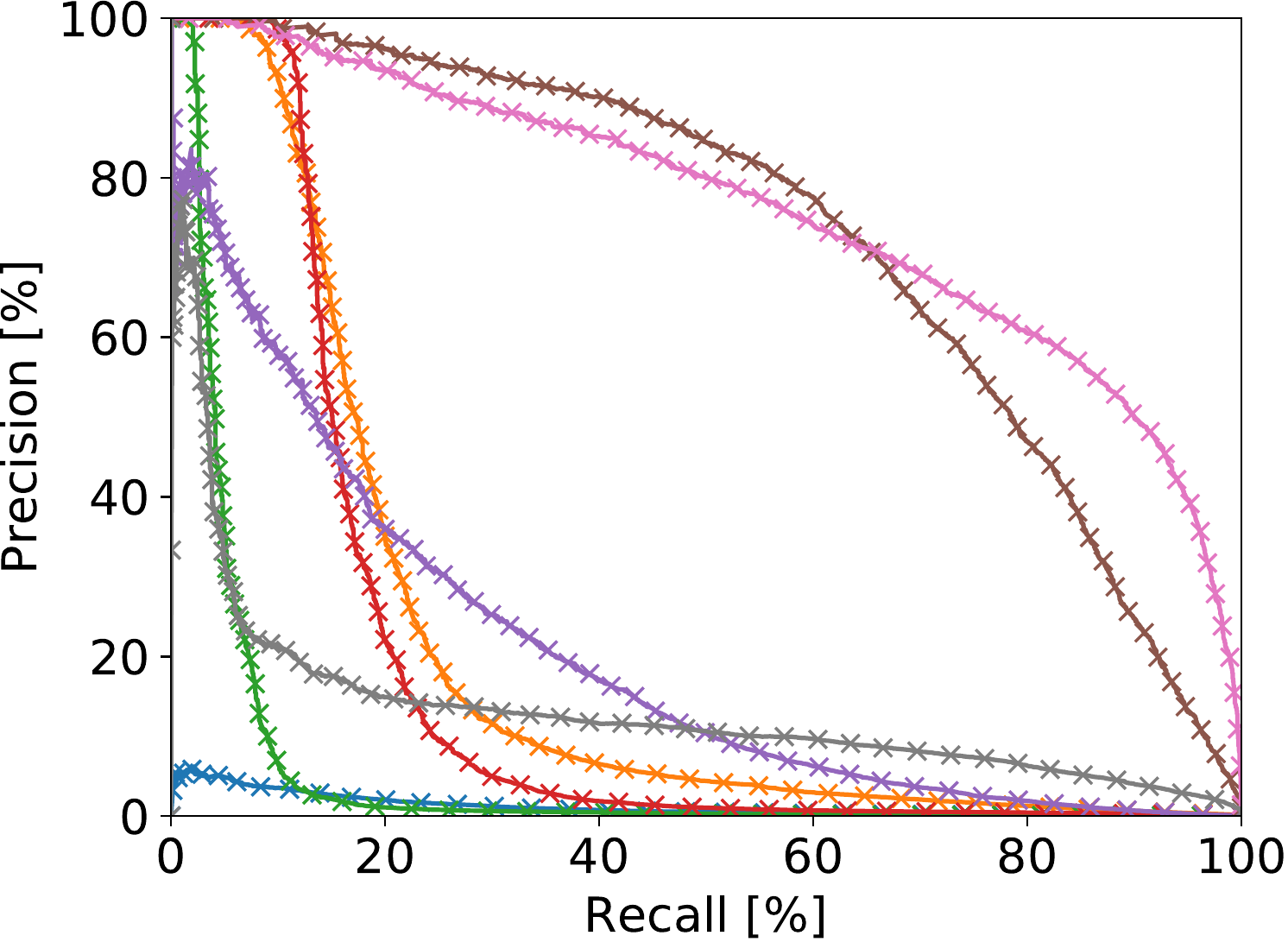} \\
\multicolumn{2}{c}{(b)~KITTI sequence 08} \\
\\
\\
\includegraphics[width=\linewidth]{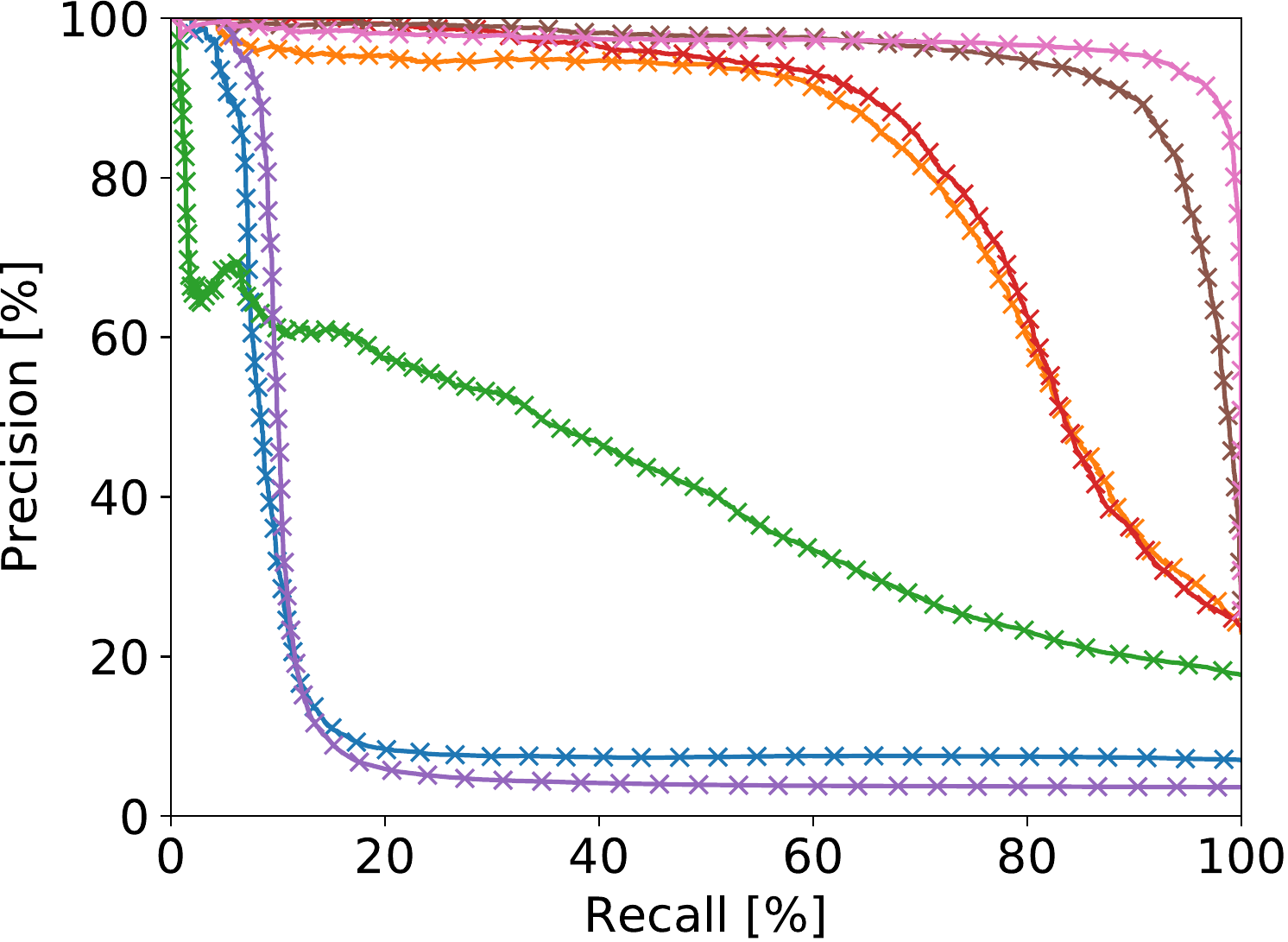} & \includegraphics[width=\linewidth]{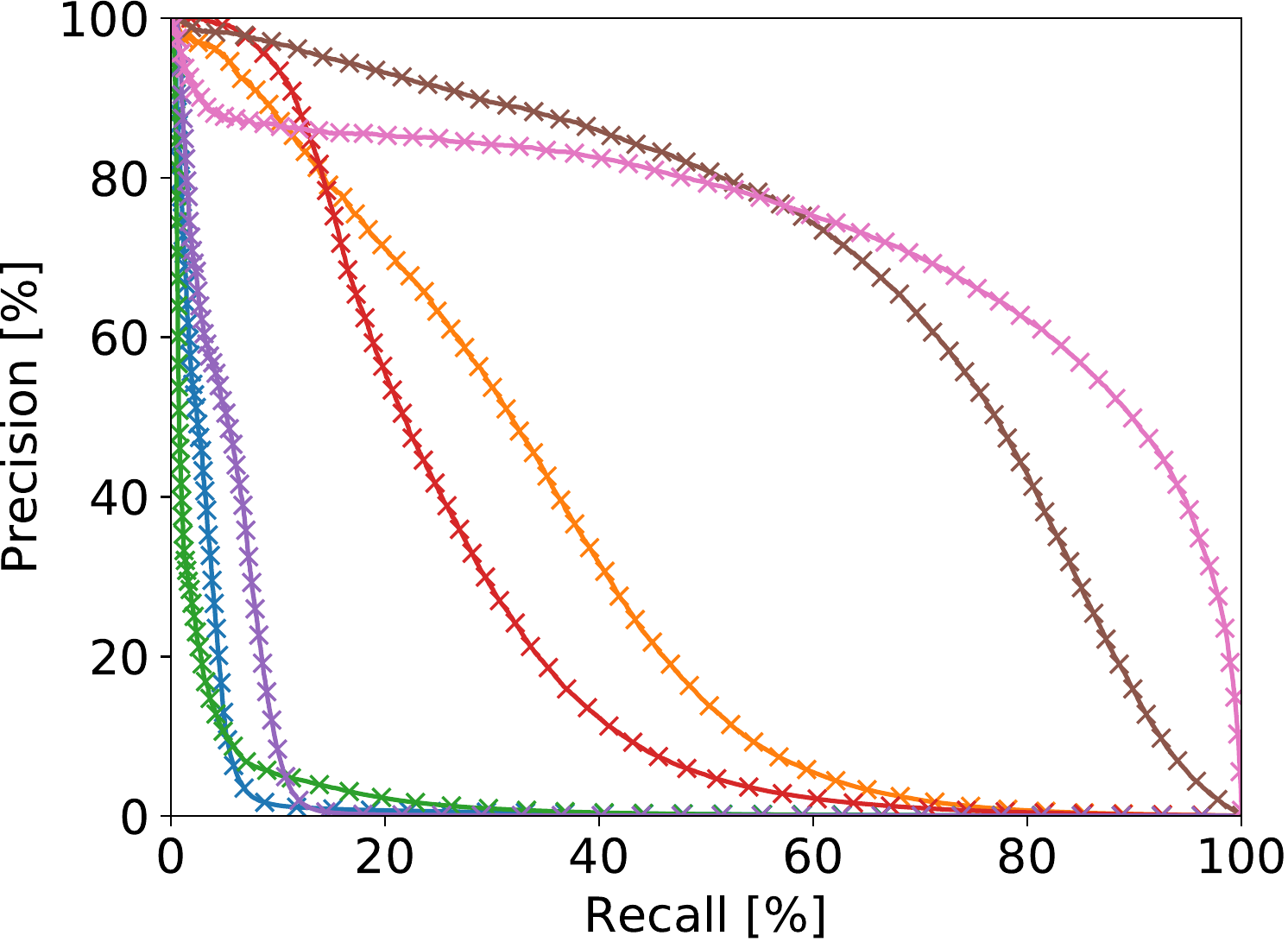} \\
\multicolumn{2}{c}{(c)~KITTI-360 sequence 02} \\
\\
\\
\includegraphics[width=\linewidth]{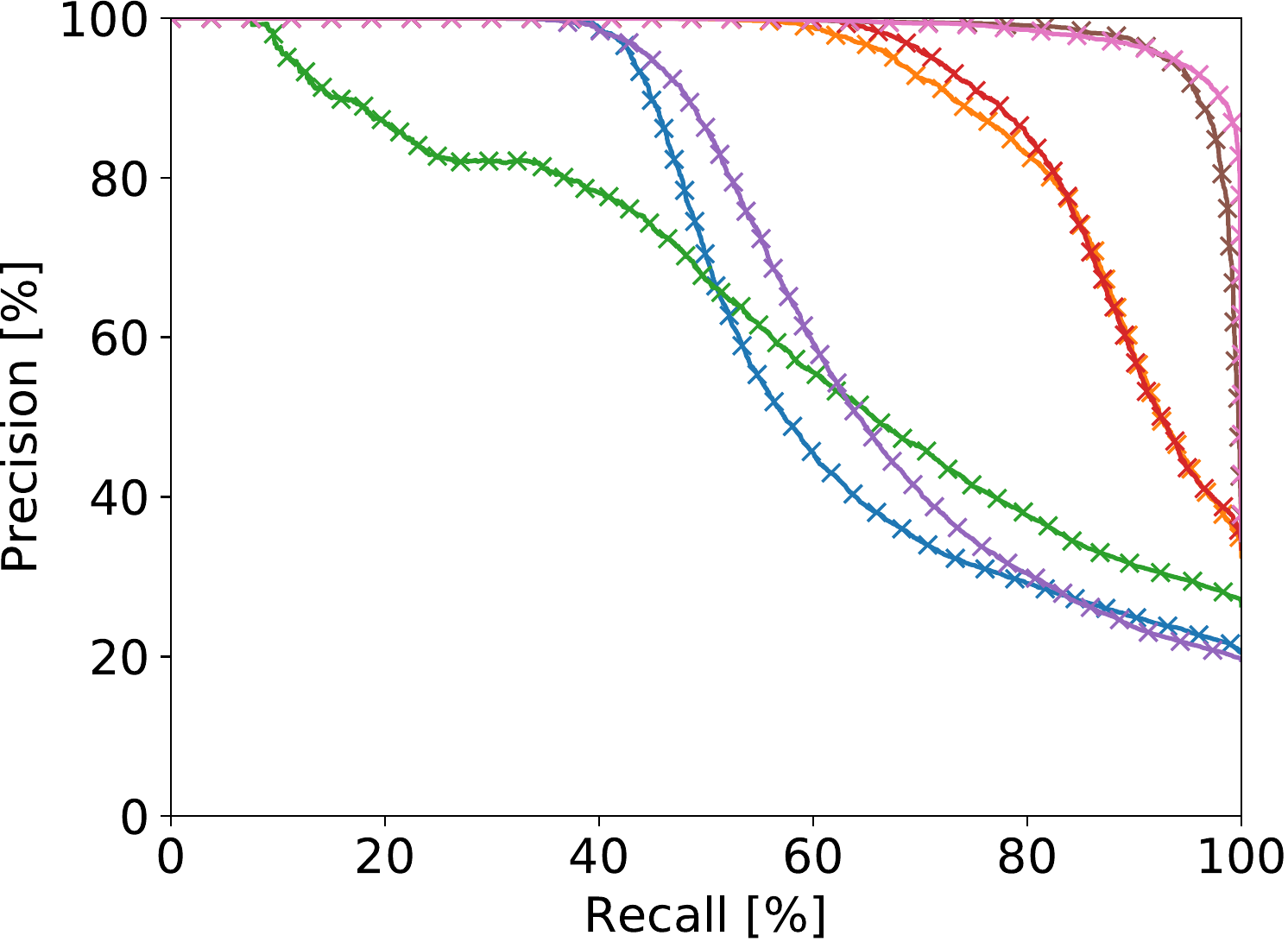} & \includegraphics[width=\linewidth]{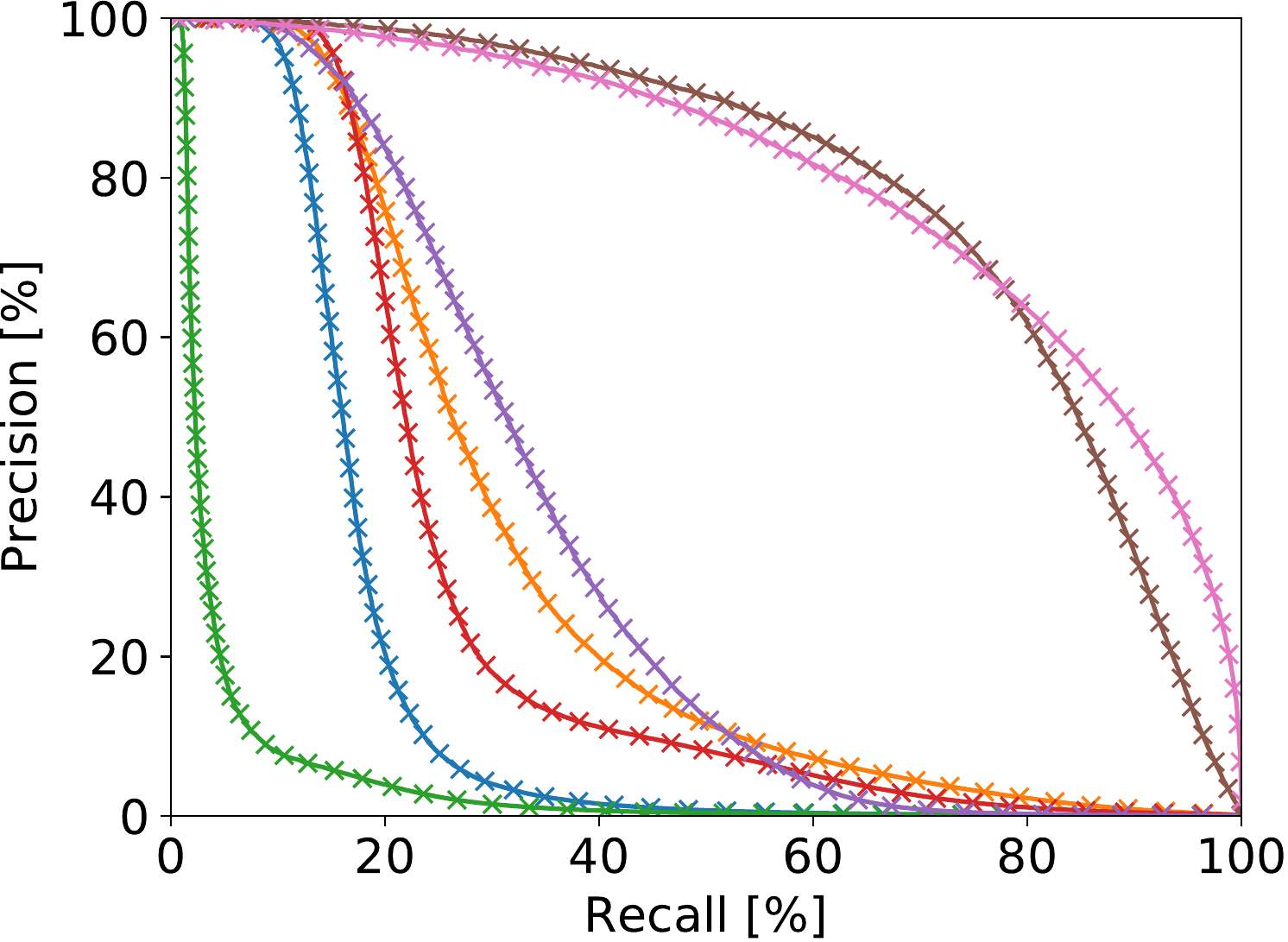} \\
\multicolumn{2}{c}{(d)~KITTI-360 sequence 09} \\
\\
\\
\multicolumn{2}{c}{\includegraphics[width=12cm]{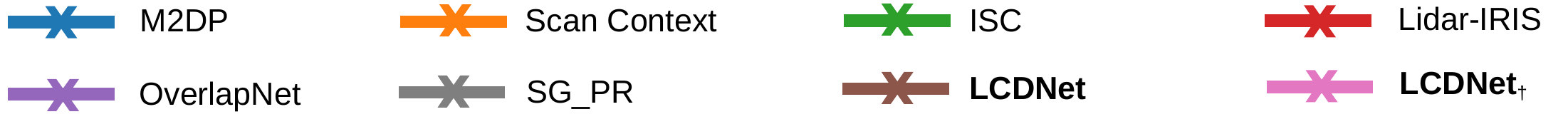}} \\
\end{tabular}}
\caption{Comparison of loop closure detection precision-recall curves on KITTI (a-b) and on KITTI-360 (c-d) datasets evaluated using both protocols. Our proposed \net$_\dagger$ achieves the best performance in all the experiments, followed by our \net\ as second best method. The improvement over previous state-of-the-art approaches is even more prominent when dealing with reverse direction loops, as observed in (b).}
\label{fig:prcurve}
\end{figure*}

In both protocols, by varying the threshold $th$ we obtain a set of pairs (precision, recall), that we use to generate the precision-recall curve and to compute the AP.

We compare our approach with state-of-the-art handcrafted methods: M2DP~\cite{He2016}, Scan-Context~\cite{Kim2018}, Intensity Scan-Context (ISC)~\cite{9196764}, and LiDAR-IRIS~\cite{IRIS_2020}, as well as \ac{DNN}-based methods OverlapNet~\cite{chen2020rss}, and Semantic Graph Place Recognition (SG\_PR)~\cite{kong2020semantic}. For all these approaches, we used the official code published by the respective authors, and the pretrained models that are provided by the authors for \ac{DNN}-based methods. OverlapNet only provides the model trained with geometric information, we refer to this model as OverlapNet (\textit{Geo}).
All the \ac{DNN}-based methods except for \net$_\dagger$ are trained on the KITTI dataset as described in \Cref{sec:dataset}, and evaluated individually on sequences from both KITTI and KITTI-360 datasets.

We present results with the AP metric for protocol~1 and protocol~2 in \cref{tab:comparison-ap}. The best method is highlighted in bold, and the second best is underlined. Moreover, we present the precision-recall curves for both protocols in \cref{fig:prcurve}. We observe that while most approaches achieve satisfactory results in detecting loop closures in the same direction (\cref{fig:prcurve}~(a)), this is not the case for reverse loops as shown in~\cref{fig:prcurve}~(b). M2DP and SG\_PR completely fail on the KITTI sequence~08; Scan Context, OverlapNet (\textit{Geo}) and LiDAR-Iris also show a strong decrease in performance when dealing with reverse loops. For instance, the previous state-of-the-art method Scan Context achieved an AP of $0.96$ in sequence 00 of the KITTI dataset (which contains only same direction loops), and $0.65$ in sequence 08. Our proposed \net, on the other hand, performs equally well for both reverse and same direction loops, achieving an AP of $0.94$ and $0.97$ respectively. This is even more noticeable in the results using protocol~2 where all the other approaches show a substantial decrease in performance, while our \net\ achieves an AP score that is even better for detecting reverse loops than the same direction loops. We also observe that the model trained on the KITTI-360 dataset (\net$_\dagger$) achieves the best performance on all the sequences, thereby setting the new state-of-the-art on both KITTI and KITTI-360.\looseness-1

\subsection{Evaluation of Relative Pose Estimation}
\label{sec:comparison-pos}

\begin{table*}
  \centering
  \caption{Comparison of relative pose errors (rotation and translation) between positive pairs on the KITTI dataset.}
  \label{tab:comparison-pos1}
  \begin{threeparttable}
  \begin{tabular}{clcccccc}
  \toprule
  &Approach & \multicolumn{3}{c}{Seq. 00} & \multicolumn{3}{c}{Seq. 08} \\ \cmidrule(lr){3-5} \cmidrule(lr){6-8}
   && Success & TE [m] (succ. / all) & RE [deg] (succ. / all) & Success & TE [m] (succ. / all) & RE [deg] (succ. / all) \\ \midrule
   \multirow{7}{*}{\rotatebox[origin=c]{90}{\parbox[c]{2cm}{\centering\scriptsize Handcrafted}}} &Scan Context$^*$~\cite{Kim2018} & 97.66\% & - / - & 1.34 / 1.92 & 98.21\% & - / - & 1.71 / 3.11 \\
   & ISC$^*$~\cite{9196764} & 32.07\% & - / - & 1.39 / 2.13 & 81.28\% & - / - & 2.07 / 6.27 \\
  &LiDAR-Iris$^*$~\cite{IRIS_2020} & 98.83\% & - / - & 0.65 / 1.69 & \underline{99.29\%} & - / - & 0.93 / 1.84 \\ 
  & ICP (P2p)~\cite{zhang1994iterative} & 35.57\% & 0.97 / 2.08 & 1.36 / 8.98 & 0\% & - / 2.43 & - / 160.46 \\
  & ICP (P2pl)~\cite{zhang1994iterative} & 35.54\% & 1.00 / 2.11 & 1.39 / 8.99 & 0\% & - / 2.44 & - / 160.45 \\
  & RANSAC~\cite{Rusu_ICRA_2009} & 33.95\% & 0.98 / 2.75 & 1.37 / 12.01 & 15.61\% & 1.33 / 4.57 & 1.79 / 37.31 \\
  & FGR~\cite{Zhou_ECCV_2016} & 34.54\% & 0.98 / 5972.31 & 1.2 / 12.79 & 17.16\% & 1.32 / 35109.13 & 1.76 / 28.98 \\
  & {TEASER++}~\cite{Yang2020tro} & {34.06\%} & {0.98 / 2.72} & {1.33 / 15.85} & {17.13\%} & {1.34 / 3.83} & {1.93 / 29.19} \\\midrule
  \multirow{4}{*}{\rotatebox[origin=c]{90}{\parbox[c]{1.2cm}{\centering\scriptsize DNN-based}}} &OverlapNet$^*$~\cite{chen2020rss} & 83.86\% & - / - & 1.28 / 3.89 & 0.10\% & - / - & 2.03 / 65.45 \\ 
  &RPMNet~\cite{yew2020} & 47.31\% & 1.05 / 2.07 & 0.60 / 1.88 & 27.80\% & 1.28 / 2.42 & 1.77 / 13.13 \\
  &{DCP~\cite{wang2019deep}} & {50.71\%} & {0.98 / 1.83} & {1.14 / 6.61} & {0\%} & {- / 4.01} & {- / 161.24} \\
  &{PCAM~\cite{cao21pcam}} & {99.68\%} & {\textbf{0.07 / 0.08}} & {0.35 / 0.74} & {94.90\%} & {0.19 / 0.41} & {0.51 / 6.01} \\
   \midrule
  \multirow{3}{*}{\rotatebox[origin=c]{90}{\parbox[c]{1.3cm}{\centering\scriptsize \textbf{Ours}}}}&\net\ (fast) & 93.03\% & 0.65 / 0.77 & 0.86 / 1.07 & 60.71\% & 1.02 / 1.62 & 1.65 / 3.13 \\
  & \net & \textbf{100\%} & \underline{0.11 / 0.11} & \textbf{0.12 / 0.12} & \textbf{100\%} & \textbf{0.15 / 0.15} & \textbf{0.34 / 0.34} \\
  &\net$_\dagger$ (fast) & \underline{99.79\%} & 0.28 / 0.29 & 0.30 / 0.30 & 88.51\% & 0.66 / 0.93 & 1.00 / 1.31 \\
  & \net$_\dagger$ & \textbf{100\%} & 0.14 / 0.14 & \underline{0.14 / 0.14} & \textbf{100\%} & \underline{0.18 / 0.18} & \underline{0.36 / 0.36} \\
  \midrule
  
  & \net\ + ICP & 100\% & 0.04 / 0.04 & 0.09 / 0.09 & 100\% & 0.09 / 0.09 & 0.33 / 0.33 \\
  & \net$_\dagger$ + ICP & 100\% & 0.04 / 0.04 & 0.08 / 0.08 & 100\% & 0.07 / 0.07 & 0.32 / 0.32 \\ 
  & {\net\ + TEASER} & {94.39\%} & {0.66 / 0.77} & {0.09 / 0.10} & {71.99\%} & {1.05 / 1.62} & {0.33 / 0.35} \\
  & {\net$_\dagger$ + TEASER} & {99.78\%} & {0.28 / 0.29} & {0.09 / 0.09} & {89.39\%} & {0.67 / 0.93} & {0.33 / 0.34} \\
  \bottomrule
  \end{tabular}
  \begin{tablenotes}[para,flushleft]
       \footnotesize      
       $^*$ these approaches only estimate the rotation between two point clouds, therefore are not directly comparable with the other approaches which estimate the full 6-DoF transformation {under driving conditions}.
     \end{tablenotes}
   \end{threeparttable}
   \vspace{-3mm}
\end{table*}

\begin{table*}
  \centering
  \caption{Comparison of relative pose errors (rotation and translation) between positive pairs on the KITTI-360 dataset.}
  \label{tab:comparison-pos2}
  \begin{threeparttable}
  \begin{tabular}{clcccccc}
  \toprule
  &Approach & \multicolumn{3}{c}{Seq. 02} & \multicolumn{3}{c}{Seq. 09} \\ \cmidrule(lr){3-5} \cmidrule(lr){6-8}
   && Success & TE [m] (succ. / all) & RE [deg] (succ. / all) & Success & TE [m] (succ. / all) & RE [deg] (succ. / all) \\ \midrule
   \multirow{7}{*}{\rotatebox[origin=c]{90}{\parbox[c]{2cm}{\centering\scriptsize Handcrafted}}} &Scan Context$^*$~\cite{Kim2018} & 92.31\% & - / - & 1.60 / 5.49 & 95.25\% & - / - & 1.40 / 6.80 \\
   & ISC$^*$~\cite{9196764} & 83.15\% & - / - & 1.71 / 3.44 & 86.26\% & - / - & 1.51 / 7.08 \\
  &LiDAR-Iris$^*$~\cite{IRIS_2020} & 96.54\% & - / - & 1.07 / 2.24 & 97.63\% & - / - & 0.72 / 3.80 \\ 
  & ICP (P2p)~\cite{zhang1994iterative} & 4.19\% & 1.10 / 2.26 & 1.74 / 149.76 & 21.24\% & 1.06 / 2.22 & 1.34 / 66.34 \\
  & ICP (P2pl)~\cite{zhang1994iterative} & 4.19\% & 1.11 / 2.30 & 1.18 / 149.39 & 21.29\% & 1.07 / 2.24 & 1.38 / 66.23 \\
  & RANSAC~\cite{Rusu_ICRA_2009} & 24.78\% & 1.24 / 3.67 & 1.83 / 32.22 & 29.69\% & 1.12 / 3.14 & 1.48 / 23.42 \\
  & FGR~\cite{Zhou_ECCV_2016} & 27.92\% & 1.23 / 6758.87 & 1.85 / 18.16 & 30.46\% & 1.12 / 6011.39 & 1.44 / 17.35 \\
  & {TEASER++}~\cite{Yang2020tro} & {27.02\%} & {1.25 / 3.16} & {1.83 / 19.16} & {30.32\%} & {1.14 / 2.91} & {1.46 / 19.22} \\\midrule
  \multirow{4}{*}{\rotatebox[origin=c]{90}{\parbox[c]{1.2cm}{\centering\scriptsize DNN-based}}} &OverlapNet$^*$~\cite{chen2020rss} & 11.42\% & - / - & 1.79 / 76.74 & 54.33\% & - / - & 1.38 / 33.62 \\ 
  &RPMNet~\cite{yew2020} & 37.99\% & 1.18 / 2.26 & 1.30 / 5.97 & 41.42\% & 1.13 / 2.21 & 1.02 / 3.95 \\
  &{DCP~\cite{wang2019deep}} & {5.62\%} & {1.09 / 3.14} & {1.36 / 149.27} & {30.10\%} & {1.04 / 2.30} & {1.06 / 64.86} \\
   &{PCAM~\cite{cao21pcam}} & {97.46\%} & {\textbf{0.20 / 0.30}} & {0.75 / 1.36} & {\underline{99.78\%}} & {\textbf{0.12 / 0.13}} & {0.51 / 0.64} \\
  \midrule
  \multirow{3}{*}{\rotatebox[origin=c]{90}{\parbox[c]{1.3cm}{\centering\scriptsize \textbf{Ours}}}}&\net\ (fast) & 83.92\% & 0.84 / 1.10 & 1.28 / 1.67 & 89.49\% & 0.76 / 0.94 & 0.99 / 1.19 \\
  & \net & \textbf{98.62\%} & 0.28 / 0.32 & \underline{0.32 / 0.35} & \textbf{100\%} & \underline{0.18 / 0.18} & \textbf{0.20 / 0.20} \\ 
  &\net$_\dagger$ (fast) & 89.07\% & 0.40 / 0.45 & 0.57 / 0.62 & 98.87\% & 0.43 / 0.44 & 0.59 / 0.63 \\
  & \net$_\dagger$ & \underline{98.55\%} & \underline{0.27 / 0.32} & \textbf{0.32 / 0.34} & \textbf{100\%} & 0.20 / 0.20 & \underline{0.22 / 0.22} \\
  \midrule
  & \net\ + ICP & 98.51\% & 0.20 / 0.25 & 0.24 / 0.27 & 100\% & 0.10 / 0.10 & 0.15 / 0.15 \\
  & \net$_\dagger$ + ICP & 98.51\% & 0.20 / 0.25 & 0.24 / 0.27 & 100\% & 0.11 / 0.11 & 0.15 / 0.15 \\ 
  & {\net\ + TEASER} & {86.63\%} & {0.85 / 1.10} & {0.40 / 0.52} & {90.57\%} & {0.76 / 0.94} & {0.22 / 0.25} \\
  & {\net$_\dagger$ + TEASER} & {98.06\%} & {0.40 / 0.45} & {0.37 / 0.45} & {99.10\%} & {0.43 / 0.44} & {0.22 / 0.23} \\ 
  \bottomrule
  \end{tabular}
  \begin{tablenotes}[para,flushleft]
       \footnotesize      
       $^*$ these approaches only estimate the rotation between two point clouds, therefore are not directly comparable with the other approaches which estimate the full 6-DoF transformation {under driving conditions}.
     \end{tablenotes}
   \end{threeparttable}
   \vspace{-3mm}
\end{table*}

In this section, we evaluate the relative pose estimation between two point clouds. Our proposed \net\ provides a full 6-DoF transformation {under driving conditions} between two points clouds. However, Scan Context, ISC, LiDAR-Iris, and OverlapNet only provide an estimation of the yaw angle. As M2DP, and SG\_PR do not provide any information about the relative pose, we do not include them in the results presented in this section.
{Moreover, we compare our approach with state-of-the-art handcrafted methods for point cloud registration: \ac{ICP}~\cite{zhang1994iterative} using point-to-point and point-to-plane distances, \ac{RANSAC} with FPFH features~\cite{Rusu_ICRA_2009} and Fast Global Registration (FGR)~\cite{Zhou_ECCV_2016}, all implemented in the Open3D library~\cite{Zhou2018}, and TEASER++~\cite{Yang2020tro} using the official implementation.
We also compare with \ac{DNN}-based methods RPMNet~\cite{yew2020}, Deep Closest Point (DCP)~\cite{wang2019deep} and Product of Cross-Attention Matrices (PCAM)~\cite{cao21pcam}.
To provide a fair comparison, we trained all the latter \ac{DNN}-based approaches on the same data, following the same protocol, and using the same number of keypoints used to train our \net.
Following~\cite{choy2020deep}, for the aforementioned handcrafted methods we first downsample the point clouds using a voxel size of 0.3 meter, while the latter \ac{DNN}-based methods and our \net\ perform point cloud registration using 4096 sampled points, which is a much sparser representation. Scan-Context, LiDAR-Iris, ISC, and OverlapNet, on the other hand, operate on spherical projections of the points, and thus they process almost all the points in the original cloud.}
We evaluate two versions of our method. The first one, denoted as \textit{\net\ (fast)}, leverages the output of the UOT-based relative position head to estimate the transformation. In the second version, denoted as \textit{\net}, we replace the UOT-based head with a RANSAC estimator, as described in \cref{sec:posehead}.
The models trained on KITTI-360 are denoted as \textit{\net$_\dagger$ (fast)} and \textit{\net$_\dagger$}, respectively.
{We also evaluate the performance of \net\ followed by a further \ac{ICP} registration. We report the latter evaluation only as a reference to show the best alignment achievable.
Finally, we further investigate whether TEASER++ is a better pose estimator by replacing RANSAC in \net.
}

We evaluate all the methods in terms of success rate (percentage of successfully aligned pairs), translation error (TE), and rotation error (RE) averaged over successful pairs as well as over all the positive pairs. We consider two pairs to be aligned successfully if the final {rotation and translation} error is below five degrees and two meters, respectively. The results on the KITTI and KITTI-360 datasets are reported in~\cref{tab:comparison-pos1,tab:comparison-pos2}.
{We observe that LiDAR-Iris achieves the best performance among the handcrafted methods and PCAM demonstrates superior results compared to existing \ac{DNN}-based approaches when dealing with same and reverse direction pairs.
However, as opposed to the other methods, PCAM only performs point cloud registration and do not provide any information regarding loop closure detection.
Whereas, our proposed \net\ and \net$_\dagger$ achieve the highest success rates and lowest rotation errors compared to all the methods
, with a success rate of 100\% in three out of four sequences.
PCAM, on the other hand, achieves the lowest translation errors in most sequences, but is not robust to registration under partial overlap, as we discuss in \cref{sec:partial}.}
The fast versions of our method achieve results comparable with, and in some sequences even better than existing approaches, while being much faster than most point cloud registration methods, as we discuss in \cref{sec:runtime}.
{We observe that by replacing RANSAC in \net\ and \net$_\dagger$ with TEASER++ the success rates decrease and the translation errors significantly increase, while the rotation errors remain similar.
During our experimental evaluations, we also observed that while the rotation and translation invariance obtained by our \net\ primarily arise from our data augmentation scheme, many existing loop closure detection approaches (not reported in the comparison) did not converge at all when trained with the same scheme. Therefore, we argue that data augmentation by itself is not sufficient, and a well-designed architecture and loss function is necessary to achieve invariance.}

\subsection{{Partial Overlap}}
\label{sec:partial}
{
In this section, we evaluate the ability of \net\ in detecting loops and regressing the relative pose between point clouds that only overlap partially. To do so, we follow the same evaluation protocol that we use in \cref{sec:experiment_lcd} (protocol 1) and \cref{sec:comparison-pos}. We simulate partial overlapping pairs by removing a random section of each point cloud. We compare \net\ against state-of-the-art approaches on the sequence~08 of the KITTI dataset under two settings: by removing a random \SI{45}{\degree} and \SI{90}{\degree} sector, respectively. \Cref{tab:comparison-partial} reports the results of this experiment in terms of average precision (AP), success rate, mean translation error and mean rotation error. Although the AP of \net\ drops moderately when a \ang{90} section is removed, \net$_\dagger$ still achieves an AP higher than all the existing approaches evaluated on the complete overlap test (\cref{tab:comparison-ap}). We observe that PCAM which achieves remarkable results in the full overlap registration test, struggles when dealing with partial overlapping point clouds with a success rate that drops from 95\% to 56\%, a translation error that increases from \SI{0.41}{\meter} to \SI{3.32}{\meter}, and a rotation error that raises from \ang{6.01} to \ang{34.64}. \net\ and \net$_\dagger$, on the other hand, retain an almost perfect success rate and slightly lower translation and rotation errors.\looseness=-1

We also investigated the MulRan dataset~\cite{gskim-2020-mulran} for this experiment, as the LiDAR mounted on their vehicle is obstructed by the radar sensor for approximately \ang{70} rear FOV. Therefore, in reverse direction scenarios, the scans share only a very limited overlap. In preliminary evaluations, all the considered approaches failed in detecting reverse loops. We argue that this is a limitation of all scan-to-scan methods, and that scan-to-map approaches should be considered in these scenarios.

\begin{table*}

  \centering
  \caption{Comparison of loop closure detection (AP) and relative pose errors (rotation and translation) under partial overlap on the sequence~08 of the KITTI dataset.}
  \label{tab:comparison-partial}
  \begin{threeparttable}
  \begin{tabular}{clcccccccc}
  \toprule
  &Approach & \multicolumn{4}{c}{\SI{45}{\degree}} & \multicolumn{4}{c}{\SI{90}{\degree}} \\ \cmidrule(lr){3-6} \cmidrule(lr){7-10}
  && AP & Success & TE [m] (all) & RE [deg] (all) & AP & Success  & TE [m] (all) & RE [deg] (all) \\ \midrule
  \multirow{7}{*}{\rotatebox[origin=c]{90}{\parbox[c]{2cm}{\centering\scriptsize Handcrafted}}} &Scan Context$^*$~\cite{Kim2018} & 0.52 & 27.33\% & - & 57.70 & 0.40 & 17.40\% & - & 72.05 \\
  & LiDAR-Iris$^*$~\cite{IRIS_2020} & 0.43 & 97.84\% & - & 2.78 & 0.22 & 96.28\% & - & 5.13 \\
  & ICP (P2p)~\cite{zhang1994iterative} & - & 0\% & 2.42 & 160.46 & - & 0\% & 2.42 & 160.46 \\
  & ICP (P2pl)~\cite{zhang1994iterative} & - & 0\% & 2.45 & 160.46 & - & 0\% & 2.45 & 160.42 \\
  & RANSAC~\cite{Rusu_ICRA_2009} & - & 15.51\% & 4.88 & 43.77 & - & 13.78\% & 5.50 & 48.74 \\
  & FGR~\cite{Zhou_ECCV_2016} & - & 16.55\% & \num{44439.37} & 30.30 & - & 14.49\% & \num{235332.54} & 34.20 \\
  & TEASER++~\cite{Yang2020tro} & - & 16.42\% & 4.03 & 30.32 & - & 15.98\% & 4.37 & 34.99 \\ \midrule
  \multirow{2}{*}{\rotatebox[origin=c]{90}{\parbox[c]{0.5cm}{\centering\scriptsize DNN-based}}} &OverlapNet$^*$~\cite{chen2020rss} & 0.09 & 1.11\% & - & 70.69 & 0.01 & 0.68\% & - & 85.68 \\ 
  &PCAM~\cite{cao21pcam} & - & 84.67\% & 1.04 & 11.80 & - & 55.62\% & 3.32 & 34.64 \\
  \midrule
  \multirow{2}{*}{\rotatebox[origin=c]{90}{\parbox[c]{0.5cm}{\centering\scriptsize \textbf{Ours}}}}& \net & \underline{0.79} & \textbf{100\%} & \underline{0.20} & \underline{0.38} & \underline{0.59} & \underline{99.93\%} & \underline{0.24} & \underline{0.46} \\ 
  & \net$_\dagger$ & \textbf{0.83} & \textbf{100\%} & \textbf{0.19} & \textbf{0.36} & \textbf{0.70} & \textbf{100\%} & \textbf{0.21} & \textbf{0.37}  \\
  \bottomrule
  \end{tabular}
  \begin{tablenotes}[para,flushleft]
      \footnotesize      
      $^*$ these approaches only estimate the rotation between two point clouds, therefore are not directly comparable with the other approaches which estimate the full 6-DoF transformation {under driving conditions}.
     \end{tablenotes}
  \end{threeparttable}
  \vspace{-3mm}
\end{table*}
}

\subsection{Ablation Studies}
\label{sec:ablation}

\begin{table}
  \centering
  \caption{Ablation study on the backbone network architecture.}
  \label{tab:ablationBackbone}
  \setlength\tabcolsep{10pt}
\begin{tabular}{lccc}
	\toprule
  Backbone & AP & TE [m] & RE [deg] \\
	\midrule
	 PointNet~\cite{Qi_2017_CVPR} & \underline{0.67} & 5.15 & 34.14 \\
	 EdgeConv~\cite{dgccn2019} & 0.52 & 5.44 & \underline{16.85} \\
	 RandLA-Net~\cite{hu2020randla} & 0.55 & \underline{3.55} & 20.08 \\
	 \textbf{PVRCNN}~\cite{shi2020pv} & \textbf{0.94} & \textbf{1.62} & \textbf{3.13} \\
	\bottomrule
\end{tabular}
\vspace{-3mm}
\end{table}

\begin{table*}
  \centering
  \caption{Ablation study on the different architectural components of our \net\ evaluated on sequence~08 of the KITTI dataset.}
  \label{tab:ablation}
  \setlength\tabcolsep{6pt}
\begin{tabular}{p{2.5cm}p{1.3cm}p{1.5cm}p{1.5cm}|p{0.5cm}p{1cm}p{1.5cm}p{1.5cm}}
	\toprule
	Relative & Feature & Auxiliary & Num & & AP & TE [m] & RE [deg] \\
	Pose Head & Size D & Loss & Keypoints & & & &  \\
	\midrule
	 UOT & \multirow{4}{*}{640} & \multirow{4}{*}{\ding{51}} & \multirow{4}{*}{4096} & & \textbf{0.94} & \underline{1.62} & \underline{3.13} \\
	 MLP (sin-cos) & & & & & 0.75 & 2.14 & 21.05 \\ 
	 MLP (quat) & & & & & 0.78 & 2.43 & 35.16 \\ 
	 MLP (bingham) & & & & & 0.75 & 2.27 & 22.69 \\ \midrule
  \multirow{4}{*}{UOT} & 640 & \multirow{4}{*}{\ding{51}} & \multirow{4}{*}{4096} & & \textbf{0.94} & \underline{1.62} & \underline{3.13} \\
	 & 128 &  &  & & \underline{0.92} & 1.85 & 3.19 \\
	 & 64 &  &  & & \underline{0.92} & 1.99 & 3.33 \\
	 & 32 &  &  & & 0.86 & 2.23 & 4.09 \\ \midrule
   \multirow{2}{*}{UOT} & \multirow{2}{*}{640} & \ding{51} & \multirow{2}{*}{4096} & & \textbf{0.94} & \underline{1.62} & \underline{3.13} \\
	 &  & \ding{55} &  & & 0.83 & 6.00 & 4.71 \\ \midrule
\multirow{5}{*}{UOT} & \multirow{5}{*}{640} & \multirow{5}{*}{\ding{51}} & 8192 & & \textbf{0.94} & \textbf{1.28} & \textbf{1.99} \\
	 &  &  & 4096 & & \textbf{0.94} & \underline{1.62} & \underline{3.13} \\
	 &  &  & 2048 & & 0.85 & 4.68 & 3.73 \\
   &  &  & 1024 & & 0.69 & 5.17 & 4.75 \\
	 &  &  & 512 & & 0.50 & 4.79 & 4.85 \\
	\bottomrule
\end{tabular}
\vspace{-3mm}
\end{table*}

In this section, we present ablation studies on the different architectural components of our proposed \net. All the models presented in this section are trained on the KITTI dataset, and evaluated on the sequence~08 using the AP, mean rotation error (RE) and mean translation error (TE) metrics.
We choose sequence~08 as the validation set since it is the most challenging sequence, containing only reverse direction loops.
Since \ac{RANSAC} does not influence the training of the network, in this section the rotation and translation errors are computed using the \textit{\net\ (fast) }version.

We first compare our feature extractor built upon PVRCNN presented in~\Cref{sec:feature_extractor} with three different backbones: the widely adopted feature extractor PointNet~\cite{Qi_2017_CVPR}, the dynamic graph CNN EdgeConv~\cite{dgccn2019}, and the recent state-of-the-art semantic segmentation network RandLA-Net~\cite{hu2020randla}. We modified all backbones in order to output a feature vector of size $D=640$ for $N=4096$ points, similar to our backbone. We report results in~\Cref{tab:ablationBackbone}. 
{The ability of our feature extractor presented in \Cref{sec:feature_extractor} to combine high-level features from the 3D voxel \ac{DNN} with fine-grained details provided by the PointNet-based voxel set abstraction layer is demonstrated by the superior performance compared to other backbones, outperforming them in every metric by a large margin}. Our backbone built upon PV-RCNN achieves an average precision of $0.94$ compared to $0.67$ achieved by the second best backbone. For relative pose estimation, PV-RCNN achieves a mean rotation error of \ang{3.13} and a mean translation error of \SI{1.62}{\meter} compared to \ang{16.85} achieved by EdgeConv and \SI{3.55}{\meter} achieved by RandLA-Net.

In~\Cref{tab:ablation}, we present ablation studies on the architecture of the relative pose head, the dimensionality of the extracted point features, the effect of the auxiliary optimal transport loss presented in~\Cref{eq:aux_loss}, and the number of keypoints. We first compare our UOT-based relative pose head presented in~\Cref{sec:posehead} with a \ac{MLP} that directly regresses the rotation and translation, similar to \cite{schaupp2019}. In particular, we train three models using different rotation representations. The first model, \textit{MLP(sin-cos)} uses two parameters to represent the rotation: the sine and cosine of the yaw angle. \textit{MLP(quat)} represents the rotation as unit quaternions, and \textit{MLP(bingham)} uses the Bingham representation proposed in~\cite{peretroukhin_so3_2020}. From the first set of rows in \Cref{tab:ablation}, we observe that our proposed relative pose head significantly outperforms the \ac{MLP}-based heads, especially in the rotation estimation. Our proposed relative pose head achieves a mean rotation error of $3.13^\circ$ compared to $21.05^\circ$ achieved by the best \ac{MLP} model.
Moreover, the UOT-based head favors keypoint features that are rotation and translation invariant, thus enabling the backbone to learn more discriminative features, consequently also improving the loop closure detection performance. The \ac{MLP} based heads, on the other hand, require rotation specific features in order to predict the transformation, which hinders the performance of the place recognition head, which can be observed from the lower average precision achieved by these models.

Subsequently, we study the influence of the dimensionality of point features on the performance of our approach. We train four models by varying dimensionality $D$ as 640, 128, 64, and 32. From the results shown in the second set of rows in \Cref{tab:ablation}, we observe that the performances decrease with lowering the dimensionality $D$.

We evaluate the performance of \net\ without the auxiliary loss presented in \Cref{eq:aux_loss}. From the results shown in the third set of rows of \Cref{tab:ablation}, we observe that when training without the optimal transport loss, the performance in terms of average precision and relative transformation decreases significantly. This demonstrates that the auxiliary optimal transport loss enables the network to learn more distinctive features, which benefits the performance of both loop closure detection and relative transformation estimation.

Finally, in the last set of rows of \Cref{tab:ablation} we compare the performance of \net\ to changes in the number of selected keypoints $N$. Predictably, the performances increase with adding more keypoints. However, the average precision does not improve when increasing the number of keypoints to 8192. Therefore, due to the higher memory and computation required, we use 4096 keypoints in our final model.

\subsection{ICP with Initial Guess}
\label{sec:icp}

\begin{figure*}
\centering
\footnotesize
\setlength{\tabcolsep}{1cm}
{\renewcommand{\arraystretch}{0.5}
\begin{tabular}{p{7cm}p{7cm}}
\\
\includegraphics[width=\linewidth]{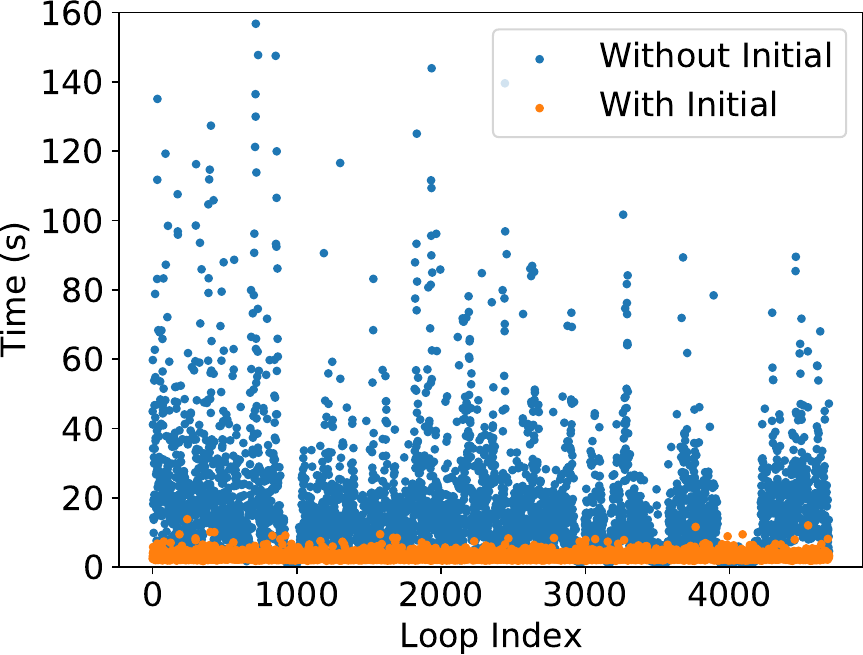} & \includegraphics[width=\linewidth]{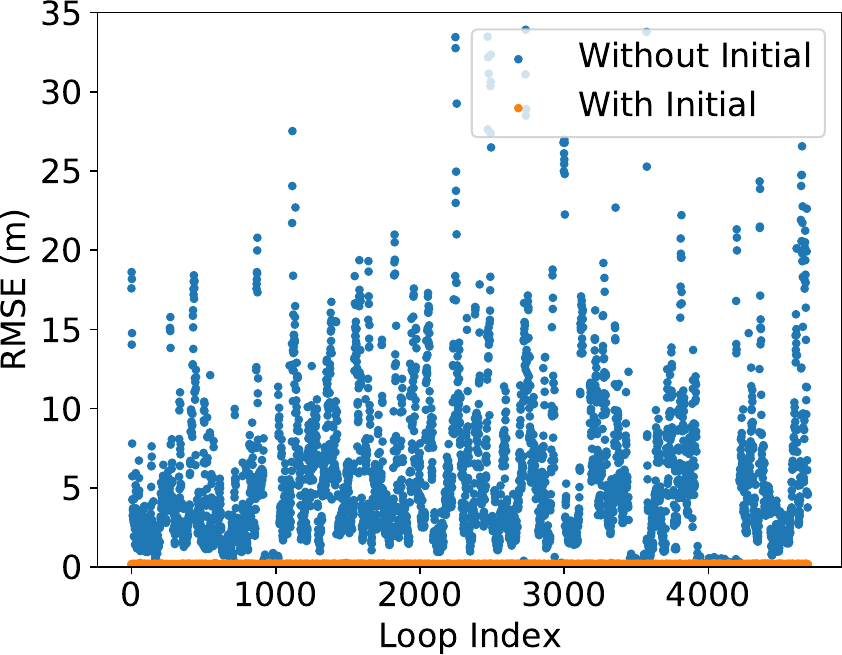} \\
\multicolumn{2}{c}{(c)~KITTI-360 sequence 02} \\
\\
\end{tabular}}
\caption{{Comparison of time (left) and RMSE (right) between ICP without initial guess and ICP with the \net\ prediction as the initial guess on the sequence~02 of the KITTI-360 dataset. Results on other sequences show similar behaviour, and are thus not reported for brevity.
The initial guess provided by our \net\ significantly reduces both runtime and final error on sequences containing reverse loops.}}
  \label{fig:icp}
  \vspace{-3mm}
\end{figure*}

\begin{figure*}
\centering
\scriptsize
\setlength{\tabcolsep}{0cm}
{\renewcommand{\arraystretch}{1}
  \begin{tabular}{P{.2\textwidth}P{.2\textwidth}P{.2\textwidth}P{.2\textwidth}P{.2\textwidth}}
    \includegraphics[width=\linewidth,trim={0 20cm 0 20cm},clip]{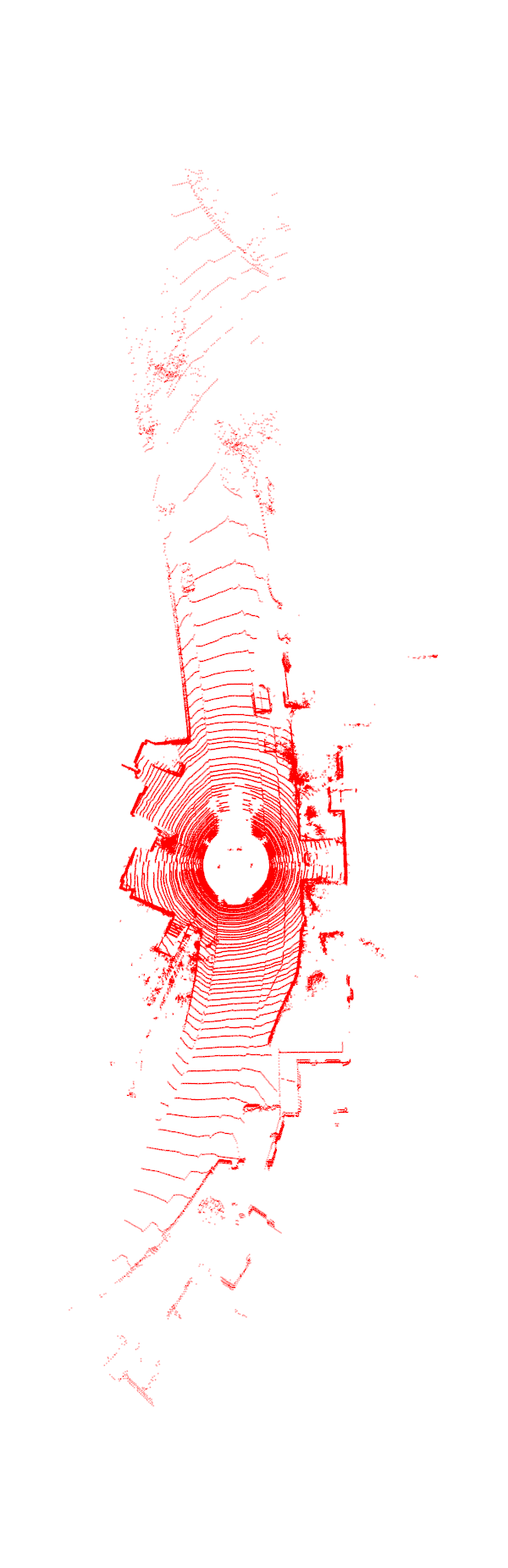} & \includegraphics[width=\linewidth,trim={0 20cm 0 20cm},clip]{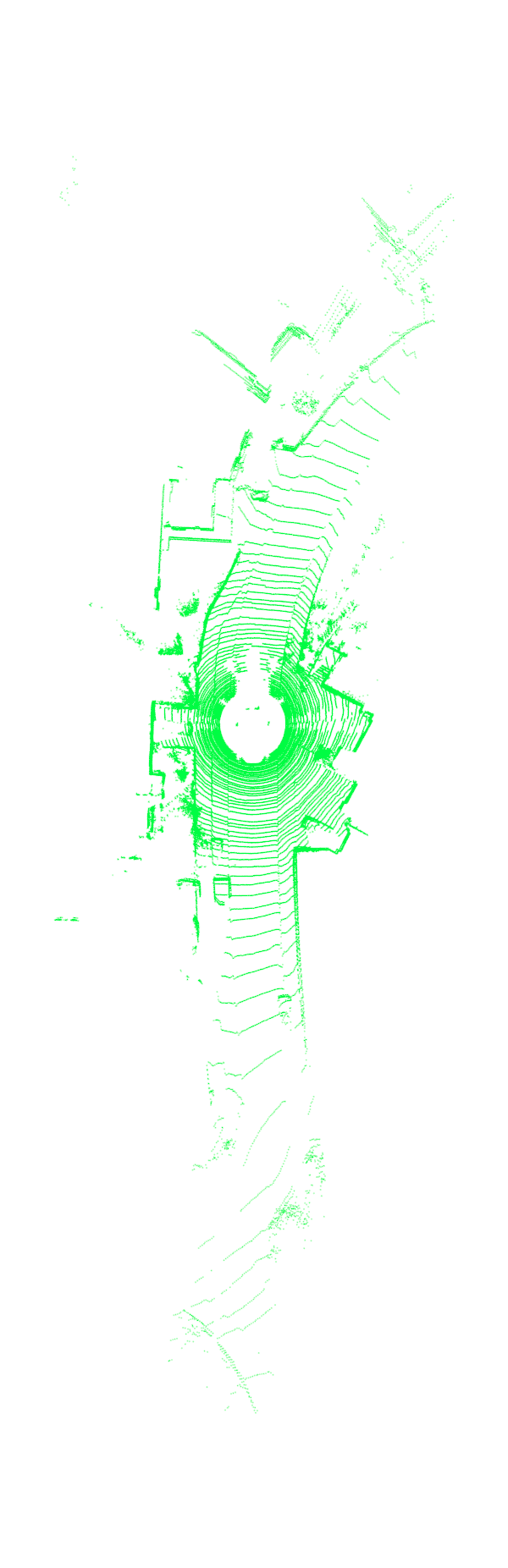} & \includegraphics[width=\linewidth,trim={0 20cm 0 20cm},clip]{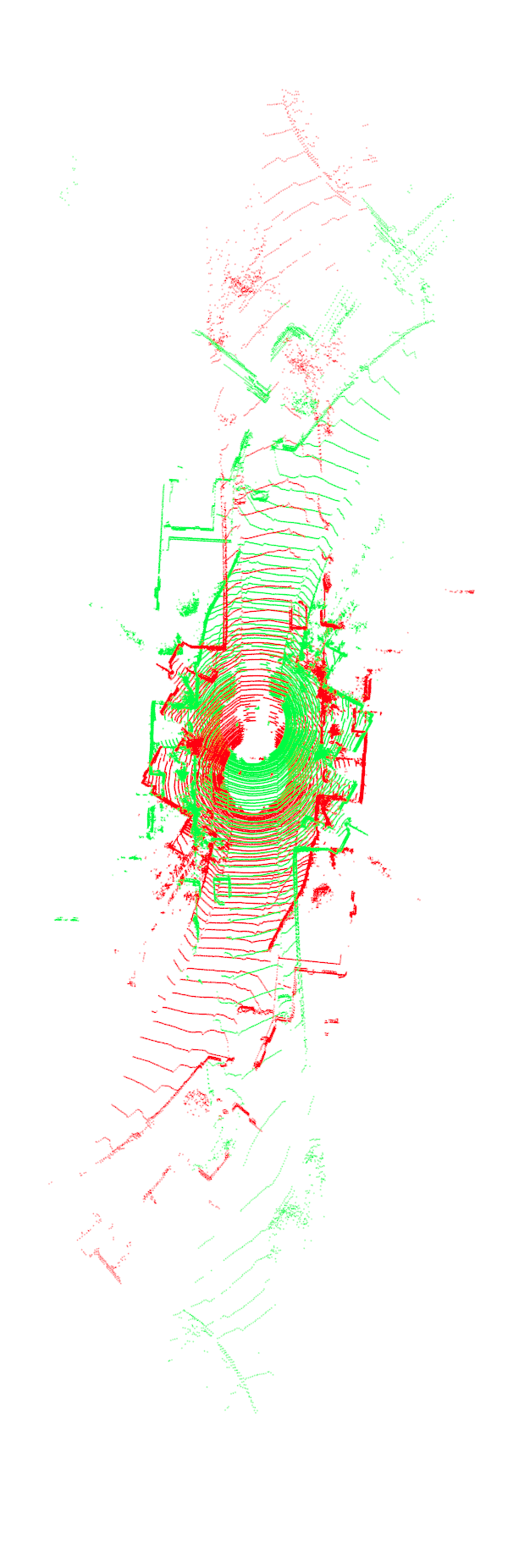} & \includegraphics[width=\linewidth,trim={0 20cm 0 20cm},clip]{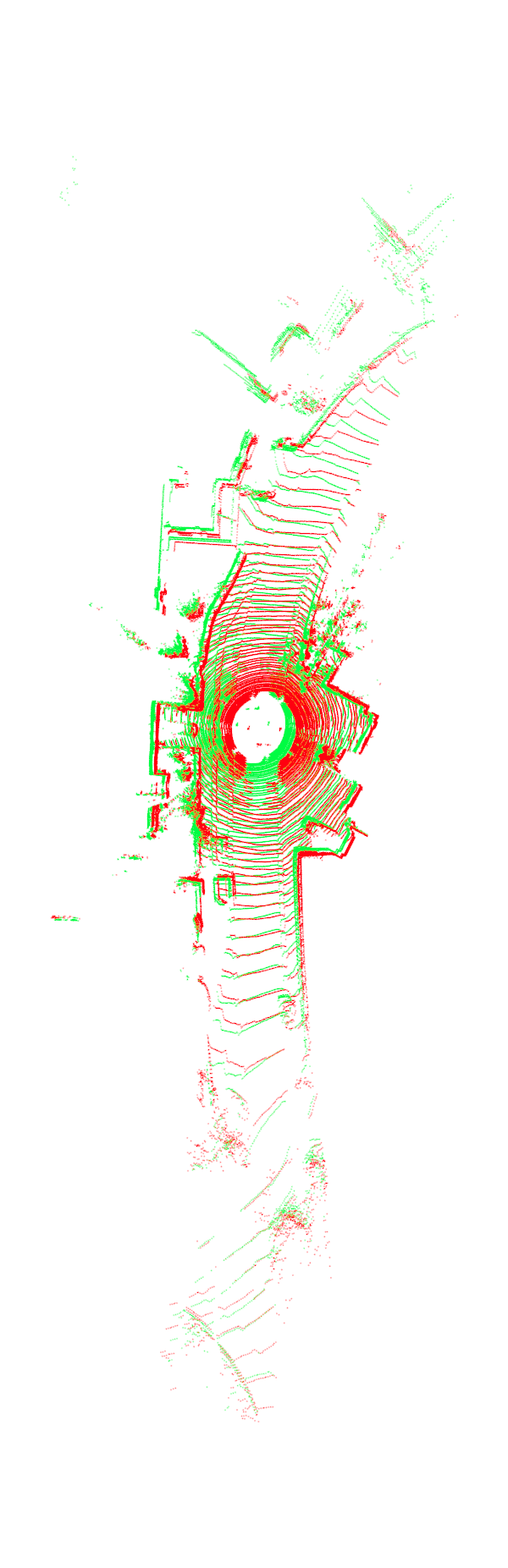} & \includegraphics[width=\linewidth,trim={0 20cm 0 20cm},clip]{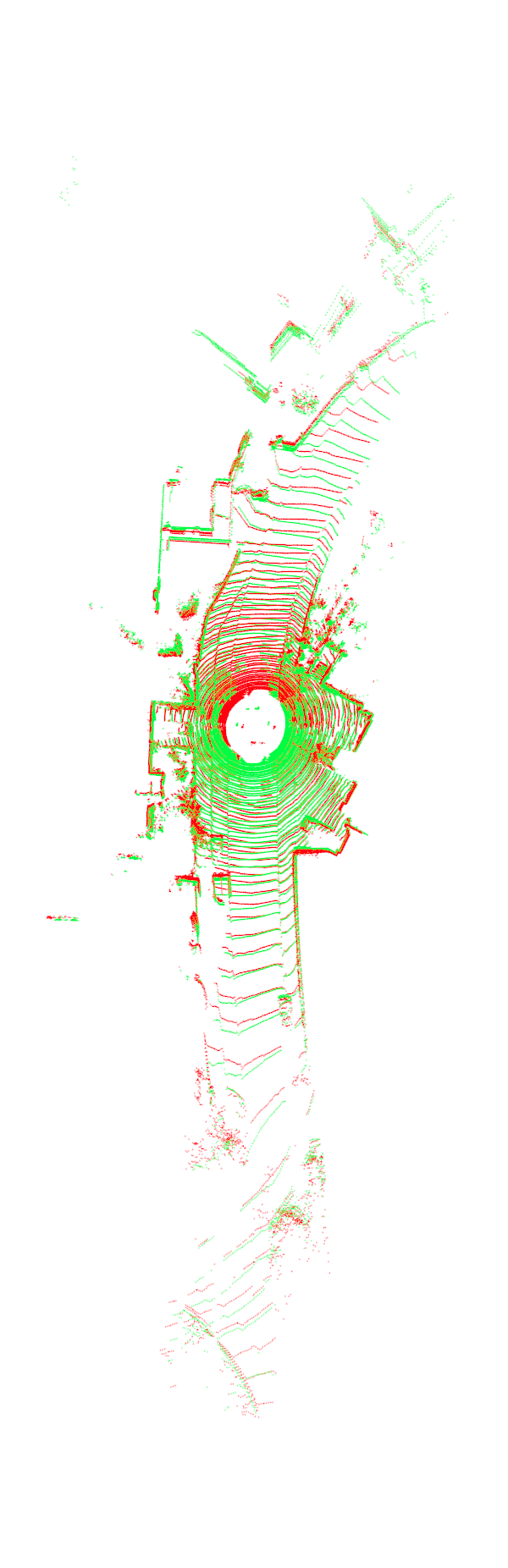} \\
    \\
    \includegraphics[width=\linewidth,trim={25cm 0 10cm 8cm},clip]{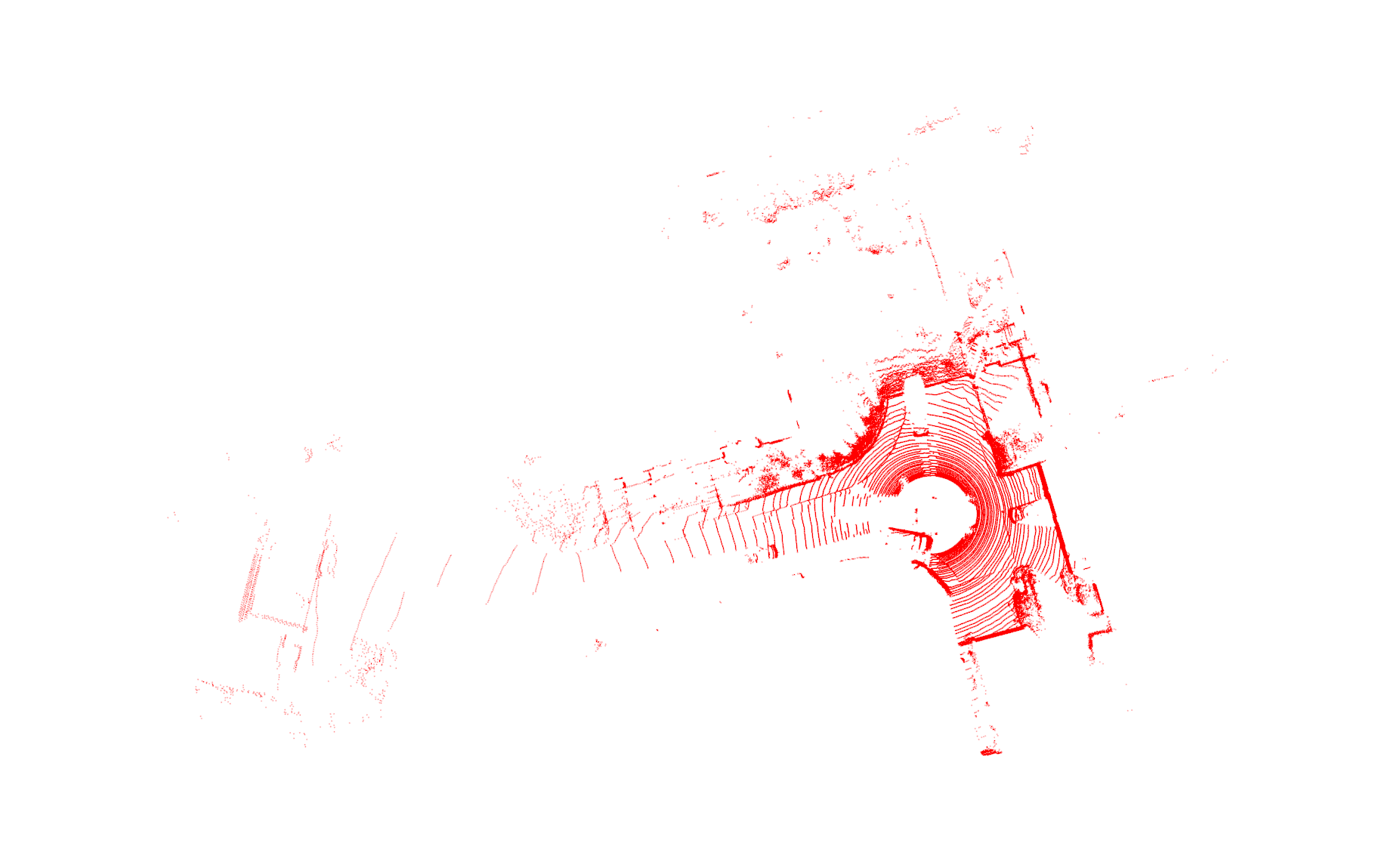} & \includegraphics[width=\linewidth,trim={20cm 0cm 20cm 15cm},clip]{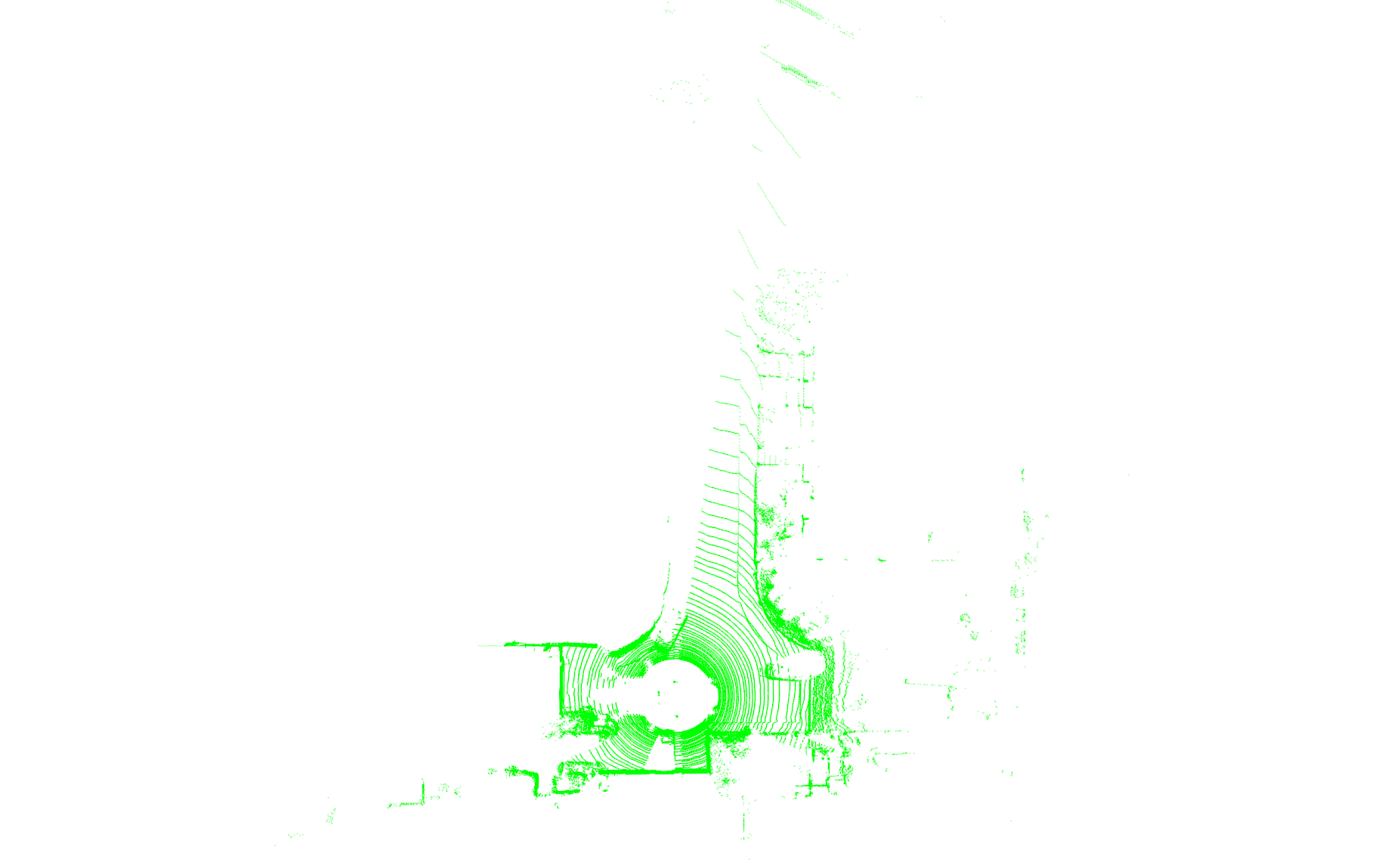} & \includegraphics[width=\linewidth,trim={20cm 0cm 20cm 15cm},clip]{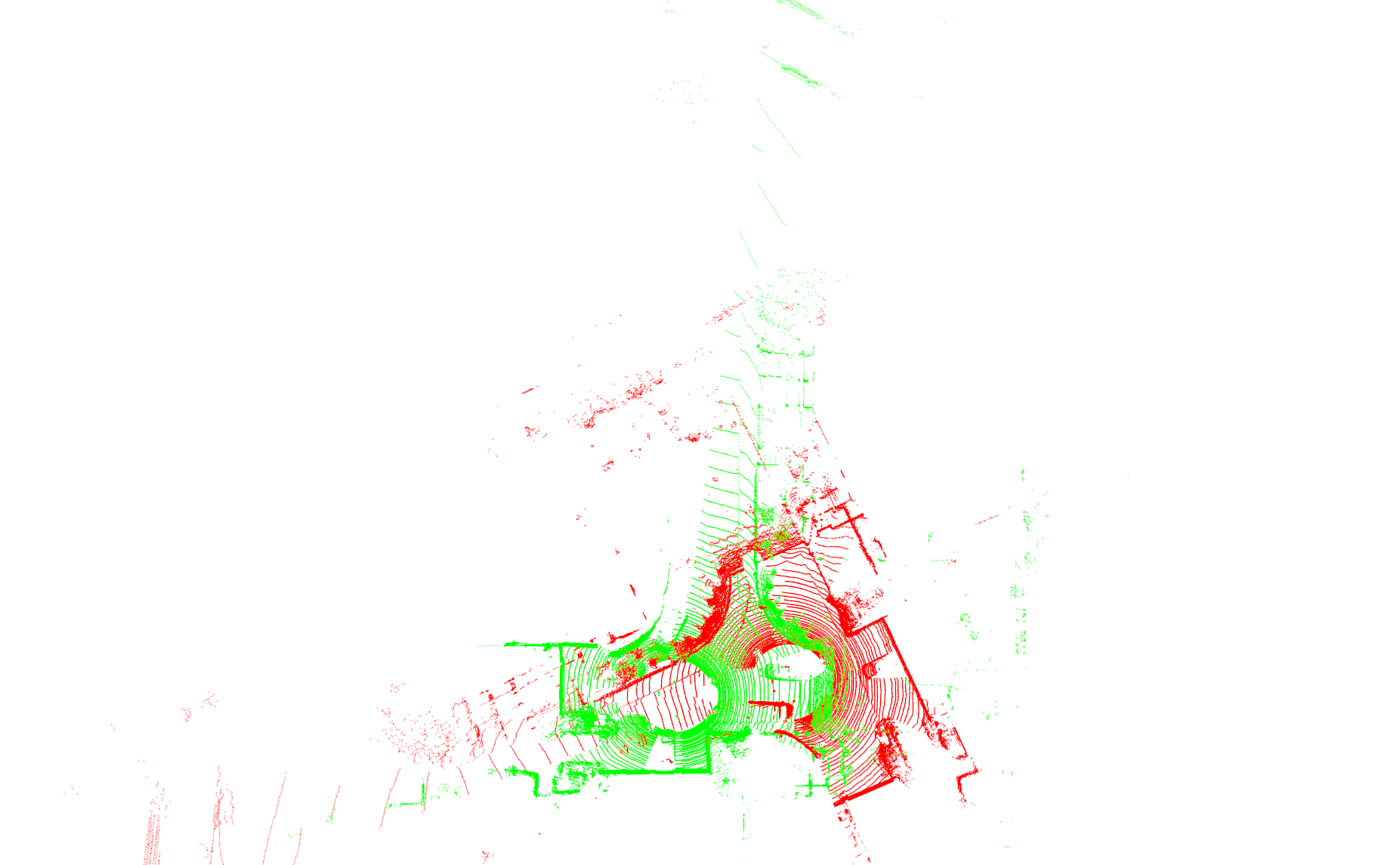} & \includegraphics[width=\linewidth,trim={20cm 0cm 20cm 15cm},clip]{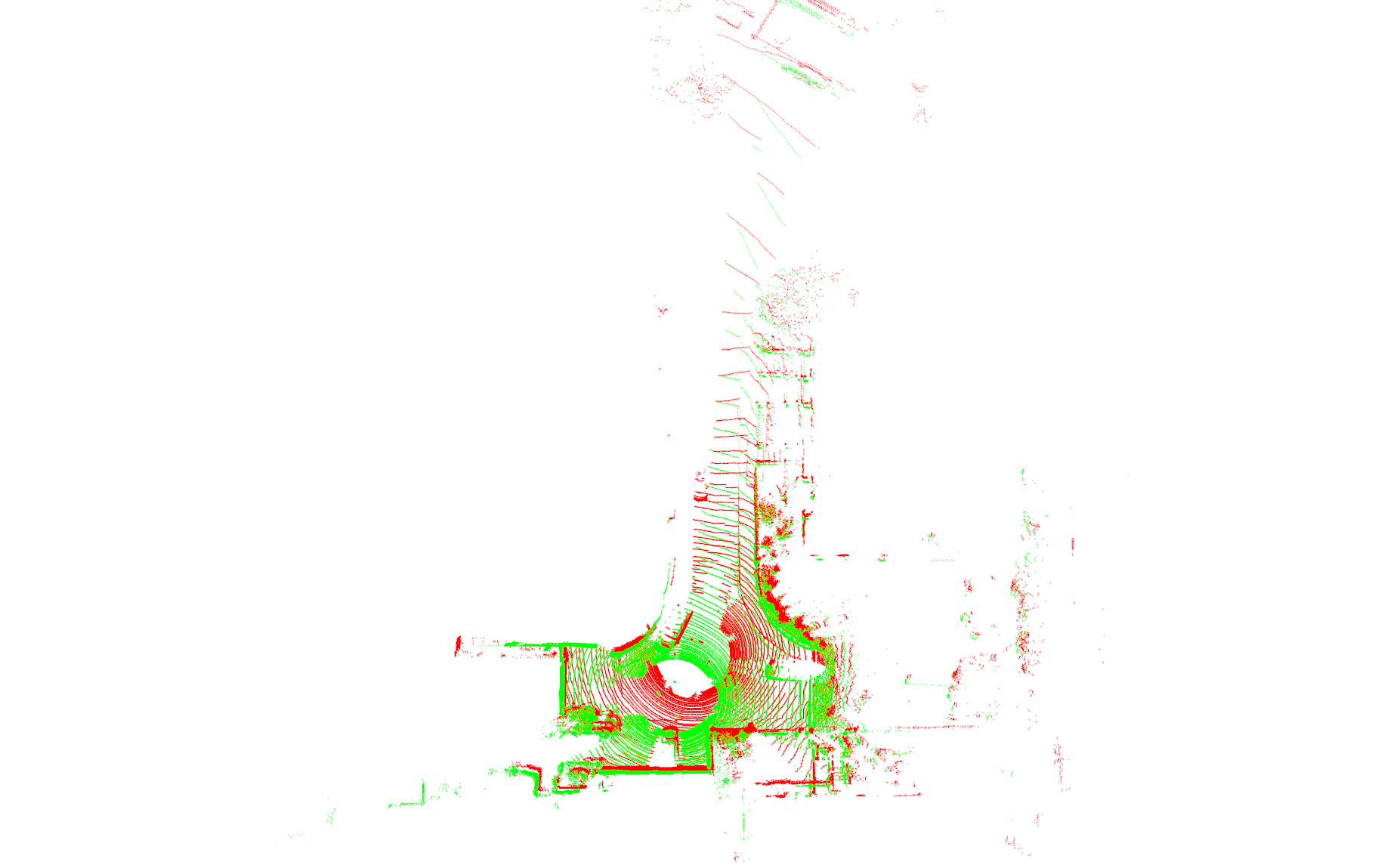} & \includegraphics[width=\linewidth,trim={20cm 0cm 20cm 15cm},clip]{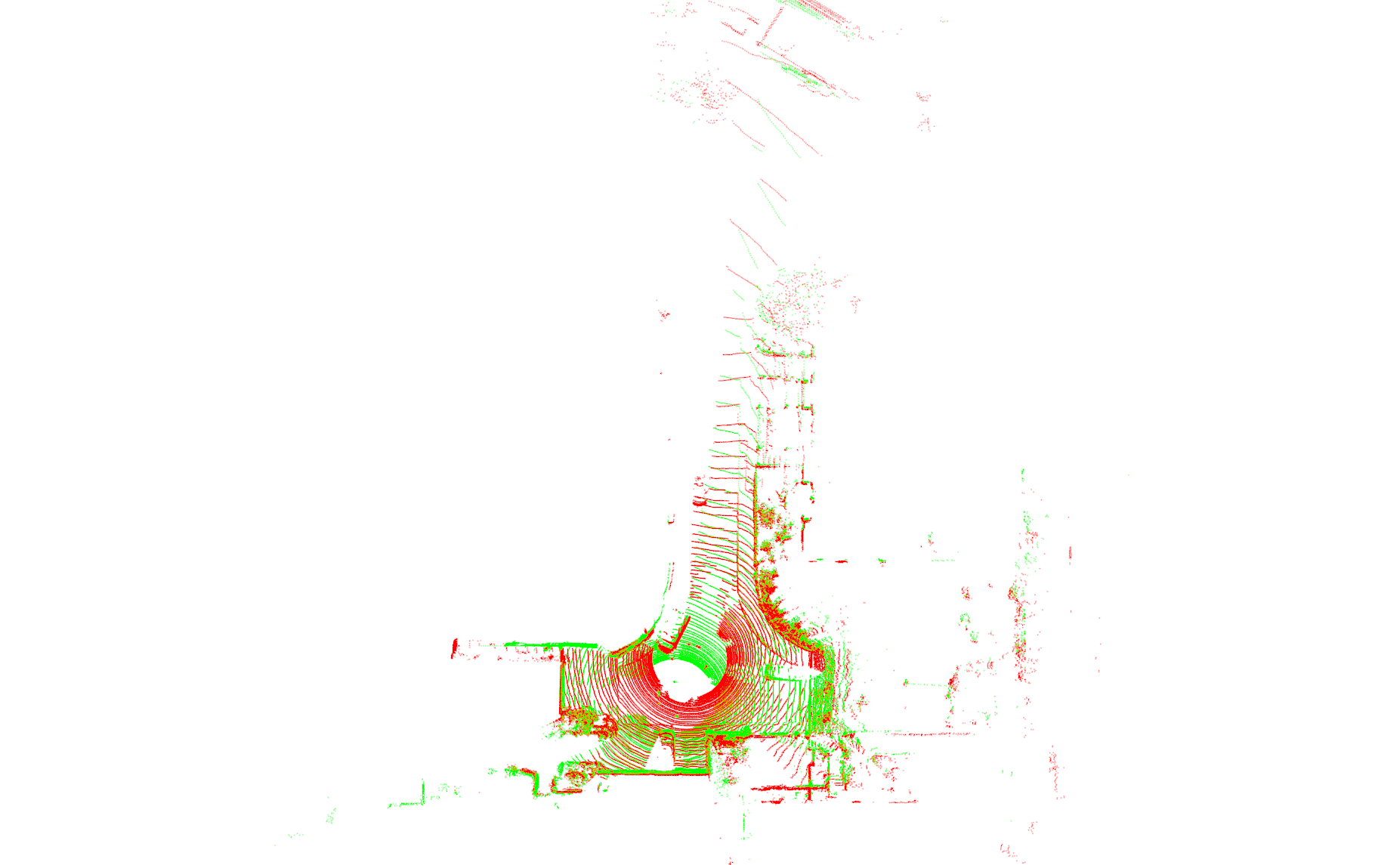} \\
    \\
    (a) Source & (b) Target & (c) ICP without initial pose & (d) Alignment from \net\ & (e) ICP with \net\ initial guess \\
  \end{tabular}}
  \caption{Qualitative comparison of ICP alignment with and without using the \net\ prediction as an initial guess. ICP alone (c) is not able to register the source (a) and the target (b) when the initial rotation misalignment is high. Whereas, our \net\ effectively aligns them (d). The final ICP alignment with the prediction of our \net\ as the initial guess further improve the results (e).}
  \label{fig:icp2}
  \vspace{-3mm}
\end{figure*}

In this experiment, we evaluate the performance of employing \net\ as an initial guess for further refinement using \ac{ICP}. We compare the runtime and the final \ac{RMSE} of \ac{ICP} without any initial guess and \ac{ICP} with \net\ relative pose estimate as an initial guess.
The time of \ac{ICP} with initial guess also includes the network inference time.
Results from this experiment are presented in \cref{fig:icp} and two qualitative results are shown in \cref{fig:icp2}. While only dealing with the same direction loops, \ac{ICP} achieves satisfactory results and the initial guess does not improve the performance significantly.
However, when reverse loops are present, \ac{ICP} often fails in accurately registering the two point clouds. In this case, the initial guess from our \net\ greatly reduces both the runtime and final errors of \ac{ICP} as observed in \cref{fig:icp}.

From \cref{fig:icp2}, we see that \ac{ICP} fails when the rotation misalignment between the two point clouds is significant. On the other hand, \net\ accurately aligns these two point clouds and it improves the results even further while using \ac{ICP} with \net\ prediction as initial guess. On average, \ac{ICP} with \net\ initial guess is 4 times faster than \ac{ICP} without any initial guess and achieves an RMSE which is 22 times lower. Note that in the results presented in \cref{fig:icp}, we use the whole point clouds to perform the registration with \ac{ICP}.

\subsection{Runtime Analysis}
\label{sec:runtime}

\begin{table}
  \centering
  \caption{Comparison of runtime analysis for the loop closure task.
  }
  \label{tab:runtime}
  \setlength\tabcolsep{5pt}
\begin{tabular}{lcccc}
	\toprule
  Method & Descriptor & Pairwise & Map & {GPU} \\
  & Extraction & Comparison & Querying & {Required} \\
  & [ms] & [ms] & [ms] & \\
	\midrule
     M2DP~\cite{He2016} & 169.28 & 0.01 & 5 & \ding{55} \\
	 SC~\cite{Kim2018} & 3.66 & 0.11 & \num{2000} & \ding{55}\\
	 SC-50~\cite{Kim2018} & 3.66 & 0.11 & 6.96 & \ding{55} \\
	 ISC~\cite{9196764} & 1.97 & 0.53 & \num{9000} & \ding{55} \\
	 LiDAR-Iris~\cite{IRIS_2020} & 8.13 & 5.39 & \num{98000} & \ding{55} \\
	 OverlapNet~\cite{chen2020rss} & 16.00 & 6.00 & \num{109000} & \ding{51}\\
	 \textbf{\net} & 94.60 & 0.01 & 5 & \ding{51}\\ \bottomrule
\end{tabular}
\vspace{-3mm}
\end{table}

\begin{table}
  \centering
  \caption{Comparison of runtime analysis for the point cloud registration task.
  }
  \label{tab:runtime_registration}
  \setlength\tabcolsep{4pt}
\begin{tabular}{clcccc}
	\toprule
  &Method & Descriptor & Pairwise & Total [ms] & {GPU} \\
  && Extract. [ms] & Reg. [ms] & &  \\
	\midrule
  \multirow{7}{*}{\rotatebox[origin=c]{90}{\parbox[c]{2cm}{\centering\scriptsize Handcrafted}}}&SC~\cite{Kim2018} & 3.66 & 0.11 & 7.43 & \ding{55} \\
	 &ISC~\cite{9196764} & 1.97 & 0.53 & 4.47 & \ding{55} \\
	 &LiDAR-Iris~\cite{IRIS_2020} & 8.13 & 5.39 & 21.65 & \ding{55} \\
	 &ICP (P2p)~\cite{zhang1994iterative} & - & 25.53 & 25.53 & \ding{55} \\
	 &ICP (P2pl)~\cite{zhang1994iterative} & 8.16 & 35.83 & 52.15 &\ding{55}  \\
	 &RANSAC~\cite{Rusu_ICRA_2009} & 24.99 & 299.66 & 349.64 & \ding{55} \\
	 &FGR~\cite{Zhou_ECCV_2016} & 24.99 & 188.74 & 238.72 & \ding{55} \\
	 &{TEASER++}~\cite{Yang2020tro} & 24.99 & 94.89 & 144.87 & \ding{55} \\\midrule
   \multirow{4}{*}{\rotatebox[origin=c]{90}{\parbox[c]{1.2cm}{\centering\scriptsize DNN-based}}} &OverlapNet~\cite{chen2020rss} & 16.00 & 6.00 & 38.00 & \ding{51} \\
	 &RPMNet~\cite{yew2020} & 366.75 & 121.29 & 854.79 & \ding{51} \\
	 &{DCP~\cite{wang2019deep}} & 19.56 & 78.76 & 117.88 & \ding{51} \\
	 &{PCAM~\cite{cao21pcam}} & 187.71 & 80.77 & 456.18 & \ding{51} \\ \midrule
	 \multirow{2}{*}{\rotatebox[origin=c]{90}{\parbox[c]{0.5cm}{\centering\scriptsize \textbf{Ours}}}}&\textbf{\net\ (fast)} & 94.60 & 4.70 & 193.9 & \ding{51} \\ 
	 &\textbf{\net} & 94.60 & 1135 & 1324.2 & \ding{51} \\ \bottomrule
\end{tabular}
\vspace{-3mm}
\end{table}

In this section, we compare the runtime of \net\ with existing state-of-the-art approaches for loop detection. All experiments were performed on a system with an Intel i7-6850K CPU and an NVIDIA GTX 1080 ti GPU. We use the official implementation of existing approaches as described in \Cref{sec:experiment_lcd,sec:comparison-pos}. Results from this experiment are presented in \Cref{tab:runtime} in which the descriptor extraction time also includes the preprocessing required by the respective method. The pairwise comparison represents the time required to compare the descriptors of two point clouds. In the map querying column, we report the time for comparing the descriptor of one scan with that of all the previous scans in the KITTI-360 sequence~02, which amounts to \num{18235} comparisons in total. For methods that do not require an ad-hoc function to compare descriptors (\net\ and M2DP), we use the efficient FAISS library~\cite{8733051} for similarity search in order to build and query the map.
Scan Context also introduces the \textit{ring key} descriptors which enable fast search for finding loop candidates, at the expense of detection performances. We also report the runtime of scan context using the ring key, denoted as \textit{Scan Context-50}. However, it is important to note that the results reported in \ref{sec:experiment_lcd} were computed without the ring key.

As shown in~\Cref{tab:runtime}, the methods that require an ad-hoc comparison function (ISC, LiDAR-Iris, and OverlapNet) are not suited for real-time applications, since they require up to 100 seconds to perform a single query. Whereas, \net\ queries more than \num{18000} scans in five milliseconds. Although it is possible to further reduce the time required to query the map when integrating the loop closure approaches in a SLAM system, such as using the covariance-based radius search~\cite{chen2020rss}, in this experiment we evaluate the runtime in the case where no prior information about the current pose is available.

We report the runtime for aligning two point clouds by \net\ and existing approaches in \cref{tab:runtime_registration}.
For methods that perform both loop closure and point cloud alignment (SC, ISC, LiDAR-IRIS, OverlapNet, and \net), the descriptor extraction time is shared between the two tasks.
{While \net\ (fast) is faster than most \ac{DNN}-based approaches for point cloud registration (RPMNet and PCAM), \net\ is slightly slower than RPMNet.
On the other hand, some approaches are much faster than both \net\ and \net\ (fast); however, they either only estimate a 1-DoF transformation (SC, ISC, and LiDAR-Iris), or achieve unsatisfactory performances (ICP, RANSAC, FGR, TEASER++, and DCP).}
It is important to note that the point cloud registration task does not need to run in realtime, since it is only required after a loop closure is detected.
Moreover, \net\ is the only method that performs both loop closure detection and 6-DoF point cloud registration {under driving conditions}.

\begin{figure*}
  \centering
  \footnotesize
  \setlength{\tabcolsep}{1cm}
  {\renewcommand{\arraystretch}{0.5}
  \begin{tabular}{p{7cm}p{7cm}}
  \multicolumn{1}{c}{\net} & \multicolumn{1}{c}{\net$_\dagger$} \\
  \\
  \includegraphics[width=\linewidth]{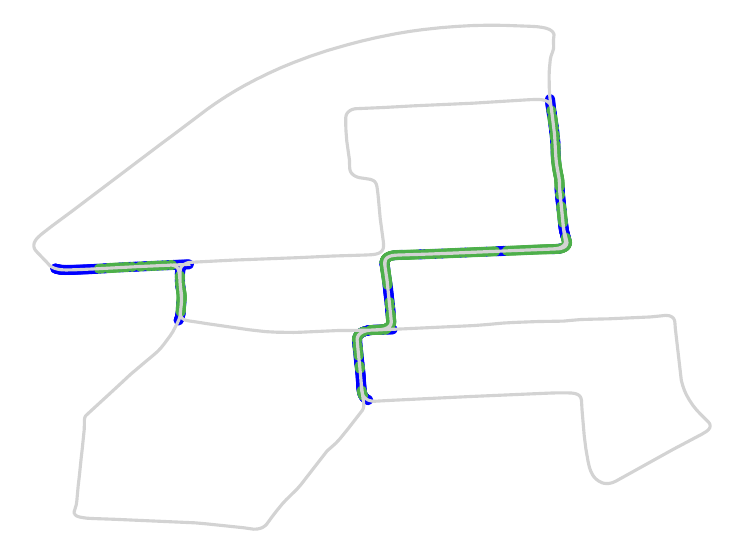} & \includegraphics[width=\linewidth]{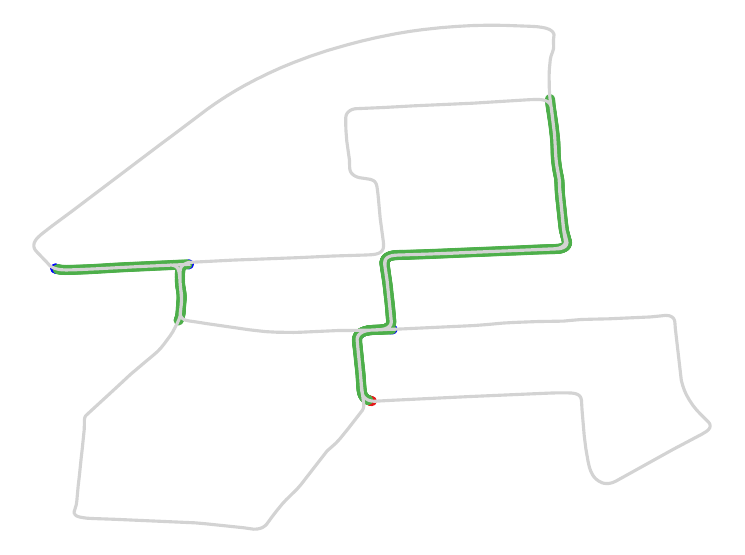} \\
  \multicolumn{2}{c}{(a)~KITTI sequence 00} \\
  \\
  \includegraphics[width=\linewidth]{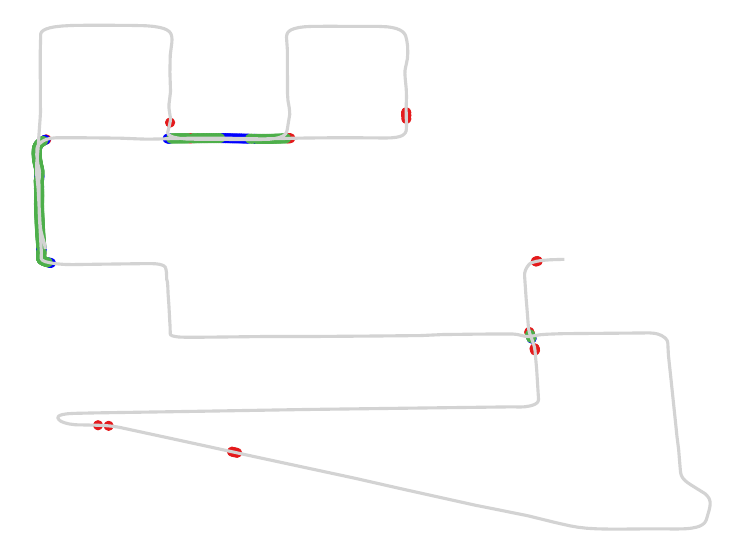} & \includegraphics[width=\linewidth]{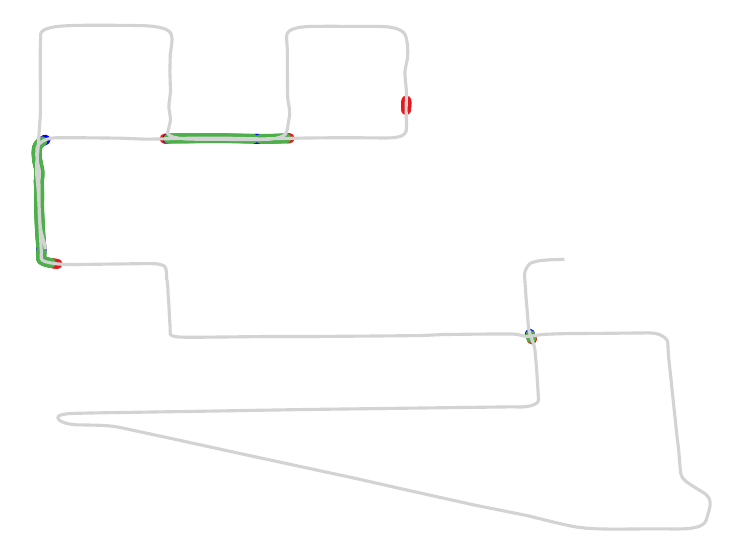} \\
  \multicolumn{2}{c}{(b)~KITTI sequence 08} \\
  \\
  \includegraphics[width=\linewidth]{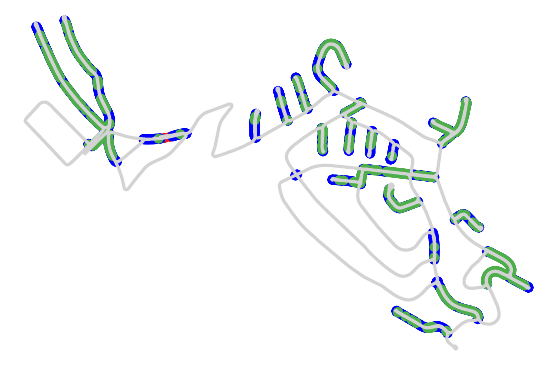} & \includegraphics[width=\linewidth]{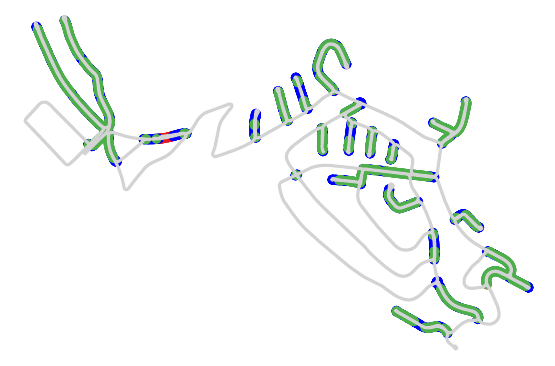} \\
  \multicolumn{2}{c}{(c)~KITTI-360 sequence 02} \\
  \\
  \includegraphics[width=\linewidth]{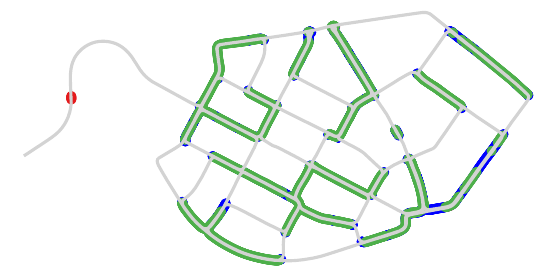} & \includegraphics[width=\linewidth]{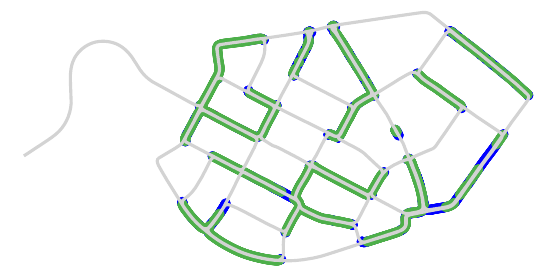} \\
  \multicolumn{2}{c}{(d)~KITTI-360 sequence 09} \\
  \\
  \end{tabular}}
  \caption{Qualitative loop closure detection results of \net\ on KITTI (a-b) and KITTI-360 (c-d) datasets. Green points \tikzcircle[pathgreen,fill=pathgreen]{3pt} are true positive detections, red points \tikzcircle[pathred,fill=pathred]{3pt} are false positive, and blue points \tikzcircle[blue,fill=blue]{3pt} are false negative. The left column shows results of \net\ trained on the KITTI dataset, while the right columns shows results of \net$_\dagger$ trained on the KITTI-360 dataset. While both \net\ and \net$_\dagger$ effectively detects loops in all the sequences, \net$_\dagger$ further reduces the number of false positive and false negative detections.}
  \label{fig:prpath}
\end{figure*}

\subsection{Qualitative Results}

\begin{figure}
  \footnotesize
  \setlength{\tabcolsep}{0.0cm}
  {\renewcommand{\arraystretch}{0.5}
  \begin{tabular}{p{3.775cm}p{4.925cm}}
  \includegraphics[width=\linewidth]{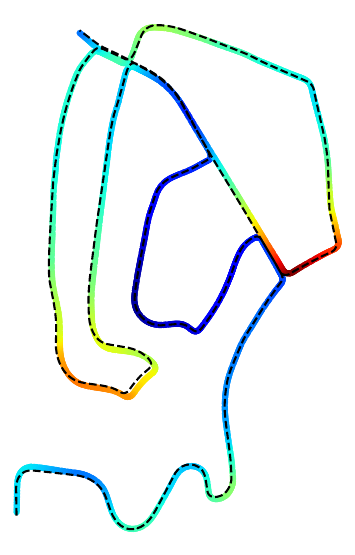} & \includegraphics[width=\linewidth]{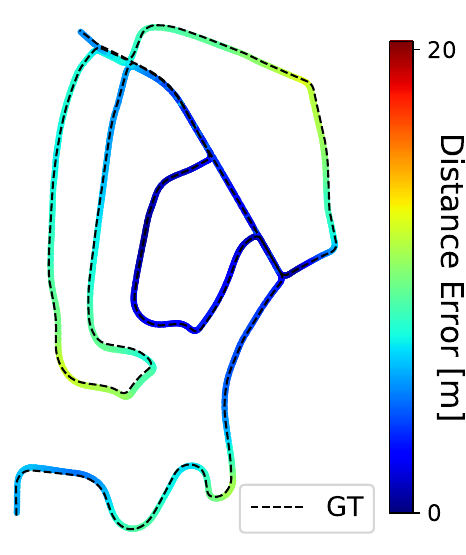} \\
  \end{tabular}}
\caption{Performance of LIO-SAM with the original loop closure detection method (left) compared to our approach (right) on sequence~02 of the KITTI dataset.
}
\label{fig:lio-sam-drift}
\vspace{-3mm}
\end{figure}

We present the qualitative results from \net\ and \net$_\dagger$ on sequences from both the KITTI and KITTI-360 datasets in \Cref{fig:prpath}. We show the true positive, false positive, and false negative scans overlaid with the respective groundtruth trajectories.
We observe that while \net\ effectively detects same direction and reverse direction loops, it also fails to detect some loops (false negative) and detects some loops where there should be no loops (false positive). \net$_\dagger$ further improves the performance by reducing the number of false positives and false negatives, while still maintaining accurate true positive detections. On the KITTI sequence 08, \net\ yields some false negative detections that are almost completely eliminated by \net$_\dagger$, although few false positive scans are still detected. On sequence 02 of the KITTI-360 dataset, \net\ presents a large amount of false negatives which are significantly reduced by \net$_\dagger$. Similarly, on the KITTI-360 sequence 09, \net\ presents a few false positive detections that are completely eliminated by \net$_\dagger$.

\subsection{Evaluation of Complete SLAM System}

We integrate our proposed approach into LIO-SAM~\cite{liosam2020shan}, which is a recent state-of-the-art LiDAR SLAM system, by replacing its loop closure detection pipeline with \net.
We evaluate the entire SLAM system on the sequence~02 of the KITTI dataset. In particular, we evaluate LIO-SAM integrated with \net\ and we compare it with the original LIO-SAM. We observed that our approach detects loop closures where the original LIO-SAM fails to do so due to the presence of the accumulated drift. In \cref{fig:lio-sam-drift}, we report the results obtained with both SLAM systems and show the distance error between LIO-SAM keyframes and groundtruth poses. We can observe that the high error (red) associated with the path of the original LIO-SAM is caused by the failed loop closure detection, since the system drifts significantly along the z-dimension. Conversely, \net\ detects such loops, perform the closure and improve the overall performance of LIO-SAM.

We publicly release the integrated LIO-SAM system with our \net\ at \url{http://rl.uni-freiburg.de/research/lidar-slam-lc}.

\subsection{Generalization Analysis}

\begin{figure}
\footnotesize
\centering
\includegraphics[width=0.9\linewidth]{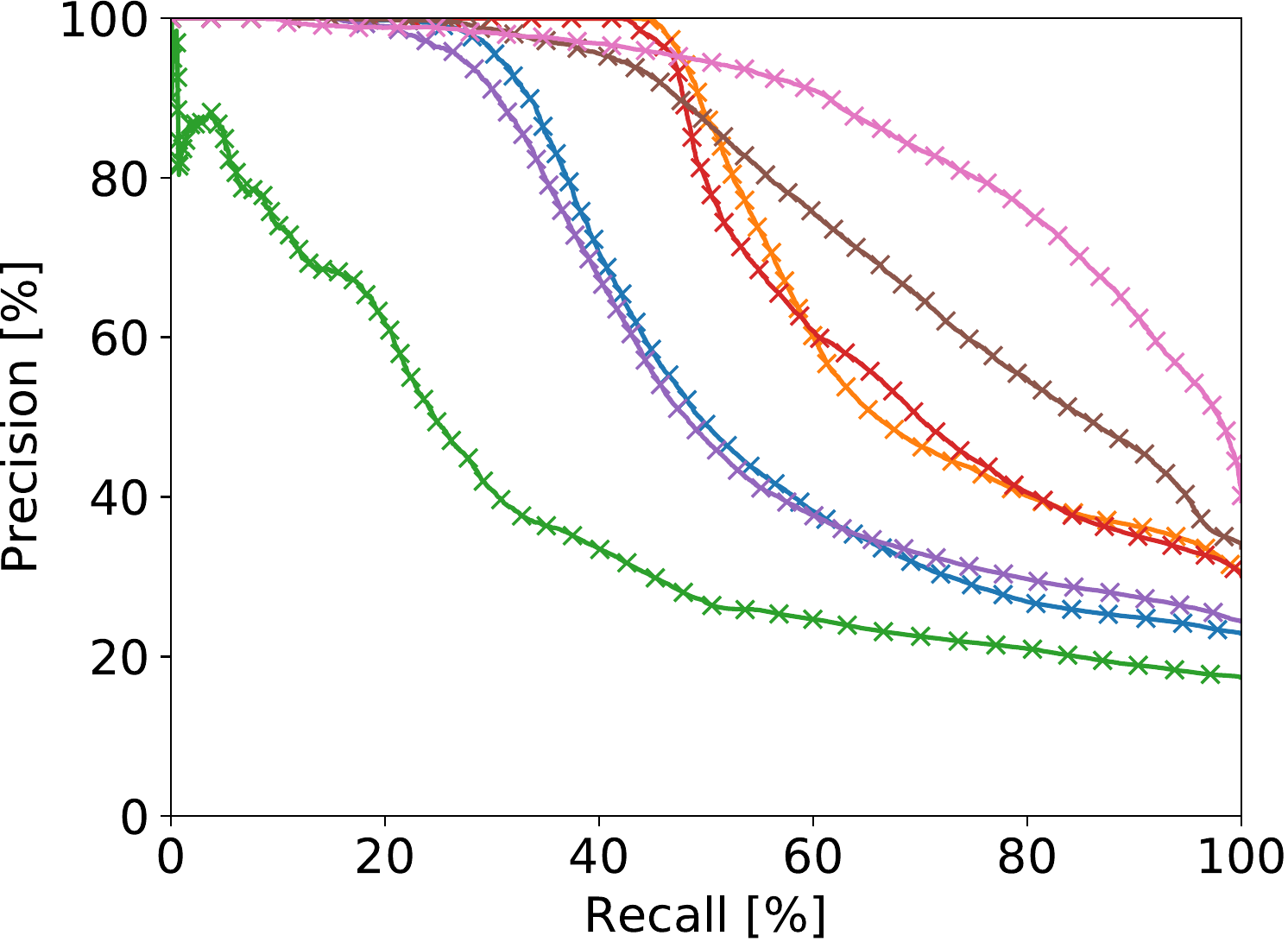}\\ \vspace{0.1cm}
\includegraphics[width=\linewidth]{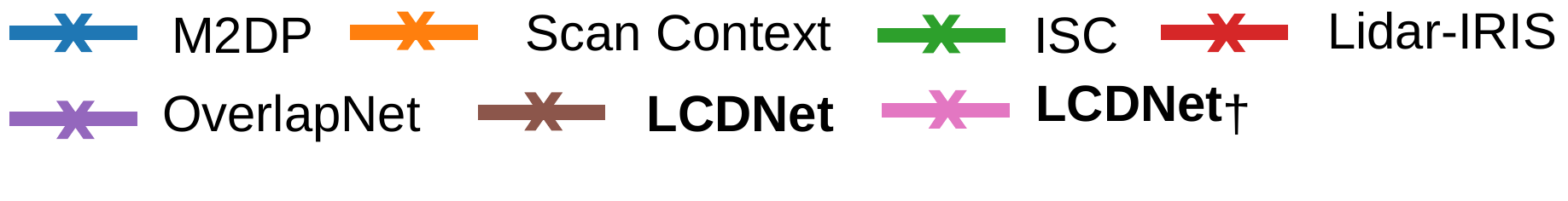}
\caption{Comparison of precision-recall curves evaluated using protocol~1 on data from the generalization experiments in Freiburg.
 }
\label{fig:pr-freiburg}
\vspace{-3mm}
\end{figure}

\begin{table}
  \centering
  \caption{Comparison with the state of the art on data from the generalization experiments in Freiburg.}
  \label{tab:AP-freiburg}
  \setlength\tabcolsep{15pt}
\begin{tabular}{clc}
	\toprule
  & Method & AP  \\
	\midrule
     \multirow{4}{*}{\rotatebox[origin=c]{90}{\parbox[c]{1cm}{\centering\scriptsize Handcrafted}}} & M2DP~\cite{He2016} & 0.60 \\
	 &Scan Context~\cite{Kim2018} & 0.74 \\
	 &ISC~\cite{9196764} & 0.38  \\
	 &LiDAR-Iris~\cite{IRIS_2020} & 0.73  \\ \midrule
	 \multirow{3}{*}{\rotatebox[origin=c]{90}{\parbox[c]{1cm}{\centering\scriptsize DNN-based}}} & OverlapNet~\cite{chen2020rss} & 0.59  \\
	 &\textbf{\net} & \underline{0.79} \\
	 &\textbf{\net$_\dagger$} & \textbf{0.88} \\ \bottomrule
\end{tabular}
\vspace{-3mm}
\end{table}

\begin{figure*}
  \centering
  \footnotesize
  \setlength{\tabcolsep}{0.2cm}
  {\renewcommand{\arraystretch}{0.5}
  \begin{tabular}{p{9.2cm}p{7.2cm}}
 \multicolumn{2}{c}{\includegraphics[width=0.925\linewidth]{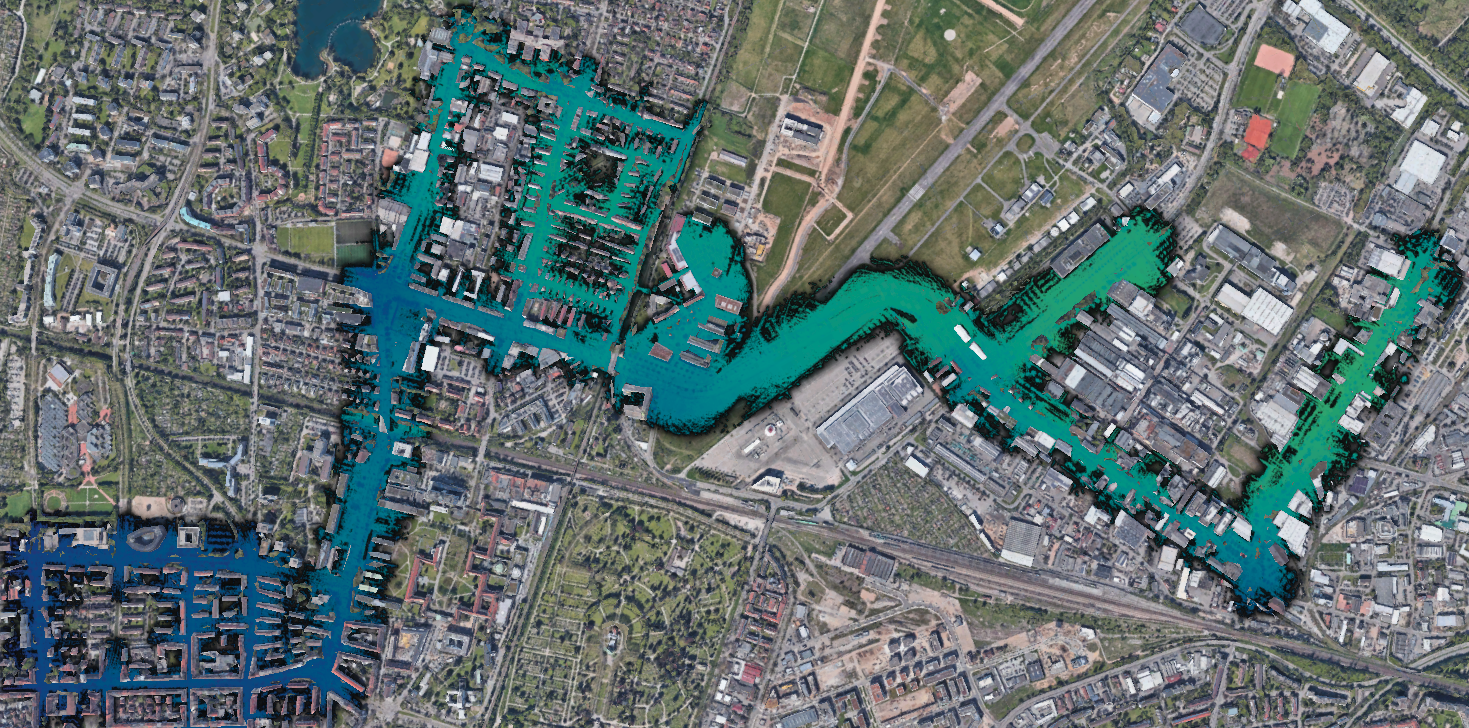}} \\
 \\
\includegraphics[width=\linewidth,trim={0 0 0 2cm},clip]{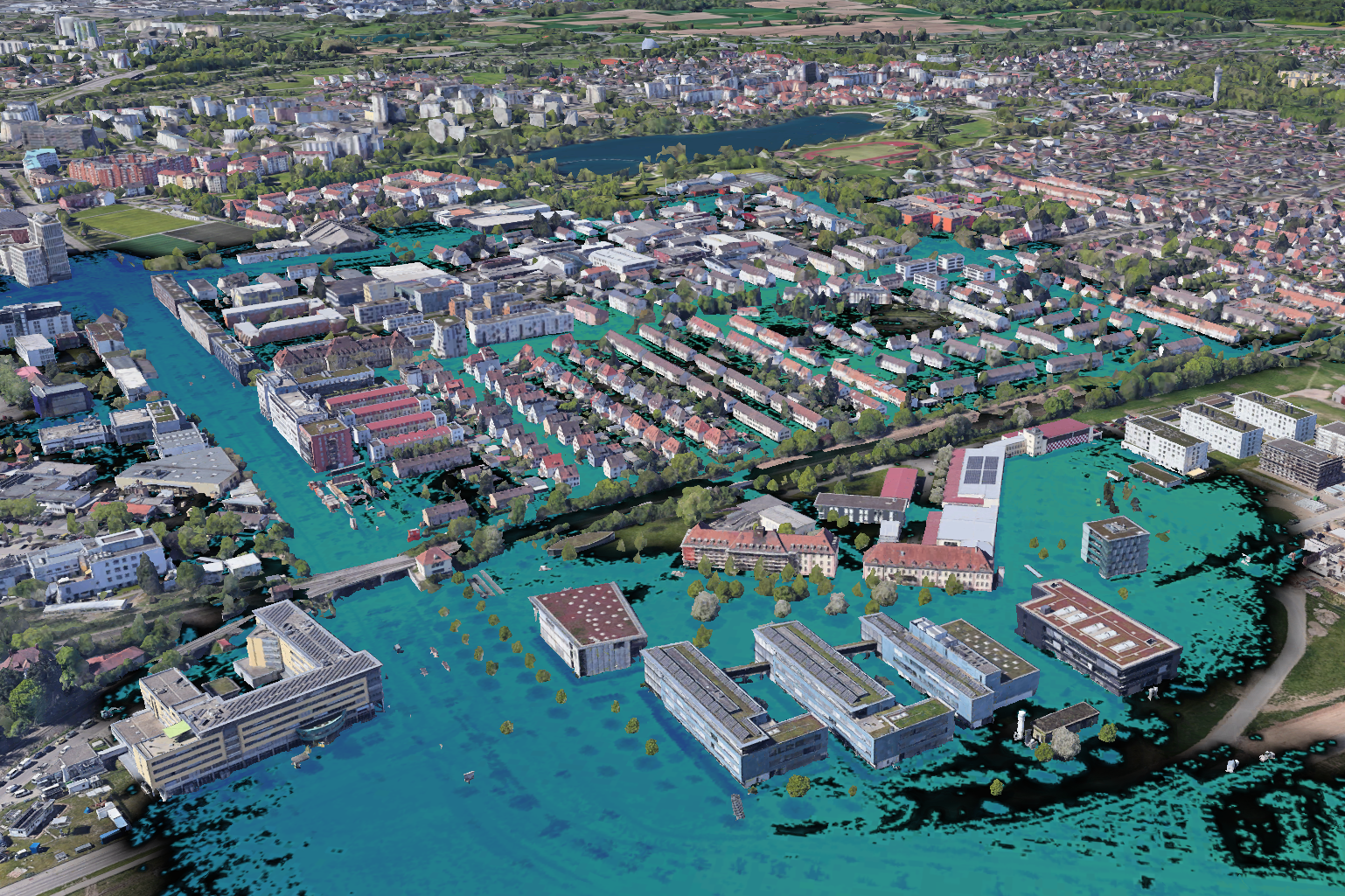} & \includegraphics[width=\linewidth]{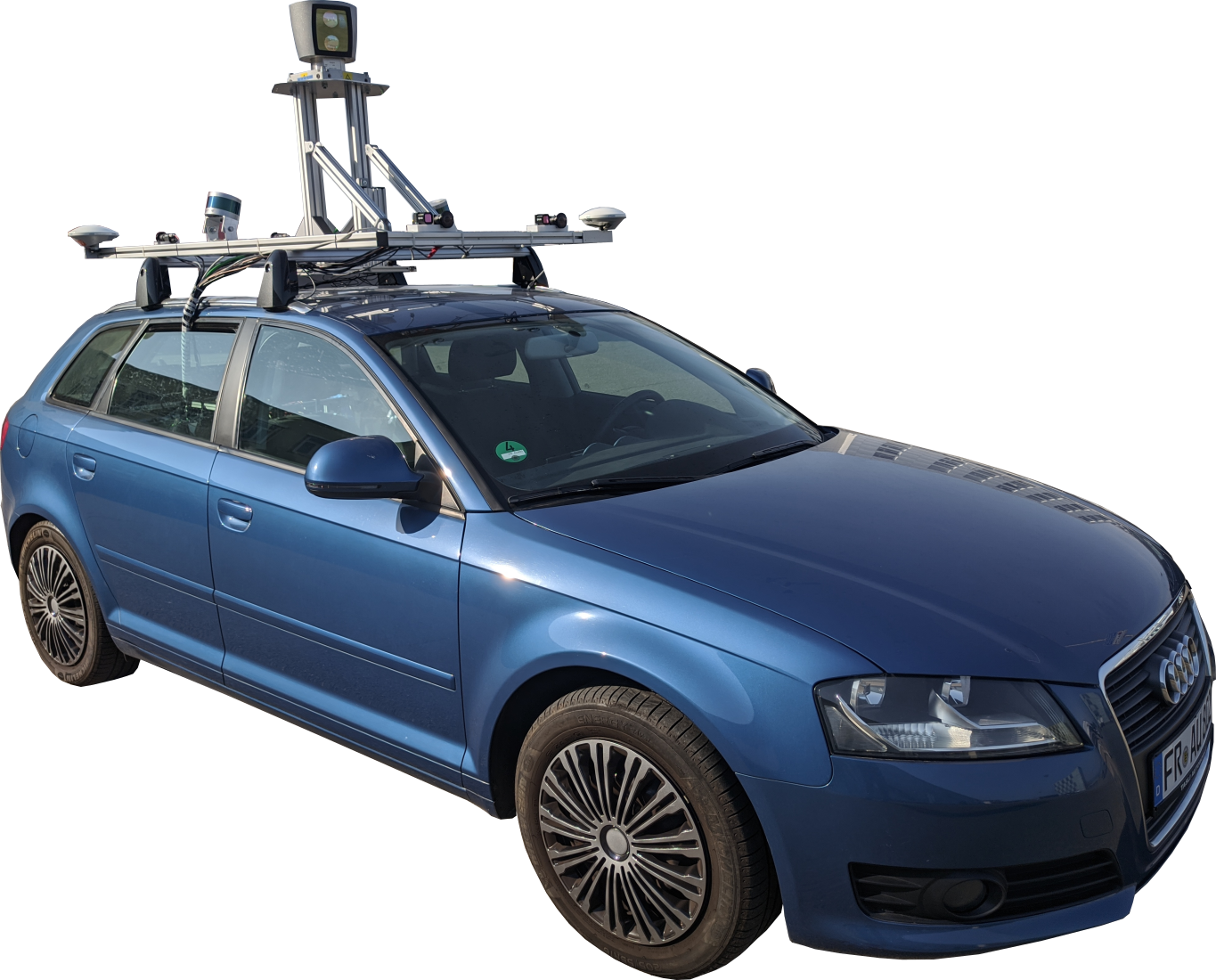} \\
  \end{tabular}}
  \caption{Qualitative results of our approach on data from the generalization experiments in Freiburg. The final map generated from LIO-SAM integrated with our \net\ is overlaid on the georeferenced aerial images. 
  Image on the top shows the entire map, while the images in the bottom show zoomed in segments and the car used to collect the dataset. {The color of the point cloud is based on the Z-coordinates of the points from lowest (blue) to highest (green).}
  }
  \label{fig:freiburgMap}
\vspace{-3mm}
\end{figure*}

Finally, in this section, we evaluate the generalization ability of our proposed \net\ by analyzing the performance in unseen environments and on different robot platforms. We evaluate both \net\ and \net$_\dagger$ in real-world experiments in Freiburg using a car with a rack of LiDAR sensors mounted on the roof as shown in \cref{fig:freiburgMap} (bottom right). Note that in these experiments, we do not retrain or fine-tune \net\ and \net$_\dagger$ on any data from Freiburg. {The KITTI and KITTI-360 datasets on which we trained our models on, were primarily recorded in narrow roads, but the streets of Freiburg also include dual carriageways, therefore we increase the range at which two scans are considered to be a real loop from 4 to 10 meters.}
In \cref{fig:pr-freiburg}, we compare the precision-recall curves of our approach with handcrafted and DNN-based methods using protocol~1 (\cref{sec:experiment_lcd}). We do not report results using protocol~2 for this experiment as there are more than 400~million positive pairs in this trajectory.

As shown in \Cref{tab:AP-freiburg}, Scan Context achieves the best performance among the handcrafted methods, with an AP of 0.74. Nevertheless, our \net\ and \net$_\dagger$ outperform all the other approaches achieving an AP of 0.79 and 0.88, respectively. Since the data from Freiburg consists of many reverse loops, existing approaches often fail to detect them, leading to a decrease in their performance. Our approach demonstrates exceptional performance even though it has never seen scans from Freiburg during training.
Moreover, we employ our modified version of LIO-SAM to generate the trajectory and the map of the experimental runs in Freiburg. In \cref{fig:freiburgMap}, we show the resulting map overlaid on the aerial image. The results show that the map is well aligned with the aerial image and there is no evidence of any drift. This demonstrates that our \net\ effectively corrects the accumulated drift.
{It is important to note that the precision-recall curve and the AP of our \net\ are computed based only on the global descriptor extracted by the place recognition head. However, in the modified SLAM system, we additionally perform a consistency check (\cref{sec:slam}) based on the transformation predicted by the relative pose head which further discard the remaining false positive detections.}

Finally, we also exploit the Freiburg dataset to demonstrate the point cloud alignment ability of our approach in new environments. Since we do not have an accurate pose for each LiDAR frame, we generate groundtruth transformations using the GPS poses and ICP, and discarding pairs that produce an inaccurate alignment. First, for each frame we identify possible pairs by considering its neighbors within a distance of $10m$. In order to avoid the pairs that are composed of consecutive frames, given a point cloud we discard the previous and the following $n=100$ frames. Secondly, for each pair we compute the yaw angle difference $\Delta_{yaw}$ and define three difficulty levels. Then, for each frame we select a random pair for every category whenever possible. We set the maximum number of ICP iterations $n_{icp} = 1000$, and we only consider pairs with a fitness score $fit >= 0.6$ and an inlier correspondences $rmse <= 0.3 m$. Finally, we randomly sample the resulting pairs to have about the same number of same direction and reverse direction loops. The resulting number of pairs amounts to 4246, of which 2106 are reverse loops.

{ The results reported in \cref{tab:generalization-pose} show that our \net\ effectively generalizes to new environments for the point cloud registration task. \net\ and \net$_\dagger$ achieve a success rate of 98.94\% and 99.81\%, respectively, compared to the second best method that achieves 92.49\%.
The overall mean translation error of \net$_\dagger$ is more than two times smaller, and the rotation error is an order of magnitude lower than PCAM.}

\begin{table}
  \centering
  \caption{Comparison of relative pose errors (rotation and translation) between positive pairs on the Freiburg dataset.}
  \label{tab:generalization-pose}
  \begin{threeparttable}
  \begin{tabular}{clccc}
  \toprule
   & Approach & Success & TE [m]  & RE [deg]  \\
   & & & (succ. / all) & (succ. / all) \\ \midrule
   \multirow{7}{*}{\rotatebox[origin=c]{90}{\parbox[c]{2cm}{\centering\scriptsize Handcrafted}}} &Scan Context~\cite{Kim2018} & 59.30\% & - / - & 1.36 / 52.70 \\
   & ISC~\cite{9196764} & 55.51\% & - / - & 1.52 / 51.02 \\
  &LiDAR-Iris~\cite{IRIS_2020} & 69.95\% & - / - & 1.52 / 51.02 \\ 
  & ICP (P2p)~\cite{zhang1994iterative} & 29.06\% & 0.83 / 2.60 & 1.21 / 89.79 \\
  & ICP (P2pl)~\cite{zhang1994iterative} & 28.73\% & 0.88 / 2.62 & 1.22 / 89.83 \\
  & RANSAC~\cite{Rusu_ICRA_2009} & 29.96\% & 1.01 / 3.54 & 1.34 / 31.29 \\
  & FGR~\cite{Zhou_ECCV_2016} & 27.72\% & 0.97 / \num{313258} & 1.27 / 13.46 \\
  & {TEASER++}~\cite{Yang2020tro} & {29.49\%} & {0.99 / 3.37} & {1.31 / 11.94} \\\midrule
  \multirow{3}{*}{\rotatebox[origin=c]{90}{\parbox[c]{1.2cm}{\centering\scriptsize DNN-based}}} &OverlapNet~\cite{chen2020rss} & 42.79\% & - / - & 1.31 / 70.91 \\ 
  &RPMNet~\cite{yew2020} & 32.05\% & 0.87 / 2.57 & 1.09 / 46.99 \\
  &{DCP~\cite{wang2019deep}} & {12.25\%} & {1.26 / 5.22} & {1.19 / 87.04} \\
&{PCAM~\cite{cao21pcam}} & {92.49\%} & {0.40 / 0.67} & {0.50 / 4.28} \\
   \midrule
  \multirow{2}{*}{\rotatebox[origin=c]{90}{\parbox[c]{0.5cm}{\centering\scriptsize \textbf{Ours}}}}
  & \net & \underline{98.94\%} & \underline{0.39 / 0.42} & \underline{0.32 / 0.37} \\
  & \net$_\dagger$ & \textbf{99.81\%} & \textbf{0.28 / 0.28} & \textbf{0.18 / 0.18} \\
  \midrule
  & \net\ + ICP & 99.79\% & 0.22 / 0.23 & 0.16 / 0.16 \\
  & \net$_\dagger$ + ICP & 99.86\% & 0.21 / 0.22 & 0.14 / 0.14 \\
  & {\net\ + TEASER} & {33.61\%} & {1.09 / 3.74} & {0.17 / 0.42} \\
  & {\net$_\dagger$ + TEASER} & {33.37\%} & {1.15 / 4.45} & {0.13 / 0.27} \\  
  \bottomrule
  \end{tabular}
   \end{threeparttable}
\vspace{-3mm}
\end{table}


\section{Conclusions}
\label{sec:conclusions}

In this paper, we presented the novel \net\ architecture for loop closure detection and point cloud registration. \net\ is composed of a shared feature extractor built upon the PV-RCNN network, a place recognition head that captures discriminative global descriptors, and a novel differentiable relative pose head based on the unbalanced optimal transport theory which effectively aligns two point clouds without any prior information regarding their initial misalignment. We identified a discrepancy in the evaluation protocols of existing methods, therefore we performed uniform evaluations of state-of-the-art handcrafted as well as \ac{DNN}-based loop closure detection methods.

We presented extensive evaluations of \net\ on the KITTI odometry and KITTI-360 datasets, which demonstrates that our approach sets the new state-of-the-art and successfully detects loops even in challenging conditions such as reverse direction loops, where existing methods fail. Our \net$_\dagger$ achieves an average precision of $0.96$ on the sequence~08 of the KITTI dataset which contains only reverse direction loops, compared to $0.65$ AP of the previous state-of-the-art method.
Our proposed relative pose head demonstrates impressive results, outperforming existing approaches for point clouds registration and loop closure detection as well as different heads based on the standard \acp{MLP}.
{ Our \net\ aligns opposite direction point clouds with an average rotation error of \ang{0.34}, and \SI{0.15}{\meter} for the translation components, compared to \ang{1.84} and \SI{0.41}{\meter} achieved by LiDAR-IRIS and PCAM, respectively. Moreover, \net\ is robust to partial overlapping point cloud, retaining a 100\% success rate when removing a \ang{90} sector from each point cloud, while the second best method drop from 95\% to 55\%. We also showed that the relative pose prediction from our approach can further be refined using ICP for accurate registration. We integrated our \net\ with LIO-SAM to provide a complete SLAM system which can detect loops even in presence of strong drift. Additionally, we demonstrated the generalization ability of our approach by evaluating it on the data from experiments using a different robotic platform and in an unseen city from that which was used for training. Finally, we have made the code and the SLAM system publicly available to encourage research in this direction.}

\section*{Acknowledgments}

This work was partly funded by the Federal Ministry of Education and Research (BMBF) of Germany under SORTIE, and by the Eva Mayr-Stihl Stiftung. The authors would like to thank Johan Vertens for his assistance in the data collection.


\bibliographystyle{IEEEtran}
\begin{small}
\bibliography{references.bib}
\end{small}

\begin{small}
\vspace{-33pt}
\begin{IEEEbiography}[{\includegraphics[width=1in,height=1.25in,clip,keepaspectratio]{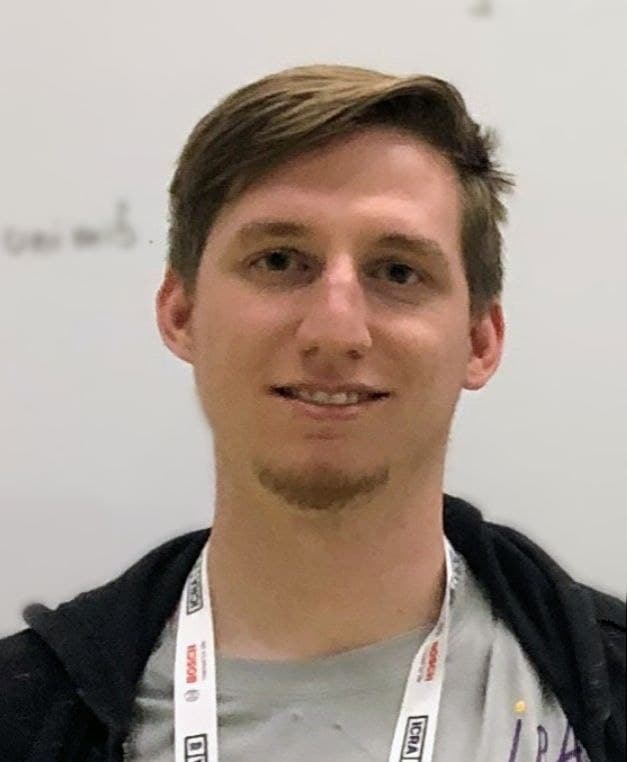}}]{Daniele Cattaneo}
  received the M.S.~degree in Computer Science from the University of Milano-Bicocca, Milan, Italy, in 2016 and the Ph.D.~degree in Computer Science from the same university in 2020.
  He is currently a Postdoctoral Researcher with the Robotic Learning Lab of the University of Freiburg, Freiburg, Germany, headed by Abhinav Valada.
  His research interest includes deep learning for robotic perception and state estimation, with a focus on sensor fusion, cross-modal matching, and domain generalization.

\end{IEEEbiography}

\vspace{-40pt}
\begin{IEEEbiography}[{\includegraphics[width=1in,height=1.25in,clip,keepaspectratio]{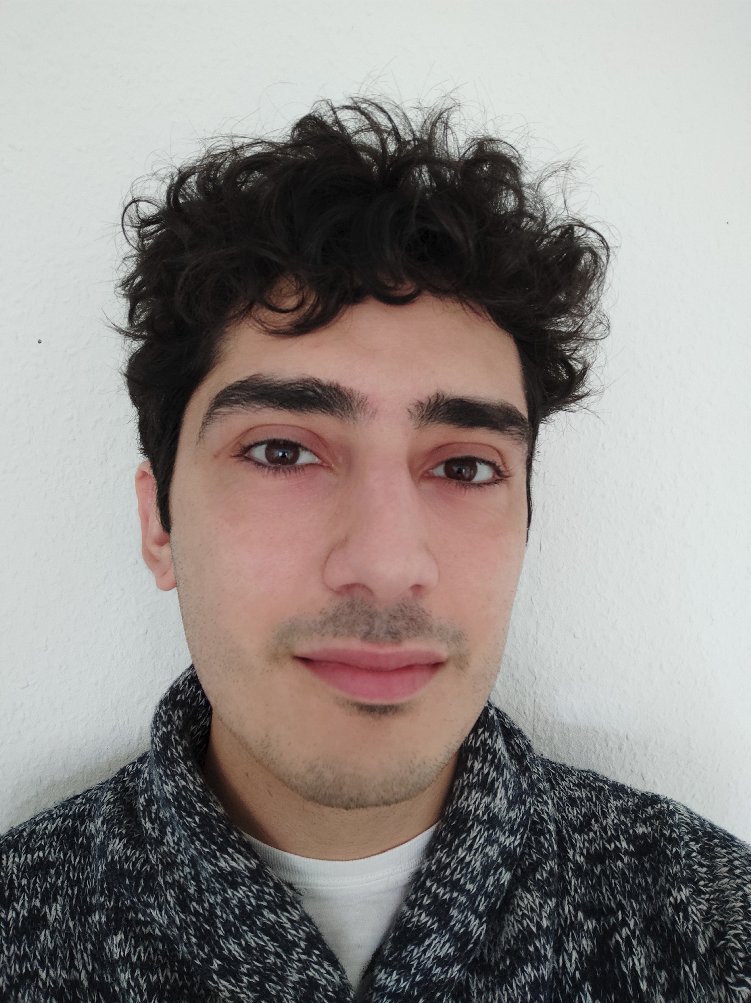}}]{Matteo Vaghi}
  received the B.S. and M.S. degrees in Computer Science from the University of Milano-Bicocca, Milan, Italy, in 2016 and 2019, respectively.
  He was a junior research assistant at the IRALab Research Group at the same university in 2019 and 2020. Currently, He is a Ph.D. student at the University of Milano-Bicocca and his research focuses on the development of techniques for addressing the vehicle localization problem in urban areas. In particular, his main research topics are computer vision, deep learning and robotics.
\end{IEEEbiography}

\vspace{-33pt}
\begin{IEEEbiography}[{\includegraphics[width=1in,height=1.25in,clip,keepaspectratio]{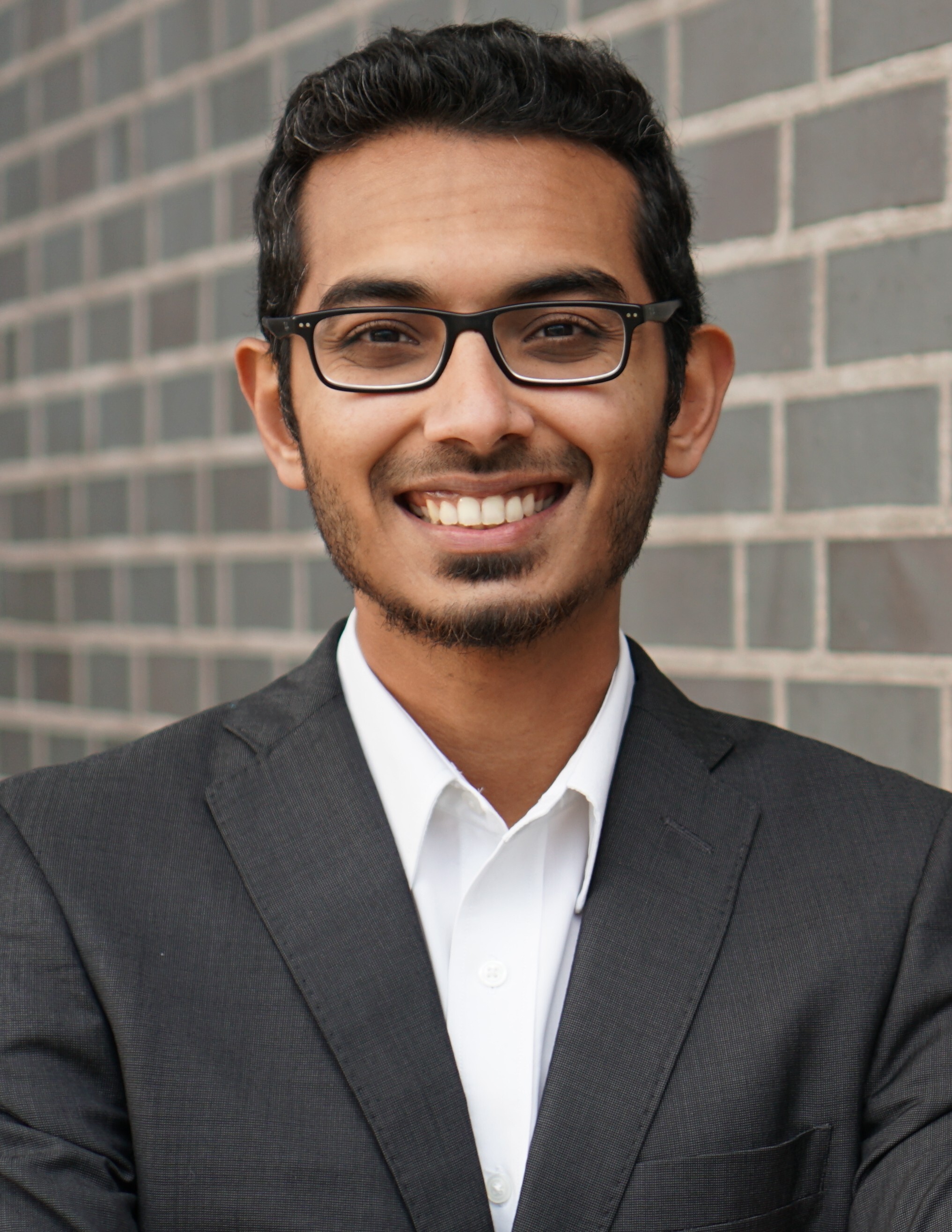}}]{Abhinav Valada}
  is an Assistant Professor and Director of the Robot Learning Lab at the University of Freiburg. He is a member of the Department of Computer Science, a principal investigator at the BrainLinks-BrainTools Center, and a founding faculty of the European Laboratory for Learning and Intelligent Systems (ELLIS) unit at Freiburg. He received his Ph.D.~in Computer Science from the University of Freiburg in 2019 and his M.S.~degree in Robotics from Carnegie Mellon University in 2013. His research lies at the intersection of robotics, machine learning and computer vision with a focus on tackling fundamental robot perception, state estimation and control problems using learning approaches in order to enable robots to reliably operate in complex and diverse domains. Abhinav Valada is a Scholar of the ELLIS Society, a DFG Emmy Noether Fellow, and co-chair of the IEEE RAS TC on Robot Learning.
\end{IEEEbiography}
\end{small}

\end{document}